\documentclass{article}
\usepackage[final]{neurips_2020}
\usepackage[utf8]{inputenc} 
\usepackage[T1]{fontenc}    
\usepackage{hyperref}       
\usepackage{url}            
\usepackage{booktabs}       
\usepackage{amsfonts}       
\usepackage{nicefrac}       
\usepackage{microtype}      

\usepackage{times}
\usepackage{epsfig}
\usepackage{graphicx}
\usepackage{amsmath}
\usepackage{amsthm}
\usepackage{amssymb}

\usepackage{multirow}
\usepackage{bm}
\usepackage{bbm}
\usepackage{mathtools}
\usepackage{enumitem}
\usepackage{kantlipsum}
\usepackage{caption}
\usepackage{subfigure}
\usepackage{comment}
\usepackage{wrapfig}
\usepackage{tikz}
\usetikzlibrary{calc}

\newtheorem{pro}{Proposition}

\newcommand{\tikzmark}[1]{\tikz[overlay,remember picture] \node (#1) {};}

\graphicspath{{./figs/}}

\title{Calibrating Deep Neural Networks using Focal Loss}

\author{
   Jishnu Mukhoti\thanks{Joint first authors, order decided by coin flip. Contact: \{jishnu, viveka,  puneet, phst\}@robots.ox.ac.uk, amartya.sanyal@cs.ox.ac.uk, sgolodetz@gxstudios.net}\\
   University of Oxford\\
   FiveAI Ltd.\\
   \And
   Viveka Kulharia$^*$ \\
   University of Oxford \\
   \AND
   Amartya Sanyal \\
   University of Oxford \\
   The Alan Turing Institute \\
   \And
   Stuart Golodetz\\
   FiveAI Ltd. \\
   \And
   Philip H. S. Torr \\
   University of Oxford \\
   FiveAI Ltd. \\
   \And
   Puneet K. Dokania \\
   University of Oxford \\
   FiveAI Ltd.
}

\begin{document}

\maketitle

\begin{abstract}
 \noindent Miscalibration -- a mismatch between a model's confidence and its correctness -- of Deep Neural Networks (DNNs) makes their predictions hard to rely on. Ideally, we want networks to be accurate, calibrated and confident. We show that, as opposed to the standard cross-entropy loss, focal loss~\citep{Lin2017} allows us to learn models that are already very well calibrated. When combined with temperature scaling, whilst preserving accuracy, it yields state-of-the-art calibrated models. We provide a thorough analysis of the factors causing miscalibration, and use the insights we glean from this to justify the empirically excellent performance of focal loss. To facilitate the use of focal loss in practice, we also provide a principled approach to automatically select the hyperparameter involved in the loss function. We perform extensive experiments on a variety of computer vision and NLP datasets, and with a wide variety of network architectures, and show that our approach achieves state-of-the-art calibration without compromising on accuracy in almost all cases. Code is available at \url{https://github.com/torrvision/focal_calibration}.
\end{abstract}

\vspace{-5mm}
\section{Introduction}
\label{sec:introduction}
\vspace{-2mm}

Deep neural networks have dominated computer vision and machine learning in recent years, and this has led to their widespread deployment in real-world systems \citep{Cao2018,Chen2018,Kamilaris2018,Ker2018,Wang2018}. However, many current multi-class classification networks in particular are poorly calibrated, in the sense that the probability values that they associate with the class labels they predict overestimate the likelihoods of those class labels being correct in the real world. This is a major problem, since if networks are routinely overconfident, then downstream components cannot trust their predictions. The underlying cause is hypothesised to be that these networks' high capacity leaves them vulnerable to overfitting on the negative log-likelihood (NLL) loss they conventionally use during training \citep{Guo2017}.

Given the importance of this problem, numerous suggestions for how to address it have been proposed. Much work has been inspired by approaches that were not originally formulated in a deep learning context, such as Platt scaling~\citep{Platt1999}, histogram binning~\citep{Zadrozny2001}, isotonic regression~\citep{Zadrozny2002}, and Bayesian binning and averaging~ \citep{Naeini2015, Naeini2016}. As deep learning has become more dominant, however, various works have begun to directly target the calibration of deep networks. For example, \cite{Guo2017} have popularised a modern variant of Platt scaling known as \emph{temperature scaling}, which works by dividing a network's logits by a scalar $T > 0$ (learnt on a validation subset) prior to performing softmax. Temperature scaling has the desirable property that it can improve the calibration of a network without in any way affecting its accuracy. However, whilst its simplicity and effectiveness have made it a popular network calibration method, it does have downsides. For example, whilst it scales the logits to reduce the network's confidence in incorrect predictions, this also slightly reduces the network's confidence in predictions that were correct \citep{Kumar2018}. Moreover, it is known that temperature scaling does not calibrate a model under data distribution shift \citep{snoek2019can}.

By contrast, \cite{Kumar2018} initially eschew temperature scaling in favour of minimising a differentiable proxy for calibration error at training time, called Maximum Mean Calibration Error (MMCE), although they do later also use temperature scaling as a post-processing step to obtain better results than cross-entropy followed by temperature scaling~\citep{Guo2017}. Separately, \cite{muller2019does} propose training models on cross-entropy loss with label smoothing instead of one-hot labels, and show that label smoothing has a very favourable effect on model calibration.

In this paper, we propose a technique for improving network calibration that works by replacing the cross-entropy loss conventionally used when training classification networks with the focal loss proposed by \cite{Lin2017}. We observe that unlike cross-entropy, which minimises the KL divergence between the predicted (softmax) distribution and the target distribution (one-hot encoding in classification tasks) over classes, focal loss minimises a regularised KL divergence between these two distributions, which ensures minimisation of the KL divergence whilst \emph{increasing the entropy} of the predicted distribution, thereby preventing the model from becoming overconfident. Since focal loss, as shown in \S\ref{sec:focalloss}, is dependent on a hyperparameter, $\gamma$, that needs to be cross-validated, we also provide a method for choosing $\gamma$ automatically for each sample, and show that it outperforms all the baseline models.

The intuition behind using focal loss is to direct the network's attention during training towards samples for which it is currently predicting a low probability for the correct class, since trying to reduce the NLL on samples for which it is already predicting a high probability for the correct class is liable to lead to NLL overfitting, and thereby miscalibration \citep{Guo2017}. More formally, we show in \S\ref{sec:focalloss} that focal loss can be seen as \emph{implicitly} regularising the weights of the network during training by causing the gradient norms for confident samples to be lower than they would have been with cross-entropy, which we would expect to reduce overfitting and improve the network's calibration.

Overall, we make the following contributions:
\begin{enumerate}[leftmargin=*,topsep=0pt,itemsep=0pt,partopsep=0pt,parsep=0pt]
\item In \S\ref{sec:cause_cali}, we study the link that \cite{Guo2017} observed between miscalibration and NLL overfitting in detail, and show that the overfitting is associated with the predicted distributions for misclassified test samples becoming peakier as the optimiser tries to increase the magnitude of the network's weights to reduce the training NLL.
\item In \S\ref{sec:focalloss}, we propose the use of focal loss for training better-calibrated networks, and provide both theoretical and empirical justifications for this approach. In addition, we provide a principled method for automatically choosing $\gamma$ for each sample during training.
\item In \S\ref{sec:experiments}, we show, via experiments on a variety of classification datasets and network architectures, that DNNs trained with focal loss are more calibrated than those trained with cross-entropy loss (both with and without label smoothing), MMCE or Brier loss~\citep{brier1950verification}. Finally, we also make the interesting observation that whilst temperature scaling may not work for detecting out-of-distribution (OoD) samples, our approach can. We show that our approach is better at detecting out-of-distribution samples, taking CIFAR-10 as the in-distribution dataset, and SVHN and CIFAR-10-C as out-of-distribution datasets.
\end{enumerate}

\section{Problem Formulation}
\vspace{-2.5mm}
Let $D = \langle(\bm{\mathrm{x}}_i, y_i)\rangle_{i=1}^N$ denote a dataset consisting of $N$ samples from a joint distribution $\mathcal{D}(\mathcal{X}, \mathcal{Y})$, where for each sample $i$, $\mathbf{x}_i \in \mathcal{X}$ is the input and $y_i \in \mathcal{Y} = \{1, 2, ..., K\}$ is the ground-truth class label. Let $\hat{p}_{i,y} = f_\theta(y|\bm{\mathrm{x}}_i)$ be the probability that a neural network $f$ with model parameters $\theta$ predicts for a class $y$ on a given input $\bm{\mathrm{x}}_i$. The class that $f$ predicts for $\mathbf{x}_i$ is computed as $\hat{y}_i = \mathrm{argmax}_{y \in \mathcal{Y}} \; \hat{p}_{i,y}$, and the predicted confidence as $\hat{p}_i = \mathrm{max}_{y \in \mathcal{Y}} \; \hat{p}_{i,y}$. The network is said to be \emph{perfectly calibrated} when, for each sample $(\bm{\mathrm{x}}, y) \in D$,  the confidence $\hat{p}$ is equal to the model accuracy $\mathbb{P}(\hat{y} = y | \hat{p})$, i.e.\ the probability that the predicted class is correct.  For instance, of all the samples to which a perfectly calibrated neural network assigns a confidence of $0.8$, $80\%$ should be correctly predicted.

A popular metric used to measure model calibration is the \textit{expected calibration error} (ECE) \citep{Naeini2015}, defined as the expected absolute difference between the model's confidence and its accuracy, i.e.\ \( \mathbb{E}_{\hat{p}} \big[ \left| \mathbb{P}(\hat{y} = y | \hat{p}) - \hat{p} \right| \big] \). Since we only have finite samples, the ECE cannot in practice be computed using this definition. Instead, we divide the interval $[0,1]$ into $M$ equispaced bins, where the $i^{\mathrm{th}}$ bin is the interval $\left(\frac{i-1}{M}, \frac{i}{M} \right]$. Let $B_i$ denote the set of samples with confidences belonging to the $i^{\mathrm{th}}$ bin. The accuracy $A_i$ of this bin is computed as \(A_i = \frac{1}{|B_i|} \sum_{j \in B_i} \mathbbm{1} \left(\hat{y}_j = y_j\right) \), where $\mathbbm{1}$ is the indicator function, and $\hat{y}_j$ and $y_j$ are the predicted and ground-truth labels for the $j^{\mathrm{th}}$ sample. Similarly, the confidence $C_i$ of the $i^{\mathrm{th}}$ bin is computed as \(C_i = \frac{1}{|B_i|} \sum_{j \in B_i} \hat{p}_j \), i.e.\ $C_i$ is the average confidence of all samples in the bin. The ECE can be approximated as a weighted average of the absolute difference between the accuracy and confidence of each bin: $\mathrm{ECE} = \sum_{i=1}^{M} \frac{|B_i|}{N} \left| A_i - C_i \right|$.

A similar metric, the \textit{maximum calibration error} (MCE) \citep{Naeini2015}, is defined as the maximum absolute difference between the accuracy and confidence of each bin: $\mathrm{MCE} = \mathrm{max}_{i \in \{1, ..., M\}}\left|A_i - C_i\right|$.

\textbf{AdaECE:} One disadvantage of ECE is the uniform bin width. For a trained model, most of the samples lie within the highest confidence bins, and hence these bins dominate the value of the ECE. We thus also consider another metric, AdaECE (Adaptive ECE), for which bin sizes are calculated so as to evenly distribute samples between bins (similar to the adaptive binning procedure in \cite{Nguyen2015posterior}): $\mathrm{AdaECE} = \sum_{i=1}^{M} \frac{|B_i|}{N} \left| A_i - C_i \right| \text{ s.t.\ } \forall i, j \cdot |B_i| = |B_j|$.

\textbf{Classwise-ECE:} The ECE metric only considers the probability of the predicted class, without considering the other scores in the softmax distribution. A stronger definition of calibration would require the probabilities of all the classes in the softmax distribution to be calibrated \citep{Kull2019beyond, Vaicenavicius2019, Widmann2019calibration, Kumar2019verified}. This can be achieved with a simple classwise extension of the ECE metric: $\mathrm{Classwise ECE} = \frac{1}{K} \sum_{i=1}^{M}\sum_{j=1}^{K} \frac{|B_{i,j}|}{N} \left| A_{i,j} - C_{i,j} \right|$, where $K$ is the number of classes, \(B_{ij}\) denotes the set of samples from the $j^{th}$ class in the $i^{th}$ bin, \(A_{ij} = \frac{1}{|B_{ij}|} \sum_{k \in B_{ij}} \mathbbm{1} \left(j = y_k\right) \) and \(C_{i,j} = \frac{1}{|B_{ij}|} \sum_{k \in B_{ij}} \hat{p}_{kj} \).

\noindent A common way of visualising calibration is to use a \emph{reliability plot} \citep{Niculescu2005}, which plots the accuracies of the confidence bins as a bar chart (see Appendix Figure~\ref{fig:rel_conf_bin_plot}). For a perfectly calibrated model, the accuracy for each bin matches the confidence, and hence all of the bars lie on the diagonal. By contrast, if most of the bars lie above the diagonal, the model is more accurate than it expects, and is under-confident, and if most of the bars lie below the diagonal, then it is over-confident.

\vspace{-2.5mm}
\section{What Causes Miscalibration?}
\label{sec:cause_cali}
\vspace{-2.5mm}
We now discuss why high-capacity neural networks, despite achieving low classification errors on well-known datasets, tend to be miscalibrated. A key empirical observation made by \cite{Guo2017} was that poor calibration of such networks appears to be linked to overfitting on the negative log-likelihood (NLL) during training. In this section, we further inspect this observation to provide new insights.

For the analysis, we train a ResNet-50 network on CIFAR-10 with state-of-the-art performance settings~\citep{PyTorchCIFAR}. We use Stochastic Gradient Descent (SGD) with a mini-batch of size 128, momentum of 0.9, and learning rate schedule of $\{0.1, 0.01, 0.001\}$ for the first 150, next 100, and last 100 epochs, respectively. We minimise cross-entropy loss (a.k.a.\ NLL) $\mathcal{L}_c$, which, in a standard classification context, is $-\log \hat{p}_{i,y_i}$, where $\hat{p}_{i,y_i}$ is the probability assigned by the network to the correct class $y_i$ for the i$^{th}$ sample. Note that the NLL is minimised when for each training sample $i$, $\hat{p}_{i,y_i} = 1$, whereas the classification error is minimised when $\hat{p}_{i,y_i} > \hat{p}_{i,y}$ for all $y \neq y_i$. This indicates that even when the classification error is $0$, the NLL can be positive, and the optimisation algorithm can still try to reduce it to $0$ by further increasing the value of $\hat{p}_{i,y_i}$ for each sample (see Appendix~\ref{rel_plots_appendix}).

\begin{figure*}[!t]
	\centering
	\subfigure[]{\includegraphics[width=0.32\linewidth]{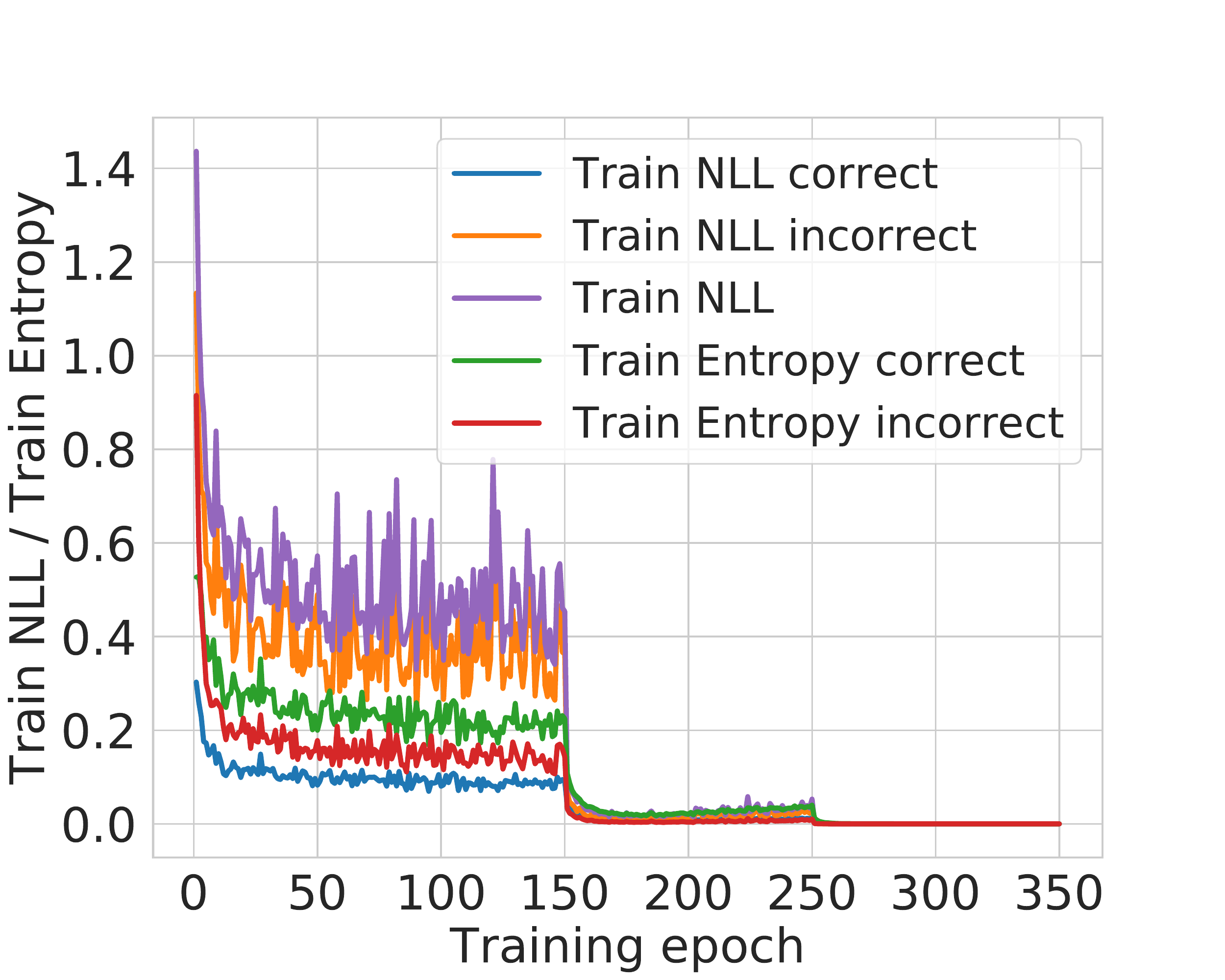}}
	\subfigure[]{\includegraphics[width=0.32\linewidth]{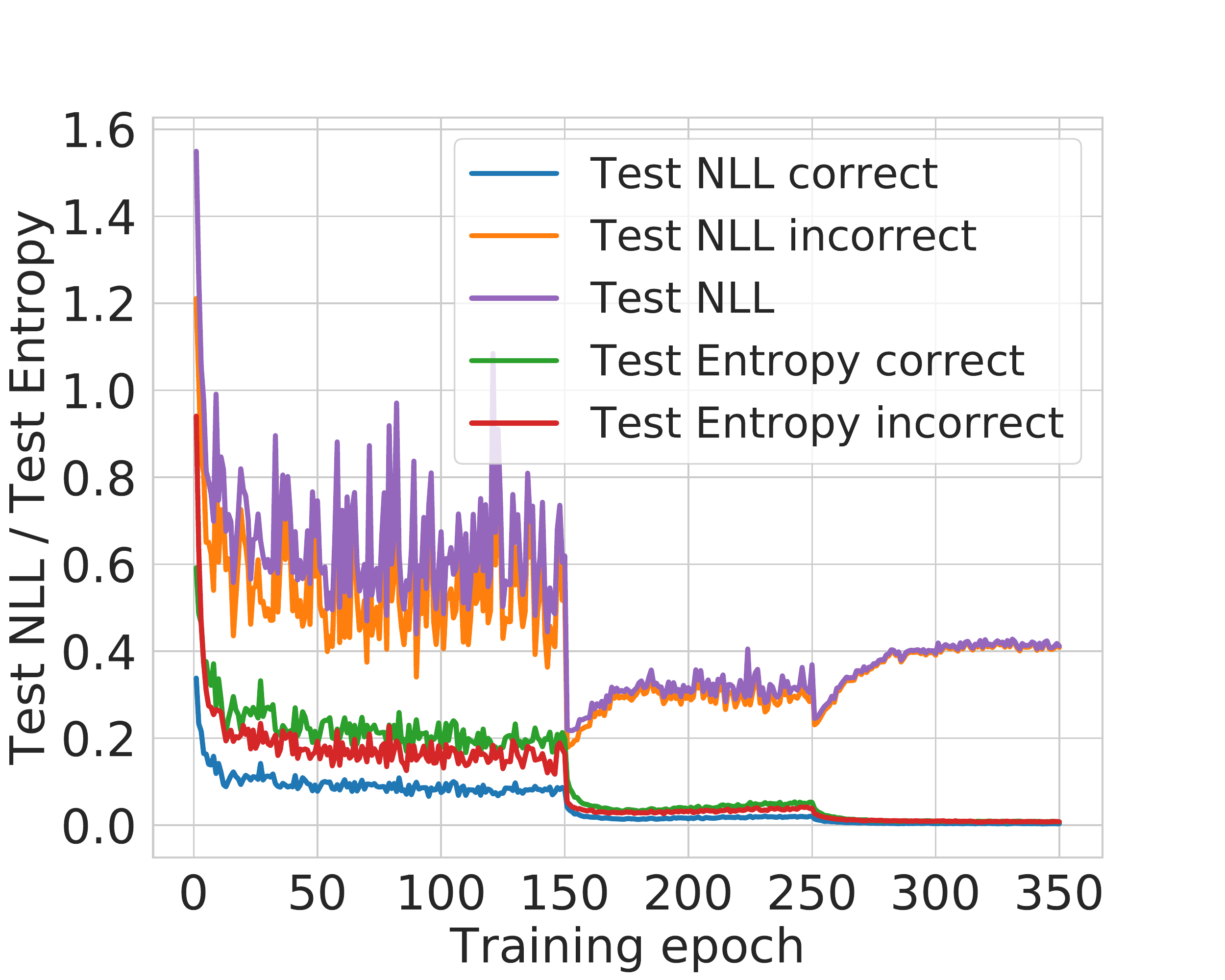}}
	\subfigure[]{\includegraphics[width=0.32\linewidth]{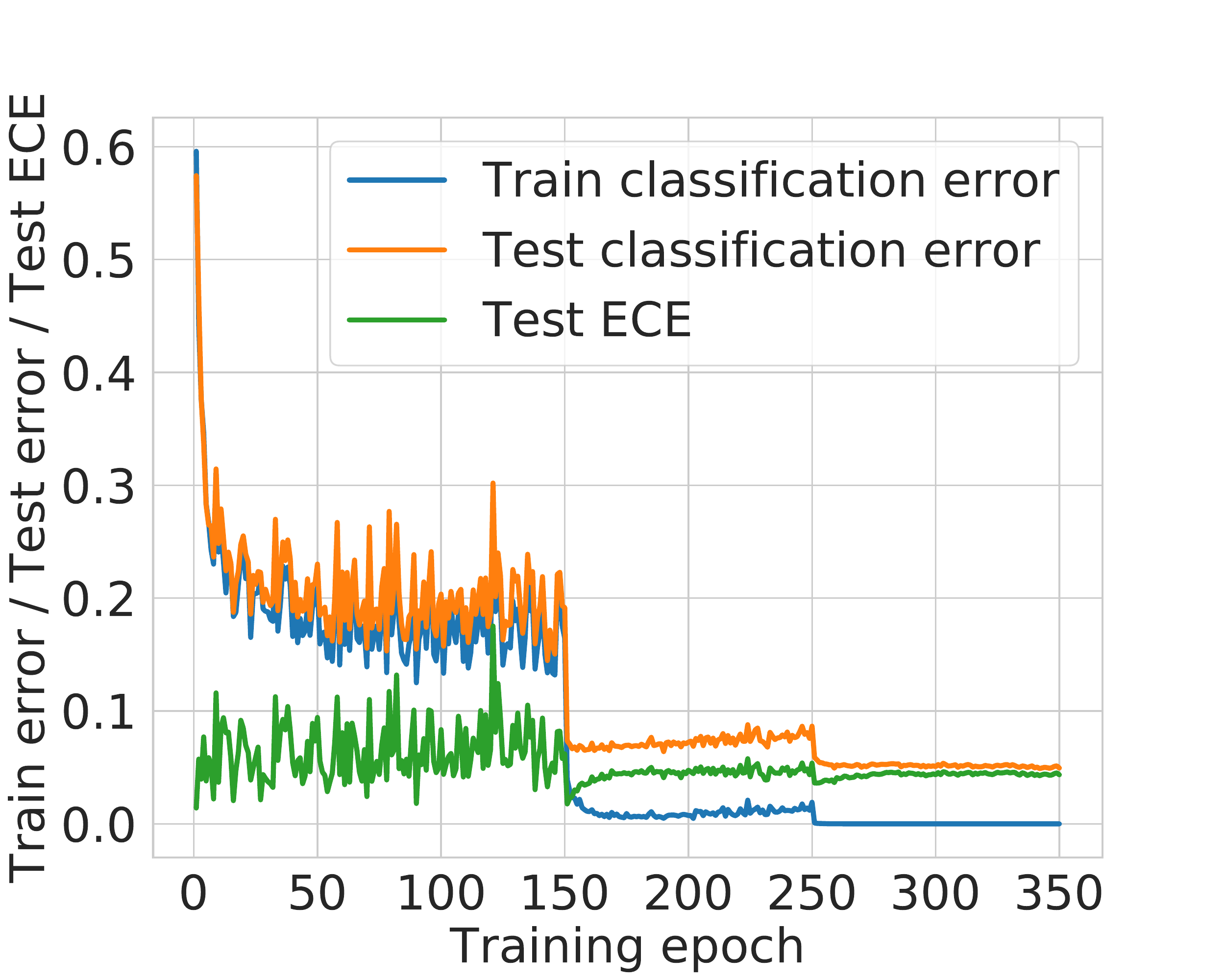}}
	\vspace{-4mm}
	\caption{Metrics related to calibration plotted whilst training a ResNet-50 network on CIFAR-10.}
	\label{fig:nll_entropy_ece}
	\vspace{-6mm}
\end{figure*}

To study how miscalibration occurs during training, we plot the average NLL for the train and test sets at each training epoch in Figures~\ref{fig:nll_entropy_ece}(a) and \ref{fig:nll_entropy_ece}(b). We also plot the average NLL and the entropy of the softmax distribution produced by the network for the correctly and incorrectly classified samples. In Figure \ref{fig:nll_entropy_ece}(c), we plot the classification errors on the train and test sets, along with the test set ECE.

\textbf{Curse of misclassified samples:} Figures \ref{fig:nll_entropy_ece}(a) and \ref{fig:nll_entropy_ece}(b) show that although the average train NLL (for both correctly and incorrectly classified training samples) broadly decreases throughout training, after the $150^{th}$ epoch (where the learning rate drops by a factor of $10$), there is a marked rise in the average test NLL, indicating that the network starts to overfit on average NLL. This increase in average test NLL is caused only by the incorrectly classified samples, as the average NLL for the correctly classified samples continues to decrease even after the $150^{th}$ epoch. We also observe that after epoch $150$, the test set ECE rises, indicating that the network is becoming miscalibrated. This corroborates the observation in \cite{Guo2017} that miscalibration and NLL overfitting are linked.

\textbf{Peak at the wrong place:} We further observe that the entropies of the softmax distributions for both the correctly and incorrectly classified {\em test} samples decrease throughout training (in other words, the distributions get peakier). This observation, coupled with the one we made above, indicates that {\em for the wrongly classified test samples, the network gradually becomes more and more confident about its incorrect predictions}.

\textbf{Weight magnification:} The increase in confidence of the network's predictions can happen if the network increases the norm of its weights $W$ to increase the magnitudes of the logits. In fact, cross-entropy loss is minimised when for each training sample $i$, $\hat{p}_{i,y_i} = 1$, which is possible only when $||W|| \to \infty$. Cross-entropy loss thus inherently induces this tendency of weight magnification in neural network optimisation. The promising performance of weight decay \citep{Guo2017} (regulating the norm of weights) on the calibration of neural networks can perhaps be explained using this. This increase in the network's confidence during training is one of the key causes of miscalibration.

\newcommand{\norm}[1]{\left\lVert#1\right\rVert}
\newcommand{\loss}{\mathcal{L}}
\newcommand{\bfw}{\mathbf{w}}
\vspace{-2.5mm}
\section{Improving Calibration using Focal Loss}
\label{sec:focalloss}
\vspace{-2.5mm}
As discussed in \S\ref{sec:cause_cali}, overfitting on NLL, which is observed as the network grows more confident on all of its predictions irrespective of their correctness, is strongly related to poor calibration. One cause of this is that the cross-entropy objective minimises the difference between the softmax distribution and the ground-truth one-hot encoding over an entire mini-batch, irrespective of how well a network classifies individual samples in the mini-batch. In this work, we study an alternative loss function, popularly known as \textit{focal loss} \citep{Lin2017}, that tackles this by weighting loss components generated from individual samples in a mini-batch by how well the model classifies them. For classification tasks where the target distribution is a one-hot encoding, it is defined as $\mathcal{L}_f = -(1 - \hat{p}_{i,y_i})^\gamma \log \hat{p}_{i,y_i}$, where $\gamma$ is a user-defined hyperparameter\footnote{We note in passing that unlike cross-entropy loss, focal loss in its general form is not a proper loss function, as minimising it does not always lead to the predicted distribution $\hat{p}$ being equal to the target distribution $q$ (see Appendix~\ref{reg_bregman} for the relevant definition and a longer discussion). However, when $q$ is a one-hot encoding (as in our case, and for most classification tasks), minimising focal loss does lead to $\hat{p}$ being equal to $q$.}.

\textbf{Why might focal loss improve calibration?} We know that cross-entropy forms an upper bound on the KL-divergence between the target distribution $q$ and the predicted distribution $\hat{p}$, i.e.\ $\mathcal{L}_c \geq \mathrm{KL}(q||\hat{p})$, so minimising cross-entropy results in minimising $\mathrm{KL}(q||\hat{p})$. Interestingly, a general form of focal loss can be shown to be an upper bound on the regularised KL-divergence, where the regulariser is the negative entropy of the predicted distribution $\hat{p}$, and the regularisation parameter is $\gamma$, the hyperparameter of focal loss (a proof of this can be found in Appendix~\ref{reg_bregman}):
\begin{equation}
\label{eq:reg_bregman}
    \mathcal{L}_f \geq \mathrm{KL}(q||\hat{p})- \gamma\mathbb{H}[\hat{p}].
\end{equation}
The most interesting property of this upper bound is that it shows that replacing cross-entropy with focal loss has the effect of adding a maximum-entropy regulariser \citep{Pereyra2017} to the implicit minimisation that was previously being performed. 
In other words, trying to minimise focal loss minimises the KL divergence between $\hat{p}$ and $q$, whilst simultaneously increasing the entropy of the predicted distribution $\hat{p}$. 
Note, in the case of ground truth with one-hot encoding, only the component of the entropy of $\hat{p}$ corresponding to the ground-truth index, $\gamma (-\hat{p}_{i,y_i} \log \hat{p}_{i,y_i})$, will be maximised (refer~Appendix~\ref{reg_bregman}). 
Encouraging the predicted distribution to have higher entropy can help avoid the overconfident predictions produced by DNNs (see the `Peak at the wrong place' paragraph of \S\ref{sec:cause_cali}), and thereby improve calibration.

\begin{figure*}[!t]
	\centering
	\subfigure[\vspace{-2mm}]{\includegraphics[width=0.19\linewidth]{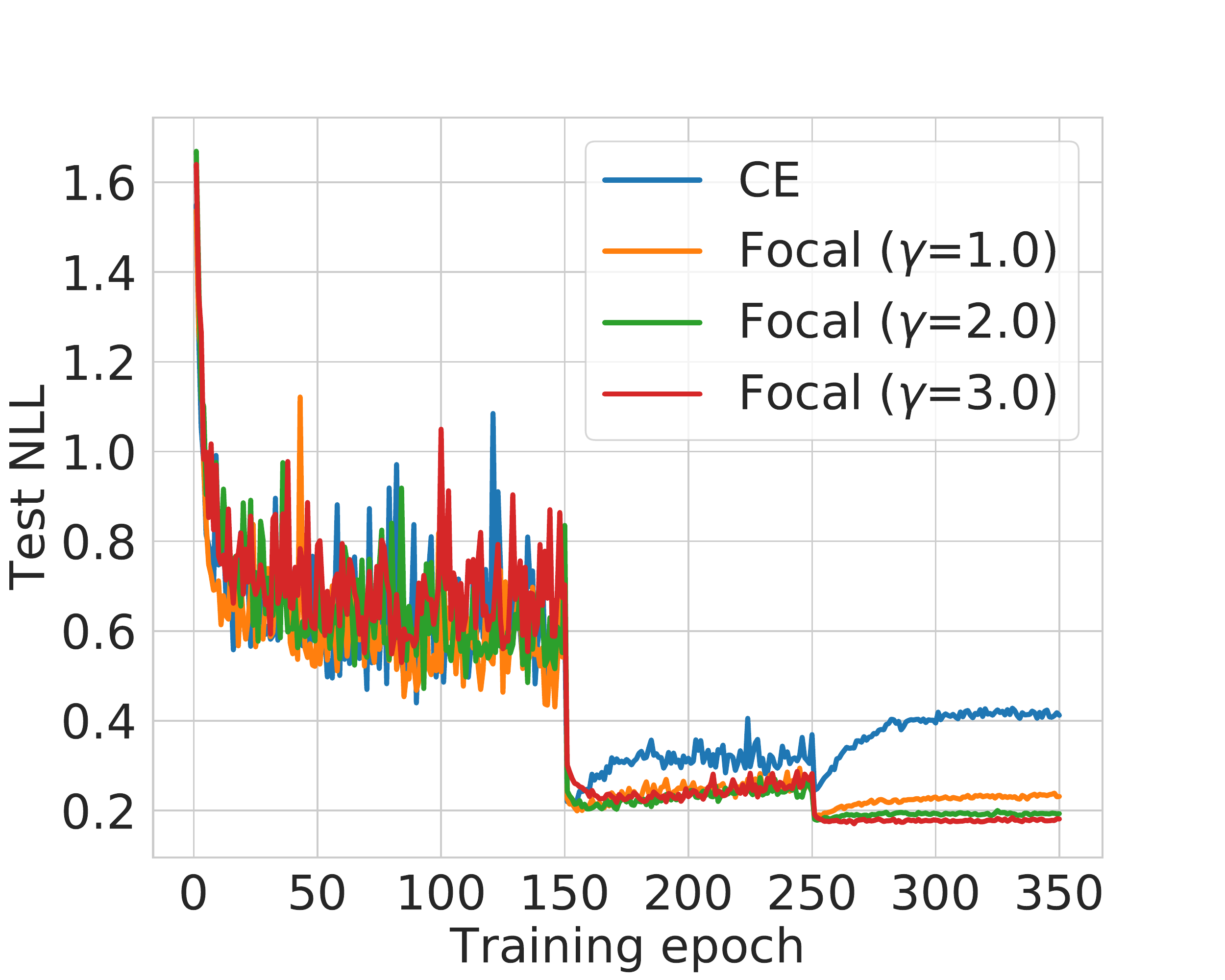}}
	\subfigure[\vspace{-2mm}]{\includegraphics[width=0.19\linewidth]{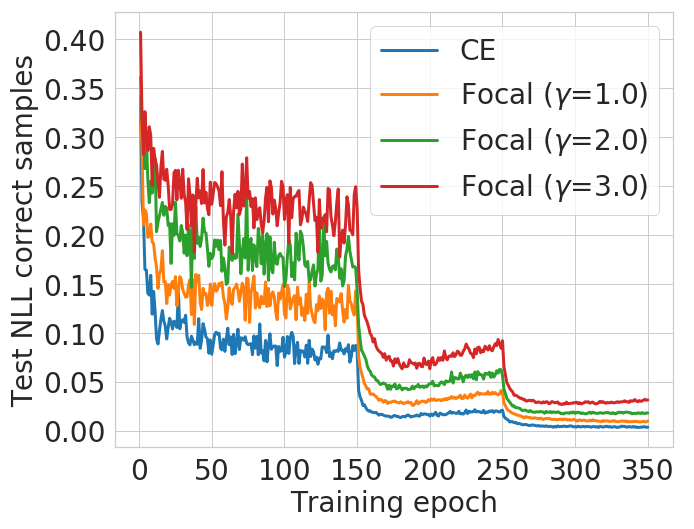}}
	\subfigure[\vspace{-2mm}]{\includegraphics[width=0.19\linewidth]{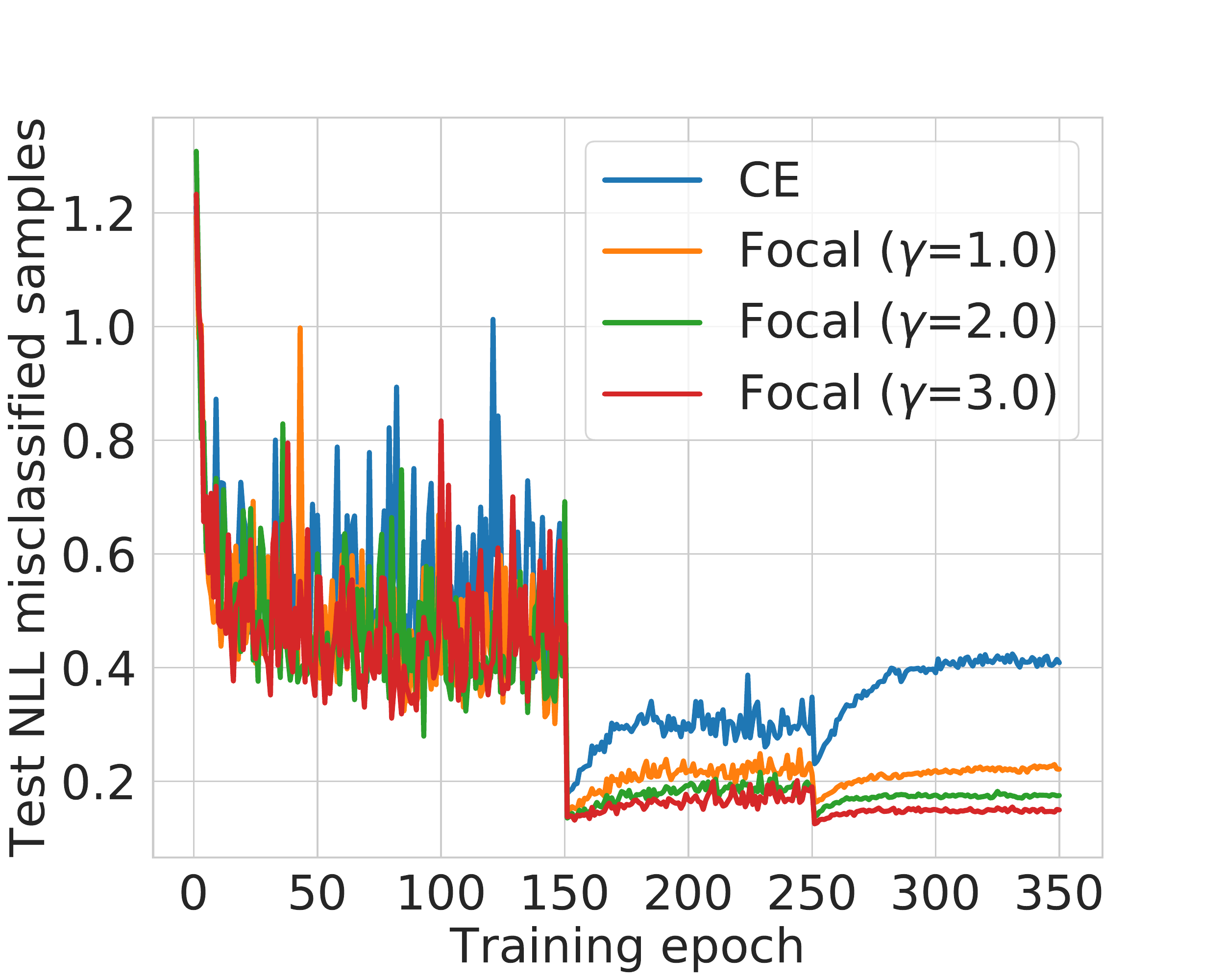}}
	\subfigure[\vspace{-2mm}]{\includegraphics[width=0.19\linewidth]{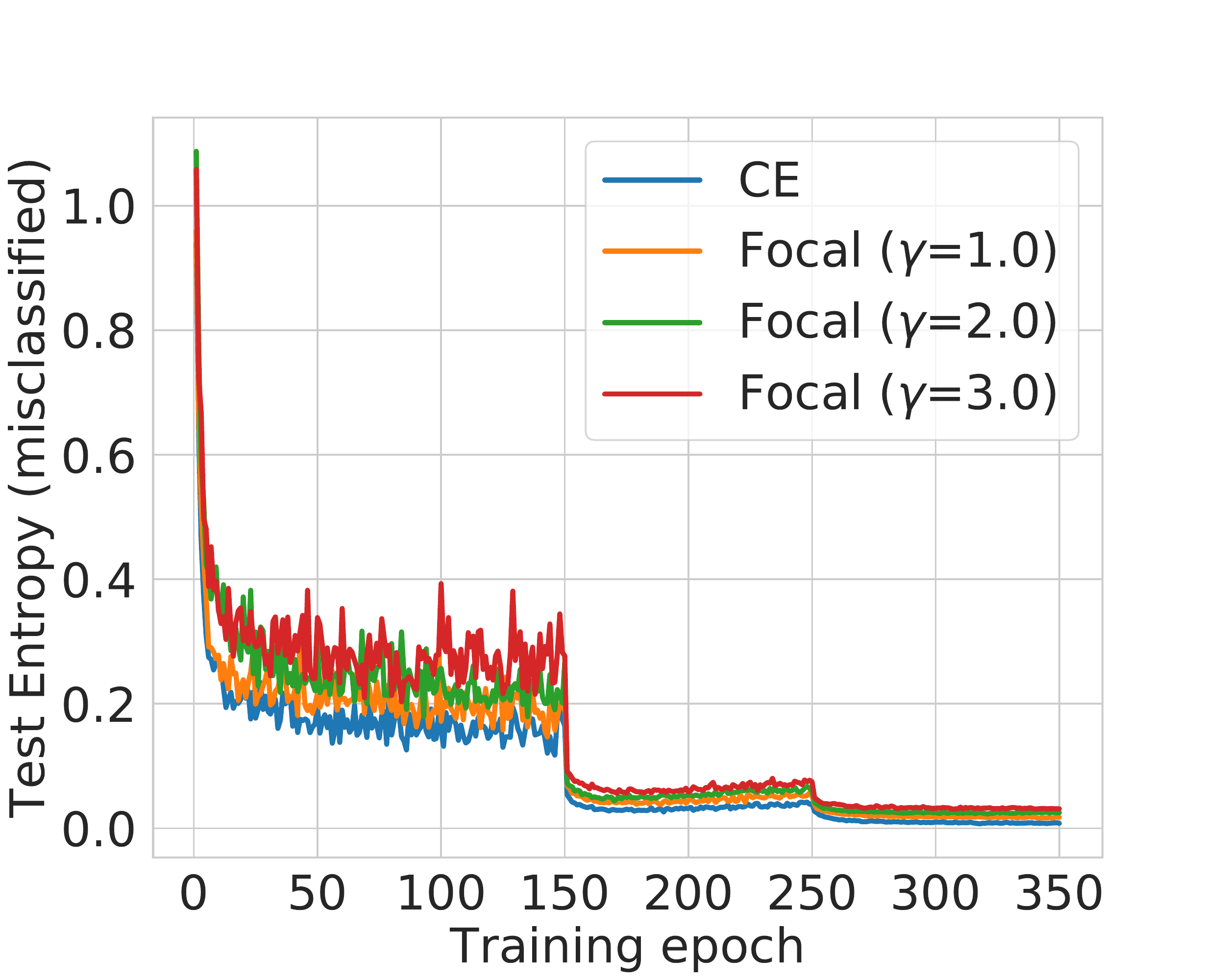}}
	\subfigure[\vspace{-2mm}]{\includegraphics[width=0.19\linewidth]{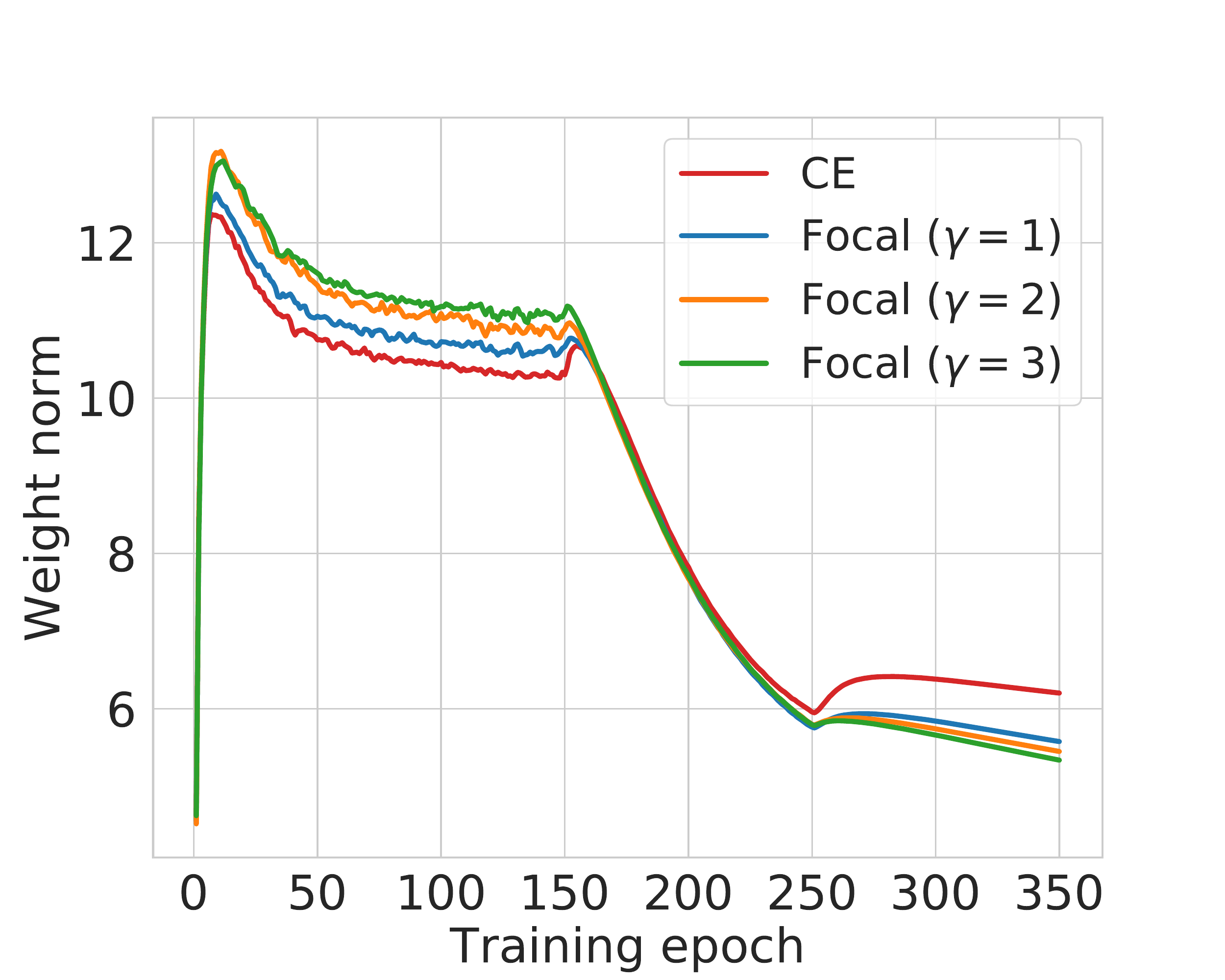}}
	\vspace{-\baselineskip}
	\caption{How metrics related to model calibration change whilst training several ResNet-50 networks on CIFAR-10, using either cross-entropy loss, or focal loss with $\gamma$ set to 1, 2 or 3.}
	\vspace{-\baselineskip}
	\label{fig:nll_corr_incorr_entropy}
\end{figure*}

\textbf{Empirical observations:} To analyse the behaviour of neural networks trained on focal loss, we use the same framework as mentioned above, and train four ResNet-50 networks on CIFAR-10, one using cross-entropy loss, and three using focal loss with $\gamma = 1, 2$ and $3$. Figure \ref{fig:nll_corr_incorr_entropy}(a) shows that the test NLL for the cross-entropy model significantly increases towards the end of training (before saturating), whereas the NLLs for the focal loss models remain low. To better understand this, we analyse the behaviour of these models for correctly and incorrectly classified samples. Figure~\ref{fig:nll_corr_incorr_entropy}(b) shows that even though the NLLs for the correctly classified samples broadly-speaking decrease over the course of training for all the models, the NLLs for the focal loss models remain consistently higher than that for the cross-entropy model throughout training, implying that the focal loss models are relatively less confident than the cross-entropy model for samples that they predict correctly. This is important, as we have already discussed that it is overconfidence that normally makes deep neural networks miscalibrated. Figure \ref{fig:nll_corr_incorr_entropy}(c) shows that in contrast to the cross-entropy model, for which the NLL for misclassified test samples increases significantly after epoch $150$, the rise in this value for the focal loss models is much less severe. Additionally, in Figure \ref{fig:nll_corr_incorr_entropy}(d), we notice that the entropy of the softmax distribution for misclassified test samples is consistently (if marginally) higher for focal loss than for cross-entropy (consistent with Equation~\ref{eq:reg_bregman}).

Note that from Figure \ref{fig:nll_corr_incorr_entropy}(a), one may think that applying early stopping when training a model on cross-entropy can provide better calibration scores. However, there is no ideal way of doing early stopping that provides the best calibration error and the best test set accuracy. For fair comparison, we chose $3$ intermediate models for each loss function with the best val set ECE, NLL and accuracy, and observed that: a) for every stopping criterion, focal loss outperforms cross-entropy in both test set accuracy and ECE, b) when using val set ECE as a stopping criterion, the intermediate model for cross-entropy indeed improves its test set ECE, but at the cost of a significantly higher test error. Please refer to Appendix~\ref{sec:early_stopping} for more details.

As per \S\ref{sec:cause_cali}, an increase in the test NLL and a decrease in the test entropy for misclassified samples, along with no corresponding increase in the test NLL for the correctly classified samples, can be interpreted as the network starting to predict softmax distributions for the misclassified samples that are ever more peaky in the wrong place. Notably, our results in Figures~\ref{fig:nll_corr_incorr_entropy}(b), \ref{fig:nll_corr_incorr_entropy}(c) and \ref{fig:nll_corr_incorr_entropy}(d) clearly show that this effect is significantly reduced when training with focal loss rather than cross-entropy, leading to a better-calibrated network whose predictions are less peaky in the wrong place.

\begin{figure*}[!t]
    \centering
    \subfigure[]{\includegraphics[width=0.24\linewidth]{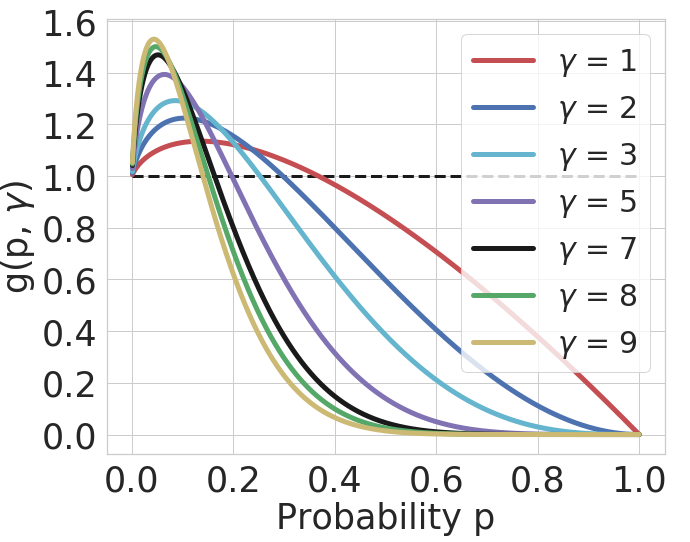}}
    \subfigure[Epoch 10]{\includegraphics[width=0.24\linewidth]{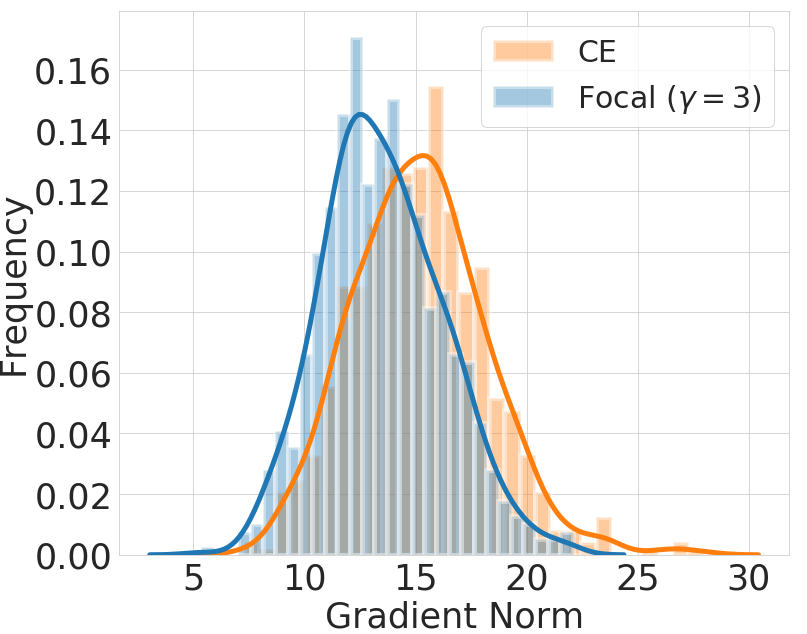}}
    \subfigure[Epoch 100]{\includegraphics[width=0.24\linewidth]{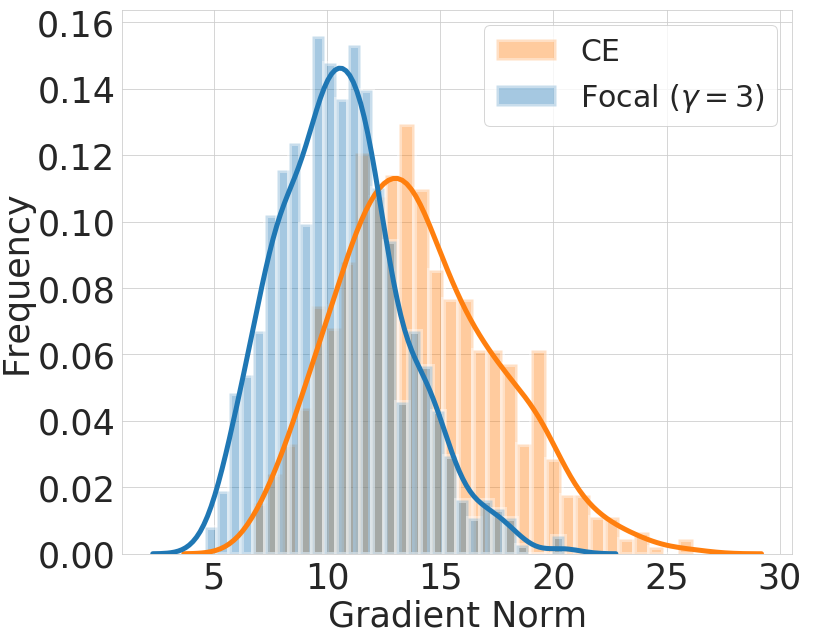}}
    \subfigure[Epoch 200]{\includegraphics[width=0.24\linewidth]{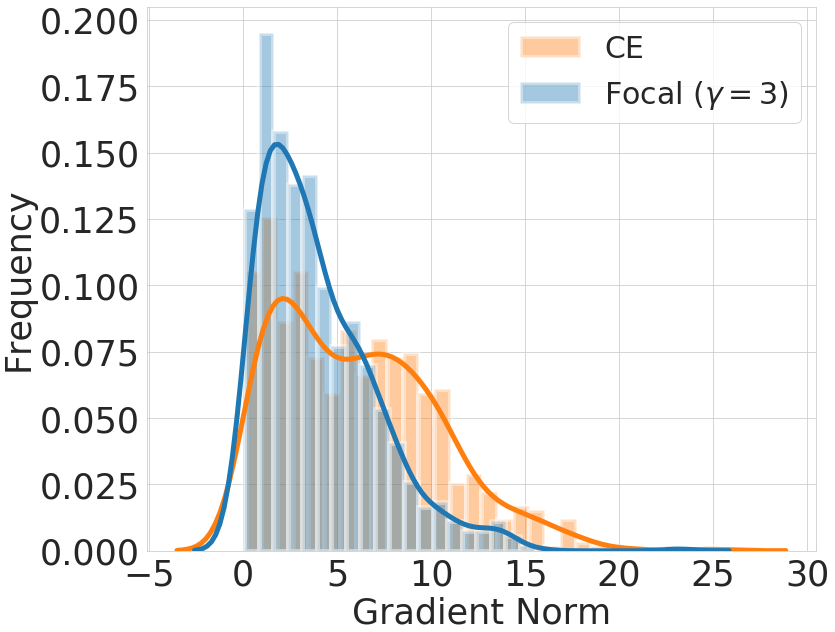}}
    \vspace{-\baselineskip}
	\caption{(a): $g(p, \gamma)$ vs.\ $p$ and (b-d): histograms of the gradient norms of the last linear layer for both cross-entropy and focal loss.}
  \label{fig:g_pt_grad_norms}
  \vspace{-5mm}
\end{figure*}

\textbf{Theoretical justification:} As mentioned previously, once a model trained using cross-entropy reaches high training accuracy, the optimiser may try to further reduce the training NLL by increasing the confidences for the correctly classified samples. It may achieve this by magnifying the network weights to increase the magnitudes of the logits. To verify this hypothesis, we plot the $L_2$ norm of the weights of the last linear layer for all four networks as a function of the training epoch (see Figure \ref{fig:nll_corr_incorr_entropy}(e)). Notably, although the norms of the weights for the models trained on focal loss are initially higher than that for the cross-entropy model, \textit{a complete reversal} in the ordering of the weight norms occurs between epochs $150$ and $250$. In other words, as the networks start to become miscalibrated, the weight norm for the cross-entropy model also starts to become greater than those for the focal loss models. In practice, this is because focal loss, by design, starts to act as a regulariser on the network's weights once the model has gained a certain amount of confidence in its predictions. This behaviour of focal loss can be observed even on a much simpler setup like a linear model (see Appendix~\ref{linear_model}). To better understand this, we start by considering the following proposition (proof in Appendix~\ref{sec:proof}):
\begin{pro}
\label{pro1}
For focal loss $\loss_f$ and cross-entropy $\loss_c$, the gradients $\frac{\partial \loss_f}{\partial \bfw} = \frac{\partial \loss_c}{\partial \bfw} g(\hat{p}_{i,y_i}, \gamma)$, where $g(p, \gamma) = (1-p)^\gamma - \gamma p (1-p)^{\gamma - 1} \log(p)$, $\gamma \in \mathbb{R}^+$ is the focal loss hyperparameter, and $\bfw$ denotes the parameters of the last linear layer. Thus $\norm{\frac{\partial  \loss_f}{\partial \bfw}} \leq \norm{\frac{\partial  \loss_c}{\partial \bfw}}$ if $g(\hat{p}_{i,y_i}, \gamma) \in [0, 1]$.
\end{pro}
\vspace{-.5\baselineskip}
Proposition~\ref{pro1} shows the relationship between the norms of the gradients of the last linear layer for focal loss and cross-entropy loss, for the same network architecture. Note that this relation depends on a function $g(p, \gamma)$, which we plot in Figure~\ref{fig:g_pt_grad_norms}(a) to understand its behaviour. It is clear that for every $\gamma$, there exists a (different) threshold $p_0$ such that for all $p \in [0,p_0]$, $g(p,\gamma) \ge 1$, and for all $p \in (p_0, 1]$, $g(p,\gamma) < 1$. (For example, for $\gamma = 1$, $p_0 \approx 0.4$.) We use this insight to further explain why focal loss provides implicit weight regularisation.

\textbf{Implicit weight regularisation:} For a network trained using focal loss with a fixed $\gamma$, during the initial stages of the training, when $\hat{p}_{i,y_i} \in (0,p_0)$, $g(\hat{p}_{i,y_i}, \gamma) > 1$. This implies that the confidences of the focal loss model's predictions will initially increase faster than they would for cross-entropy. However, as soon as $\hat{p}_{i,y_i}$ crosses the threshold $p_0$, $g(\hat{p}_{i,y_i}, \gamma)$ falls below $1$ and reduces the size of the gradient updates made to the network weights, thereby having a regularising effect on the weights. This is why, in Figure \ref{fig:nll_corr_incorr_entropy}(e), we find that the weight norms of the models trained with focal loss are initially higher than that for the model trained using cross-entropy. However, as training progresses, we find that the ordering of the weight norms reverses, as focal loss starts regularising the network weights. Moreover, we can draw similar insights from Figures~\ref{fig:g_pt_grad_norms}(b), \ref{fig:g_pt_grad_norms}(c) and \ref{fig:g_pt_grad_norms}(d), in which we plot histograms of the gradient norms of the last linear layer (over all samples in the training set) at epochs $10$, $100$ and $200$, respectively. At epoch $10$, the gradient norms for cross-entropy and focal loss are similar, but as training progresses, those for cross-entropy decrease less rapidly than those for focal loss, indicating that the gradient norms for focal loss are consistently lower than those for cross-entropy throughout training.

Finally, observe in Figure~\ref{fig:g_pt_grad_norms}(a) that for higher $\gamma$ values, the fall in $g(p,\gamma)$ is steeper. We would thus expect a greater weight regularisation effect for models that use higher values of $\gamma$. This explains why, of the three models that we trained using focal loss, the one with $\gamma = 3$ outperforms (in terms of calibration) the one with $\gamma = 2$, which in turn outperforms the model with $\gamma = 1$. Based on this observation, one might think that, in general, a higher value of gamma would lead to a more calibrated model. However, this is not the case, as we notice from Figure~\ref{fig:g_pt_grad_norms}(a) that for $\gamma \ge 7$, $g(p,\gamma)$ reduces to nearly $0$ for a relatively low value of $p$ (around $0.5$). As a result, using values of $\gamma$ that are too high will cause the gradients to die (i.e.\ reduce to nearly $0$) early, at a point at which the network's predictions remain ambiguous, thereby causing the training process to fail. 

\textbf{How to choose $\gamma$:} As discussed, focal loss provides implicit entropy and weight regularisation, which heavily depend on the value of $\gamma$. Finding an appropriate $\gamma$ is normally done using cross-validation. Also, traditionally, $\gamma$ is fixed for all samples in the dataset. However, as shown, the regularisation effect for a sample $i$ depends on $\hat{p}_{i,y_i}$, i.e.\ the predicted probability for the ground truth label for the sample. It thus makes sense to choose $\gamma$ for each sample based on the value of $\hat{p}_{i,y_i}$. To this end, we provide Proposition~\ref{pro:gamma} (proof in Appendix~\ref{sec:proof}), which we use to find a solution to this problem:
\begin{pro}
	\label{pro:gamma}
	Given a $p_0$, for $1 \geq p \geq p_0 > 0$, $g(p, \gamma) \leq 1$ for all $\gamma \geq \gamma^* = \frac{a}{b} + \frac{1}{\log a}W_{-1} \big(-\frac{a^{(1-a/b)}}{b} \log a \big)$, where $a = 1-p_0$, $b = p_0 \log p_0$, and $W_{-1}$ is the Lambert-W function~\citep{corless1996lambertw}. Moreover, for $p \geq p_0 > 0$ and $\gamma \geq \gamma^*$, the equality $g(p, \gamma) = 1$ holds only for $p = p_0$ and $\gamma = \gamma^*$.
\end{pro}
It is worth noting that there exist multiple values of $\gamma$ where $g(p, \gamma) \leq 1$ for all $p \geq p_0$. For a given $p_0$, Proposition~\ref{pro:gamma} allows us to compute $\gamma$ s.t.\ (i) $g(p_0,\gamma) = 1$; (ii) $g(p, \gamma) \ge 1$ for $p \in [0,p_0)$; and (iii) $g(p, \gamma) < 1$ for $p \in (p_0,1]$. This allows us to control the magnitude of the gradients for a particular sample $i$ based on the current value of $\hat{p}_{i,y_i}$, and gives us a way of obtaining an informed value of $\gamma$ for each sample. For instance, a reasonable policy might be to choose $\gamma$ s.t.\ $g(\hat{p}_{i,y_i}, \gamma) > 1$ if $\hat{p}_{i,y_i}$ is small (say less than $0.25$), and \ $g(\hat{p}_{i,y_i}, \gamma) < 1$ otherwise. Such a policy will have the effect of making the weight updates larger for samples having a low predicted probability for the correct class and smaller for samples with a relatively higher predicted probability for the correct class.

Following the aforementioned arguments, we choose a threshold $p_0$ of $0.25$, and use Proposition~\ref{pro:gamma} to obtain a $\gamma$ policy such that $g(p, \gamma)$ is observably greater than $1$ for $p \in [0, 0.25)$ and $g(p, \gamma) < 1$ for $p \in (0.25, 1]$. In particular, we use the following schedule: if $\hat{p}_{i,y_i} \in [0,0.2)$, then $\gamma = 5$, otherwise $\gamma = 3$ (note that $g(0.2, 5) \approx 1$ and $g(0.25, 3) \approx 1$: see Figure~\ref{fig:g_pt_grad_norms}(a)).
We find this $\gamma$ policy to perform consistently well across multiple classification datasets and network architectures. Having said that, one can calculate multiple such schedules for $\gamma$ following Proposition~\ref{pro:gamma}, using the intuition of having a relatively high $\gamma$ for low values of $\hat{p}_{i, y_i}$ and a relatively low $\gamma$ for high values of $\hat{p}_{i, y_i}$.

\vspace{-2.5mm}
\section{Experiments}
\label{sec:experiments}
\vspace{-3mm}
We conduct image and document classification experiments to test the performance of focal loss. For the former, we use CIFAR-10/100 \citep{Krizhevsky2009} and Tiny-ImageNet \citep{deng2009imagenet} , and train ResNet-50, ResNet-110 \citep{He2016}, Wide-ResNet-26-10 \citep{Zagoruyko2016} and DenseNet-121 \citep{Huang2017} models, and for the latter, we use 20 Newsgroups \citep{Lang1995} and Stanford Sentiment Treebank (SST) \citep{Socher2013} datasets and train Global Pooling CNN \citep{Lin2013} and Tree-LSTM~\citep{Tai2015} models. Further details on the datasets and training can be found in Appendix~\ref{dataset}.

\textbf{Baselines} Along with cross-entropy loss, we compare our method against the following baselines: a) \emph{MMCE} (Maximum Mean Calibration Error) \citep{Kumar2018}, a continuous and differentiable proxy for calibration error that is normally used as a regulariser alongside cross-entropy, b) \emph{Brier loss}~\citep{brier1950verification}, the squared error between the predicted softmax vector and the one-hot ground truth encoding (Brier loss is an important baseline as it can be decomposed into calibration and refinement~\citep{degroot1983comparison}), and c) \emph{Label smoothing}~\citep{muller2019does} (LS): given a one-hot ground-truth distribution $\bm{\mathrm{q}}$ and a smoothing factor $\alpha$ (hyperparameter), the smoothed vector $\bm{\mathrm{s}}$ is obtained as $\bm{\mathrm{s}}_i = (1-\alpha)\bm{\mathrm{q}}_i + \alpha(1-\bm{\mathrm{q}}_i)/(K-1)$, where $\bm{\mathrm{s}}_i$ and $\bm{\mathrm{q}}_i$ denote the $i^{th}$ elements of $\bm{\mathrm{s}}$ and $\bm{\mathrm{q}}$ respectively, and $K$ is the number of classes. Instead of $\bm{\mathrm{q}}$, $\bm{\mathrm{s}}$ is treated as the ground truth. We train models using $\alpha = 0.05$ and $\alpha = 0.1$, but find $\alpha = 0.05$ to perform better. We thus report the results obtained from LS-$0.05$ with $\alpha = 0.05$.

\textbf{Focal Loss}: As mentioned in \S\ref{sec:focalloss}, our proposed approach is the sample-dependent schedule FLSD-$53$ ($\gamma=5$ for $\hat{p}_y \in [0, 0.2)$, and $\gamma=3$ for $\hat{p}_y \in [0.2, 1]$), which we find to perform well across most classification datasets and network architectures. In addition, we also train other focal loss baselines, including ones with $\gamma$ fixed to $1,2$ and $3$, and also ones that have a training epoch-dependent schedule for $\gamma$. Among the focal loss models trained with a fixed $\gamma$, using validation set we find $\gamma = 3$ (FL-3) to perform the best. Details of all these approaches can be found in Appendix~\ref{results}. 

\begin{figure*}[!t]
    \centering
    \includegraphics[width=\linewidth]{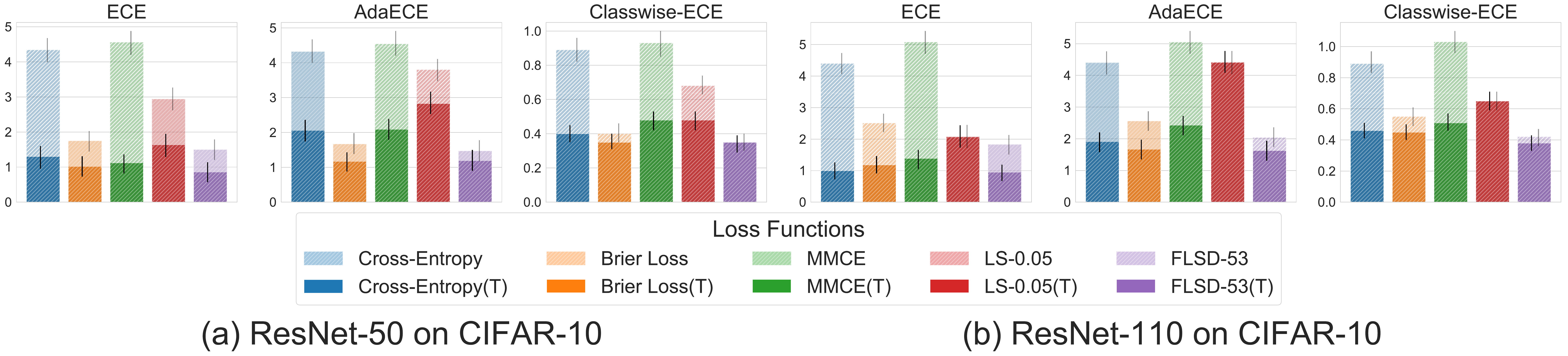}
    \vspace{-\baselineskip}
	\caption{Bar plots with confidence intervals for ECE, AdaECE and Classwise-ECE, computed for ResNet-50 (first $3$ figures) and ResNet-110 (last $3$ figures) on CIFAR-10.}
  \label{fig:error_ba}
  \vspace{-1\baselineskip}
\end{figure*}

\begin{table*}[!t]
	\centering
	\scriptsize
	\resizebox{\linewidth}{!}{%
		\begin{tabular}{cccccccccccccc}
			\toprule
			\textbf{Dataset} & \textbf{Model} & \multicolumn{2}{c}{\textbf{Cross-Entropy}} &
			\multicolumn{2}{c}{\textbf{Brier Loss}} & \multicolumn{2}{c}{\textbf{MMCE}} &
			\multicolumn{2}{c}{\textbf{LS-0.05}} & \multicolumn{2}{c}{\textbf{FL-3 (Ours)}} &
			\multicolumn{2}{c}{\textbf{FLSD-53 (Ours)}} \\
			&& Pre T & Post T & Pre T & Post T & Pre T & Post T & Pre T & Post T & Pre T & Post T & Pre T & Post T \\
			\midrule
			
			\multirow{4}{*}{CIFAR-100} & ResNet-50&17.52&3.42(2.1)&6.52&3.64(1.1)&15.32&2.38(1.8)&7.81&4.01(1.1)&\tikzmark{top left}5.13&\textbf{1.97(1.1)}&\textbf{4.5}&2.0(1.1)\\
			& ResNet-110&19.05&4.43(2.3)&\textbf{7.88}&4.65(1.2)&19.14&\textbf{3.86(2.3)}&11.02&5.89(1.1)&8.64&3.95(1.2)&8.56&4.12(1.2)\\
			& Wide-ResNet-26-10&15.33&2.88(2.2)&4.31&2.7(1.1)&13.17&4.37(1.9)&4.84&4.84(1)&\textbf{2.13}&2.13(1)&3.03&\textbf{1.64(1.1)}\\
			& DenseNet-121&20.98&4.27(2.3)&5.17&2.29(1.1)&19.13&3.06(2.1)&12.89&7.52(1.2)&4.15&\textbf{1.25(1.1)}&\textbf{3.73}&1.31(1.1)\\
			\midrule
			\multirow{4}{*}{CIFAR-10} & ResNet-50&4.35&1.35(2.5)&1.82&1.08(1.1)&4.56&1.19(2.6)&2.96&1.67(0.9)&\textbf{1.48}&1.42(1.1)&1.55&\textbf{0.95(1.1)}\\
			& ResNet-110&4.41&1.09(2.8)&2.56&1.25(1.2)&5.08&1.42(2.8)&2.09&2.09(1)&\textbf{1.55}&\textbf{1.02(1.1)}&1.87&1.07(1.1)\\
			& Wide-ResNet-26-10&3.23&0.92(2.2)&\textbf{1.25}&1.25(1)&3.29&0.86(2.2)&4.26&1.84(0.8)&1.69&0.97(0.9)&1.56&\textbf{0.84(0.9)}\\
			& DenseNet-121&4.52&1.31(2.4)&1.53&1.53(1)&5.1&1.61(2.5)&1.88&1.82(0.9)&1.32&1.26(0.9)&\textbf{1.22}&\textbf{1.22(1)}\\
			\midrule
			Tiny-ImageNet & ResNet-50&15.32&5.48(1.4)&4.44&4.13(0.9)&13.01&5.55(1.3)&15.23&6.51(0.7)&1.87&1.87(1)&\textbf{1.76}&\textbf{1.76(1)}\\
			\midrule
			20 Newsgroups & Global Pooling CNN&17.92&2.39(3.4)&13.58&3.22(2.3)&15.48&6.78(2.2)&\textbf{4.79}&2.54(1.1)&8.67&3.51(1.5)&6.92&\textbf{2.19(1.5)}\\
			\midrule
			SST Binary & Tree-LSTM&7.37&2.62(1.8)&9.01&2.79(2.5)&5.03&4.02(1.5)&\textbf{4.84}&4.11(1.2)&16.05&1.78(0.5)&9.19&\textbf{1.83(0.7)}\tikzmark{bottom right}\\
			\bottomrule
		\end{tabular}%
	}
	\caption{ECE $(\%)$ computed for different approaches both pre and post temperature scaling (cross-validating T on ECE). Optimal temperature for each method is indicated in brackets. $T\approx 1$ indicates innately calibrated model. \vspace{-3mm}}
	\label{table:ece_tab1}
\end{table*}

\begin{table*}[!t]
\centering
\scriptsize
\resizebox{\linewidth}{!}{%
\begin{tabular}{cccccccc}
\toprule
\textbf{Dataset} & \textbf{Model} & \textbf{Cross-Entropy} &
\textbf{Brier Loss} & \textbf{MMCE} & \textbf{LS-0.05} & \textbf{FL-3 (Ours)} & \textbf{FLSD-53 (Ours)} \\

\midrule
\multirow{4}{*}{CIFAR-100} & ResNet-50&23.3&23.39&23.2&23.43&22.75&23.22\\
& ResNet-110&22.73&25.1&23.07&23.43&22.92&22.51\\
& Wide-ResNet-26-10&20.7&20.59&20.73&21.19&19.69&20.11\\
& DenseNet-121&24.52&23.75&24.0&24.05&23.25&22.67\\
\midrule
\multirow{4}{*}{CIFAR-10} & ResNet-50&4.95&5.0&4.99&5.29&5.25&4.98\\
& ResNet-110&4.89&5.48&5.4&5.52&5.08&5.42\\
& Wide-ResNet-26-10&3.86&4.08&3.91&4.2&4.13&4.01\\
& DenseNet-121&5.0&5.11&5.41&5.09&5.33&5.46\\
\midrule
Tiny-ImageNet & ResNet-50&49.81&53.2&51.31&47.12&49.69&49.06\\
\midrule
20 Newsgroups & Global Pooling CNN&26.68&27.06&27.23&26.03&29.26&27.98\\
\midrule
SST Binary & Tree-LSTM&12.85&12.85&11.86&13.23&12.19&12.8\\
\bottomrule
\end{tabular}}
\caption{Test set error $(\%)$ computed for different approaches. \vspace{-3mm}}
\label{table:error_tab1}
\end{table*}

\textbf{Temperature Scaling:} In order to compute the optimal temperature, we use two different methods: (a) learning the temperature by minimising val set NLL, and (b) performing grid search over temperatures between 0 and 10, with a step of 0.1, and finding the one that minimises val set ECE. We find the second approach to produce {\em stronger baselines} and report results obtained using this approach.

\textbf{Performance Gains:} We report ECE$\%$ (computed using 15 bins) along with optimal temperatures in Table \ref{table:ece_tab1}, and test set error in Table~\ref{table:error_tab1}. We report the other calibration scores (AdaECE, Classwise-ECE, MCE and NLL) in Appendix~\ref{results}. Firstly, for all dataset-network pairs, we obtain very competitive classification accuracies (shown in Table~\ref{table:error_tab1}).  Secondly, {\em it is clear from Table~\ref{table:ece_tab1}, and Tables~\ref{table:ada_ece_tab1} and~\ref{table:sce_tab1} in the appendix, that focal loss with sample-dependent $\gamma$ and with $\gamma = 3$ outperform all the baselines: cross-entropy, label smoothing, Brier loss and MMCE.} They broadly produce the lowest calibration errors {\em both before and after temperature scaling}. This observation is particularly encouraging, as it also indicates that a principled method of obtaining values of $\gamma$ for focal loss can produce a very calibrated model, with no need to use validation set for tuning $\gamma$. As shown in Figure~\ref{fig:error_ba}, we also compute $90\%$ confidence intervals for ECE, AdaECE and Classwise-ECE using $1000$ bootstrap samples following \cite{Kumar2019verified}, and using ResNet-50/110 trained on CIFAR-10 (see Appendix~\ref{sec:bar_plots} for more results). Note that FLSD-53 produces the lowest calibration errors in general, and the difference in the metric values between FLSD-53 and other approaches (except Brier loss) is mostly statistically significant (i.e., confidence intervals don't overlap), especially before temperature scaling. In addition to the lower calibration errors, there are other advantages of focal loss as well, which we explore next.

\begin{figure*}[!t]
    \centering
    \subfigure[ResNet-110 (pre-T)]{\includegraphics[width=0.24\linewidth]{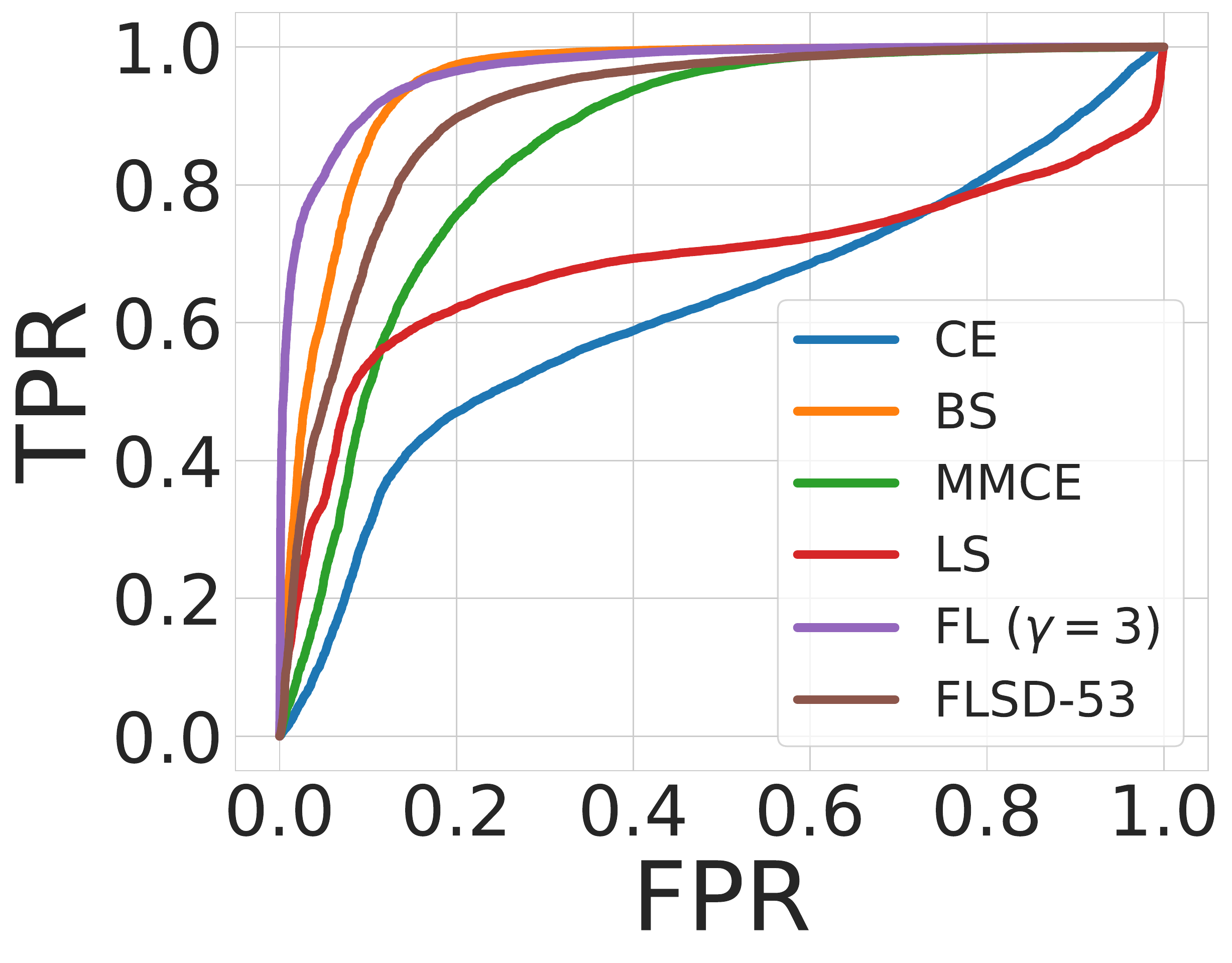}}
    \subfigure[ResNet-110 (post-T)]{\includegraphics[width=0.24\linewidth]{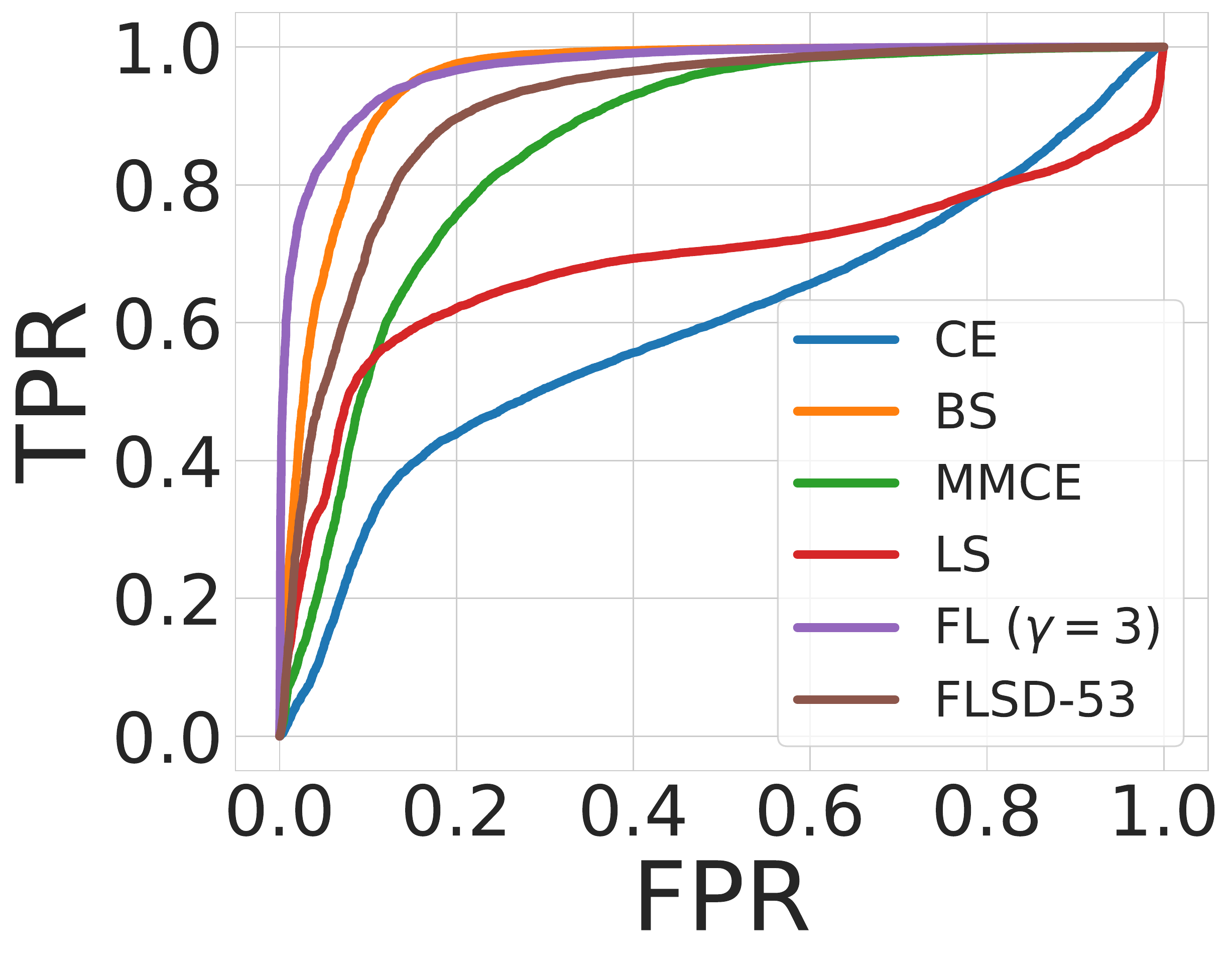}}
    \subfigure[Wide-ResNet (pre-T)]{\includegraphics[width=0.24\linewidth]{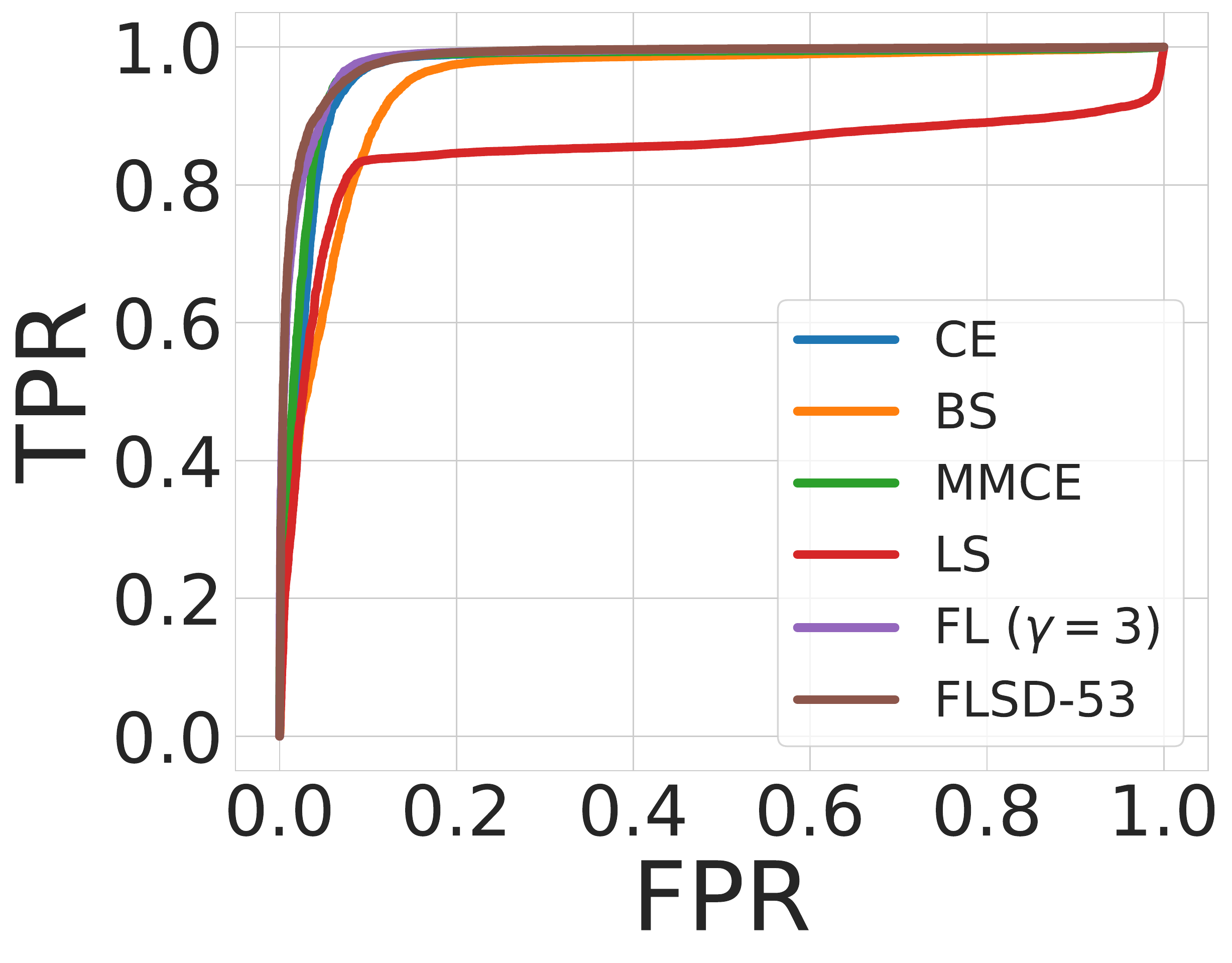}}
    \subfigure[Wide-ResNet (post-T)]{\includegraphics[width=0.24\linewidth]{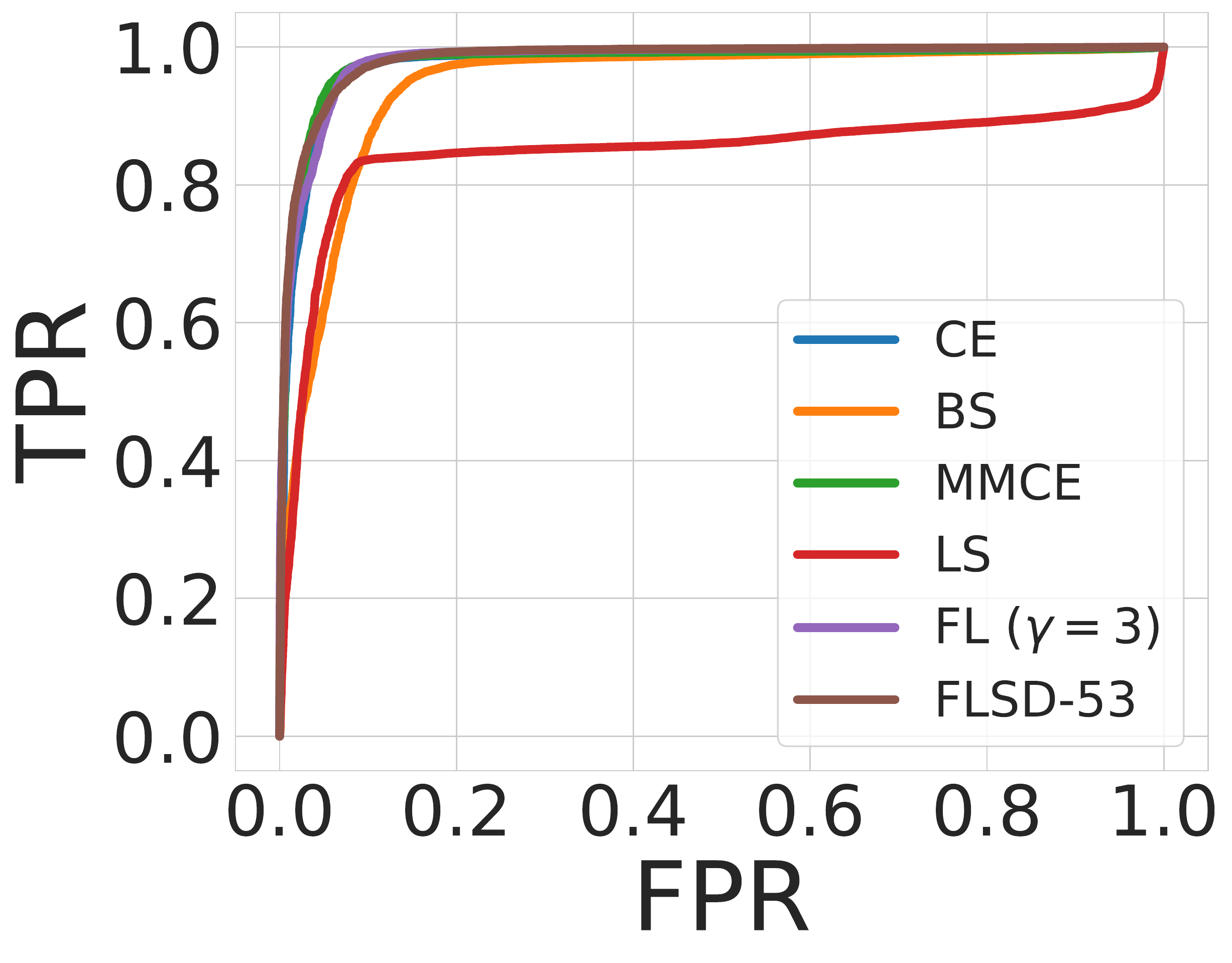}}
	\caption{ROC plots obtained from ResNet-110 and Wide-ResNet-26-10 architectures trained on CIFAR-10 (in-distribution) and tested on SVHN (OoD), both pre and post temperature scaling.}
  \label{fig:ood_roc_plots}
  \vspace{-1.2\baselineskip}
\end{figure*}

\textbf{More advantages of focal loss:}  \textit{Behaviour on Out-of-Distribution (OoD) data:} A perfectly calibrated model should have low confidence whenever it misclassifies, including when it encounters data which is OoD \citep{Thulasidasan2019mixup}. Although temperature scaling calibrates a model under the i.i.d.\ assumption, it is known to fail under distributional shift \citep{snoek2019can}. Since focal loss has implicit regularisation effects on the network (see \S\ref{sec:focalloss}), we investigate if it helps to learn representations that are more robust to OoD data. To do this, we use ResNet-110 and Wide-ResNet-26-10 trained on CIFAR-10 and consider the SVHN \citep{Netzer2011} test set and CIFAR-10-C \citep{Hendrycks2019benchmarking} with Gaussian noise corruption at severity 5 as OoD data. We use the entropy of the softmax distribution as the measure of confidence or uncertainty, and report the corresponding AUROC scores both before and after temperature scaling in Table \ref{table:auroc_tab1}. For both SVHN and CIFAR-10-C (using Gaussian noise), models trained on focal loss clearly obtain the highest AUROC scores. \textit{Note that Focal loss even without temperature scaling performs better than other methods with temperature scaling.} We also present the ROC plots pre and post temperature scaling for models trained on CIFAR-10 and tested on SVHN in Figure~\ref{fig:ood_roc_plots}. Thus, it is quite encouraging to note that models trained on focal loss are not only better calibrated under the i.i.d.\ assumption, but also seem to perform better than other competitive loss functions when we try shifting the distribution from CIFAR-10 to SVHN or CIFAR-10-C (pre and post temperature scaling).

\textit{Confident and Calibrated Models:} It is worth noting that focal loss with sample-dependent $\gamma$ has optimal temperatures that are very close to 1, mostly lying between 0.9 and 1.1 (see Table~\ref{table:ece_tab1}). This property is shown by the Brier loss and label smoothing models as well, albeit with worse calibration errors. By contrast, the temperatures for cross-entropy and MMCE models are significantly higher, with values lying between 2.0 and 2.8. An optimal temperature close to 1 indicates that the model is innately calibrated, and cannot be made significantly more calibrated by temperature scaling. In fact, a temperature much greater than 1 can make a model underconfident in general, as it is applied irrespective of the correctness of model outputs. We observe this empirically for ResNet-50 and ResNet-110 trained on CIFAR-10. Although models trained with cross-entropy have much higher confidence before temperature scaling than those trained with focal loss, after temperature scaling, focal loss models are significantly more confident in their predictions. We provide quantitative and qualitative empirical results to support this claim in Appendix~\ref{conf_and_cal}.
\begin{table*}[!t]
	\centering
	\scriptsize
	\resizebox{\linewidth}{!}{%
		\begin{tabular}{cccccccccccccc}
			\toprule
			\textbf{Dataset} & \textbf{Model} & \multicolumn{2}{c}{\textbf{Cross-Entropy}} &
			\multicolumn{2}{c}{\textbf{Brier Loss}} & \multicolumn{2}{c}{\textbf{MMCE}} &
			\multicolumn{2}{c}{\textbf{LS-0.05}} & \multicolumn{2}{c}{\textbf{FL-3 (Ours)}} &
			\multicolumn{2}{c}{\textbf{FLSD-53 (Ours)}} \\
			&& Pre T & Post T & Pre T & Post T & Pre T & Post T & Pre T & Post T & Pre T & Post T & Pre T & Post T \\
			\midrule
			\multirow{2}{*}{CIFAR-10/SVHN} & ResNet-110&61.71&59.66&94.80&95.13&85.31&85.39&68.68&68.68&\textbf{96.74}&\textbf{96.92}&90.83&90.97\\
			& Wide-ResNet-26-10&96.82&97.62&94.51&94.51&97.35&97.95&84.63&84.66&98.19&98.05&\textbf{98.29}&\textbf{98.20}\\
			\midrule
			\multirow{2}{*}{CIFAR-10/CIFAR-10-C} & ResNet-110&77.53&75.16&84.09&83.86&71.96&70.02&72.17&72.18&82.27&82.18&\textbf{85.05}&\textbf{84.70}\\
			& Wide-ResNet-26-10&81.06&80.68&85.03&85.03&82.17&81.72&71.10&71.16&82.17&81.86&\textbf{87.05}&\textbf{87.30}\\
			\bottomrule
		\end{tabular}%
	}
	\caption{AUROC $(\%)$ computed for models trained on CIFAR-10 (in-distribution), and using SVHN and CIFAR-10-C (Gaussian Noise corruption with severity level 5) respectively as the OoD datasets. \vspace{-3mm}}
	\label{table:auroc_tab1}
\end{table*}

\vspace{-2.5mm}
\section{Conclusion}
\vspace{-2mm}
In this paper, we have studied the properties of focal loss, an alternative loss function that can yield classification networks that are more naturally calibrated than those trained using the conventional cross-entropy loss, while maintaining accuracy. In particular, we show in \S\ref{sec:focalloss} that focal loss implicitly maximises entropy while minimising the KL divergence between the predicted and the target distributions. We also show that, because of its design, it naturally regularises the weights of a network during training, reducing NLL overfitting and thereby improving calibration. Furthermore, we empirically observe that models trained using focal loss are not only better calibrated under i.i.d.\ assumptions, but can also be better at detecting OoD samples which we show by taking CIFAR-10 as the in-distribution dataset and SVHN and CIFAR-10-C as out-of-distribution datasets, something which temperature scaling fails to achieve. 

\newpage
\section{Broader Impact}
Our work shows that using the right kind of loss function can lead to a calibrated model. This helps in improving the reliability of these models when used in real-world applications. It can help in deployment of the models such that users can be alerted when its prediction may not be trustworthy. We do not directly see a situation where calibrated neural networks can have a negative impact on society, but we do believe that research on making models more calibrated will help improve fairness and trust in AI.

\begin{ack}
This work was started whilst J Mukhoti was at FiveAI, and completed after he moved to the University of Oxford. V Kulharia is wholly funded by a Toyota Research Institute grant. A Sanyal acknowledges support from The Alan Turing Institute under the Turing Doctoral Studentship grant TU/C/000023. This work was also supported by the Royal Academy of Engineering under the Research Chair and Senior Research Fellowships scheme, EPSRC/MURI grant EP/N019474/1 and FiveAI.
\end{ack}

\bibliographystyle{plainnat}
\bibliography{neurips_2020}

\clearpage
\appendix

\counterwithin{figure}{section}
\counterwithin{table}{section}

\newtheorem*{prop1}{Proposition 1}
\newtheorem*{prop2}{Proposition 2}

\section*{Appendix: Calibrating Deep Neural Networks using Focal Loss}

In \S\ref{rel_plots_appendix}, we provide some empirical evidence for the observation made in \S\ref{sec:cause_cali} in the main paper using reliability plots. In \S\ref{reg_bregman}, we discuss the relation between focal loss and a regularised KL divergence, where the regulariser is the entropy of the predicted distribution. In \S\ref{linear_model}, we discuss the regularisation effect of focal loss on a simple setup, i.e.\ a generalised linear model trained on a simple data distribution. In \S\ref{sec:proof}, we show the proofs of the two propositions formulated in the main text. We then describe all the datasets and implementation details for our experiments in \S\ref{dataset}. In \S\ref{results}, we discuss additional approaches for training using focal loss, and also the results we get from these approaches. We also provide the Top-5 accuracies of several models as we speculate that calibrated models should have a higher softmax probability on the correct class even when they misclassify, as compared to models which are less calibrated. We further provide the results of evaluating our models using various metrics other than ECE (like AdaECE, Classwise-ECE, MCE and NLL). Next, in \S\ref{sec:bar_plots}, we provide additional results related to the confidence interval experiments performed in \S\ref{sec:experiments} of the main paper. In \S\ref{conf_and_cal}, we provide empirical and qualitative results to show that models trained using focal loss are calibrated, whilst maintaining their confidence on correct predictions. In \S\ref{feature_norm}, we provide a brief extension of our discussion about Figure \ref{fig:nll_corr_incorr_entropy}(e) in the main paper, with a plot of $L_2$ norms of features obtained from the last ResNet block during training. In \S\ref{sec:early_stopping}, we provide some empirical evidence to support the claims we make in \S\ref{sec:focalloss} of the main paper about early stopping. Finally, in \S\ref{sec:downstream_task}, we discuss the performance of focal loss on the downstream task of machine translation with beam search. We choose machine translation as the downstream task because in machine translation, softmax vectors from a model are directly fed into the beam search algorithm, and hence more calibrated probability vectors should intuitively produce better translations

\section{Reliability Plots}
\label{rel_plots_appendix}
\begin{figure}[!t]
\centering
    \subfigure{\includegraphics[width=0.20\linewidth]{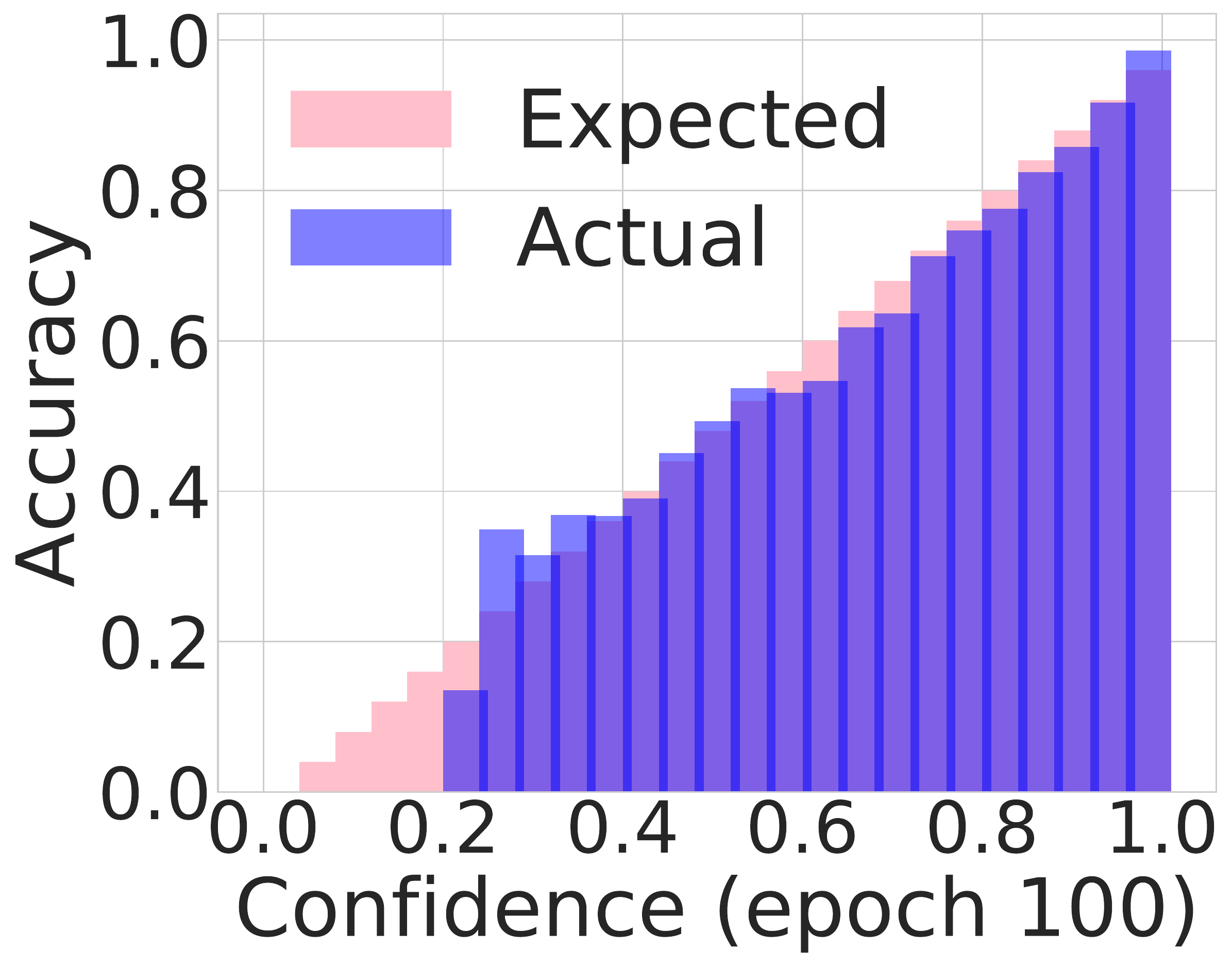}}
    \subfigure{\includegraphics[width=0.20\linewidth]{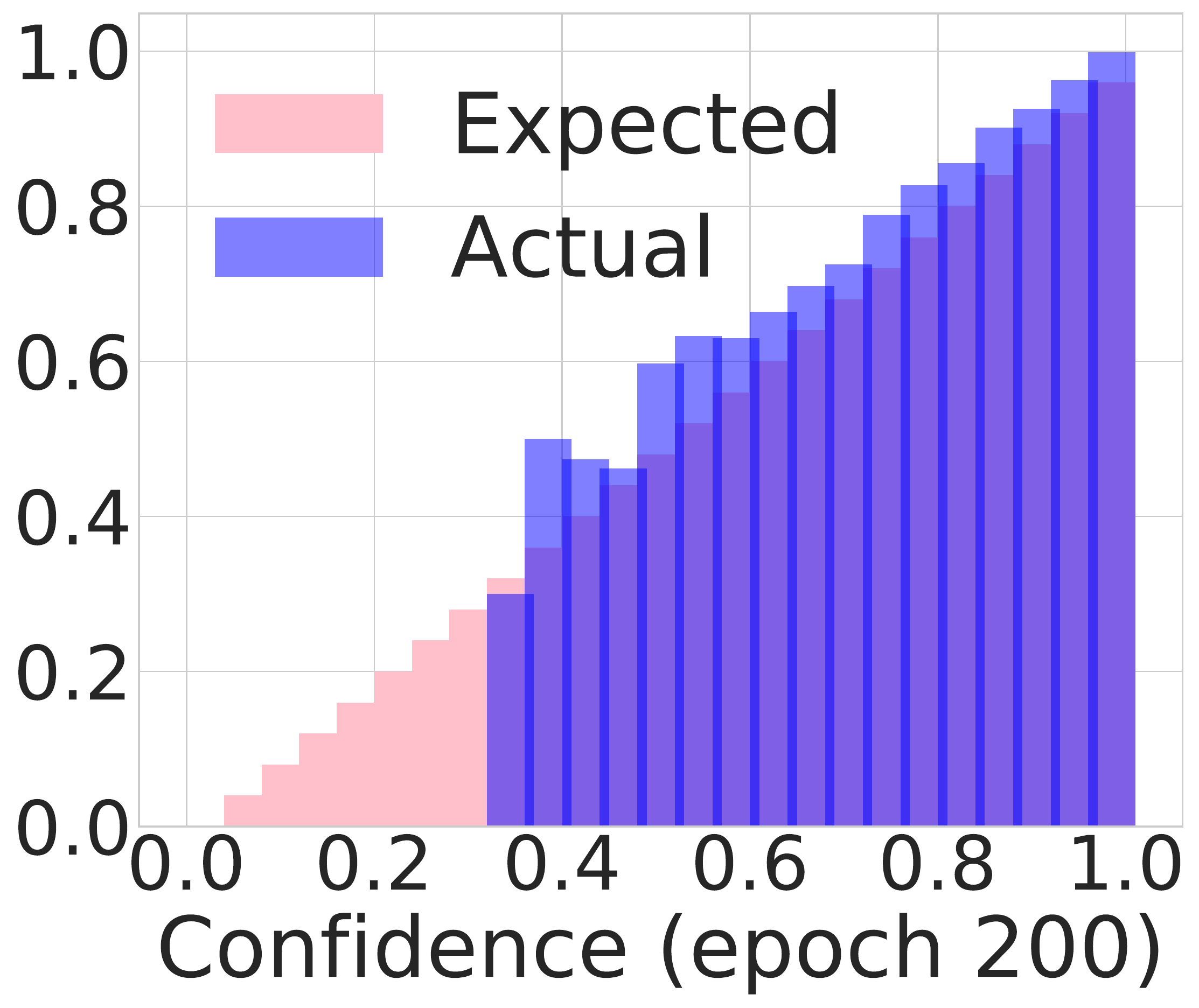}}
    \subfigure{\includegraphics[width=0.20\linewidth]{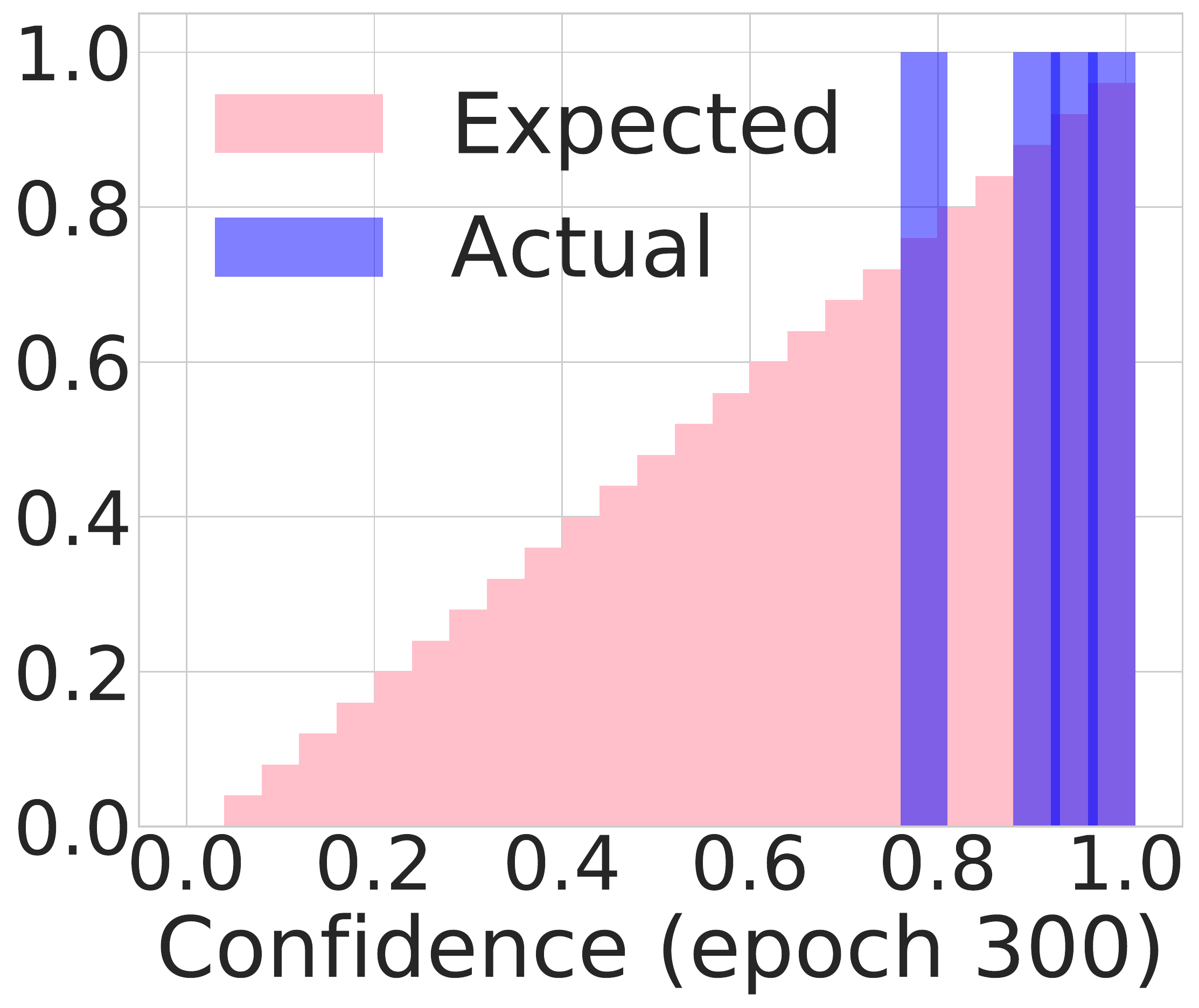}}
    \subfigure{\includegraphics[width=0.20\linewidth]{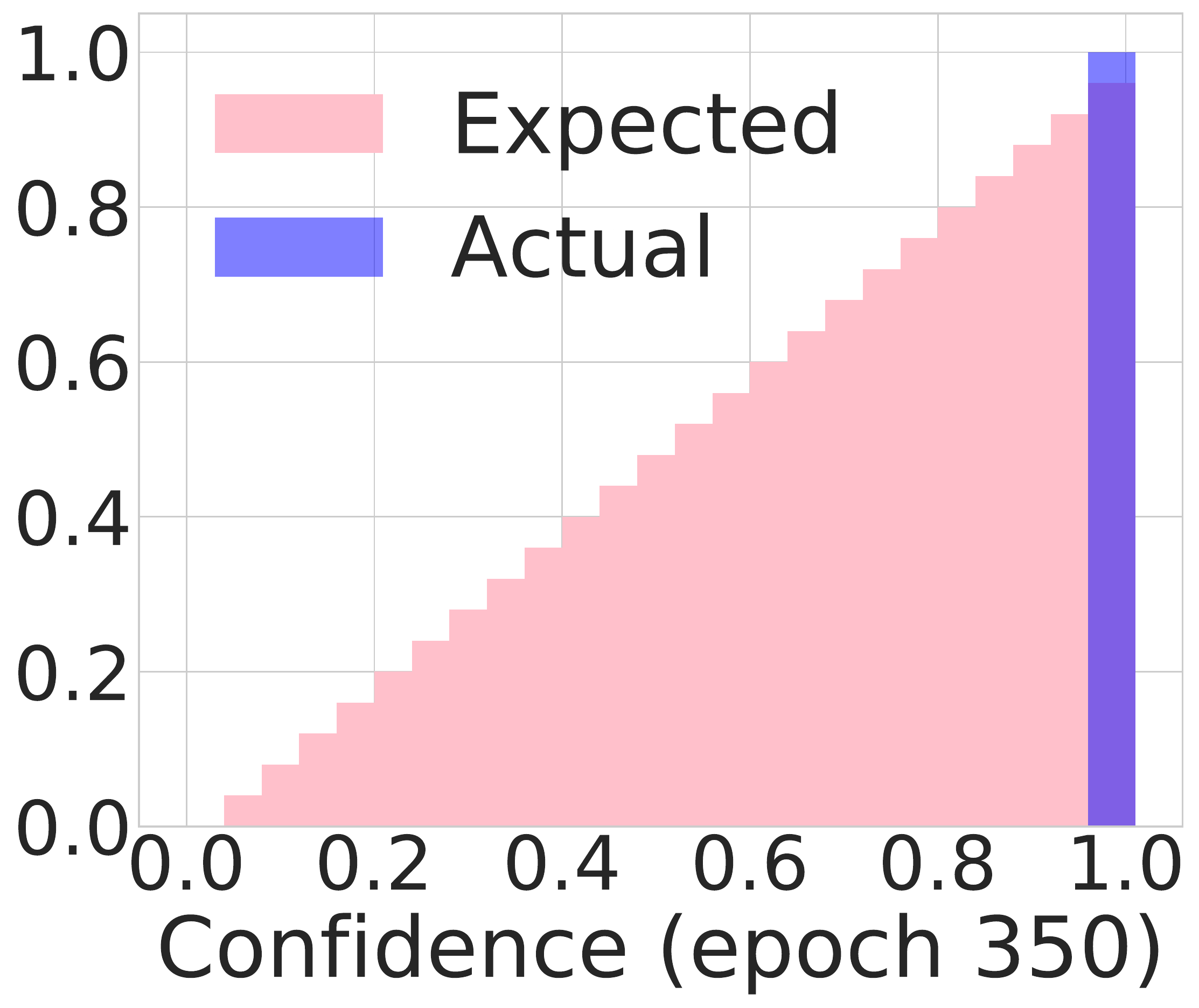}}
    \subfigure{\includegraphics[width=0.20\linewidth]{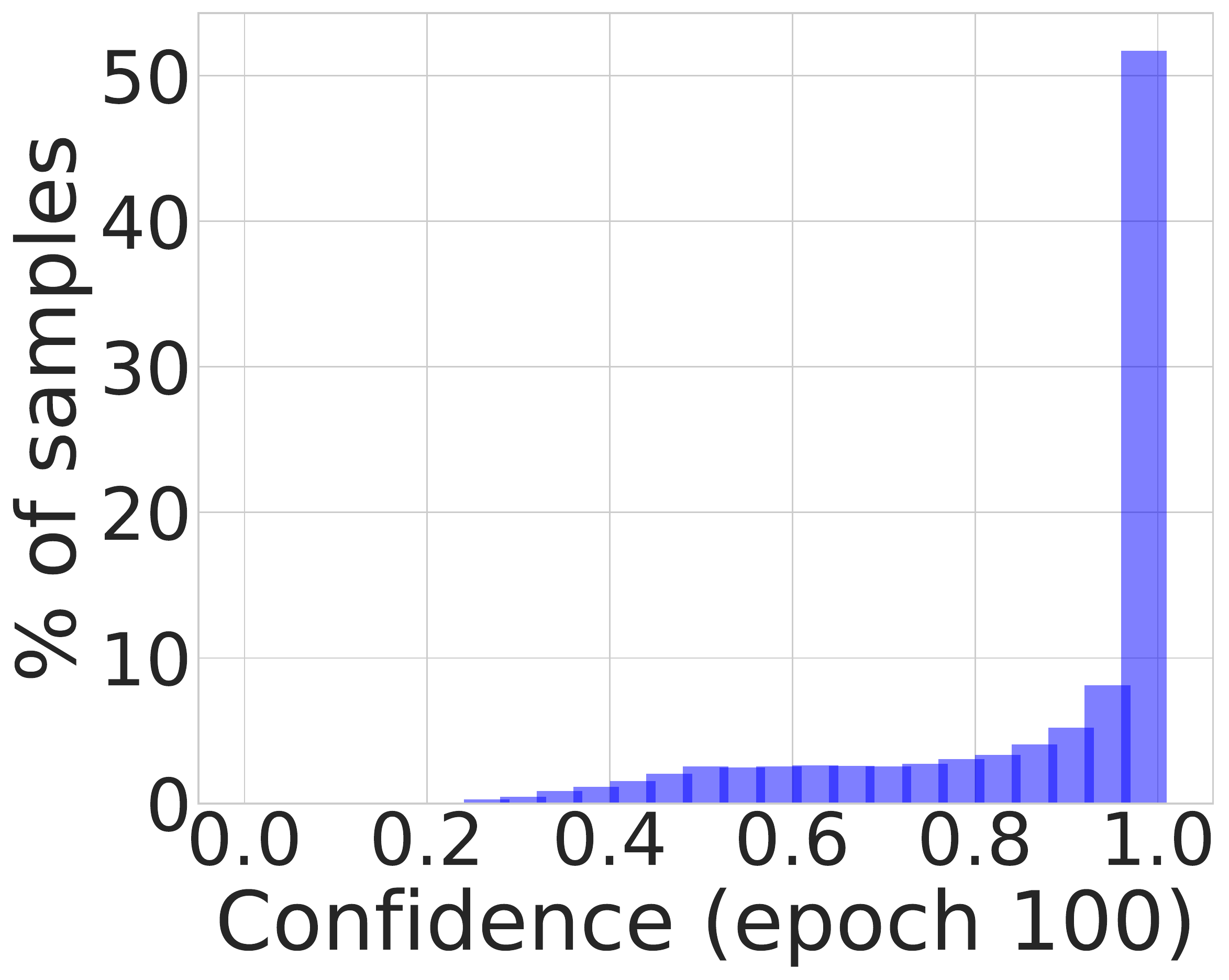}}
    \subfigure{\includegraphics[width=0.20\linewidth]{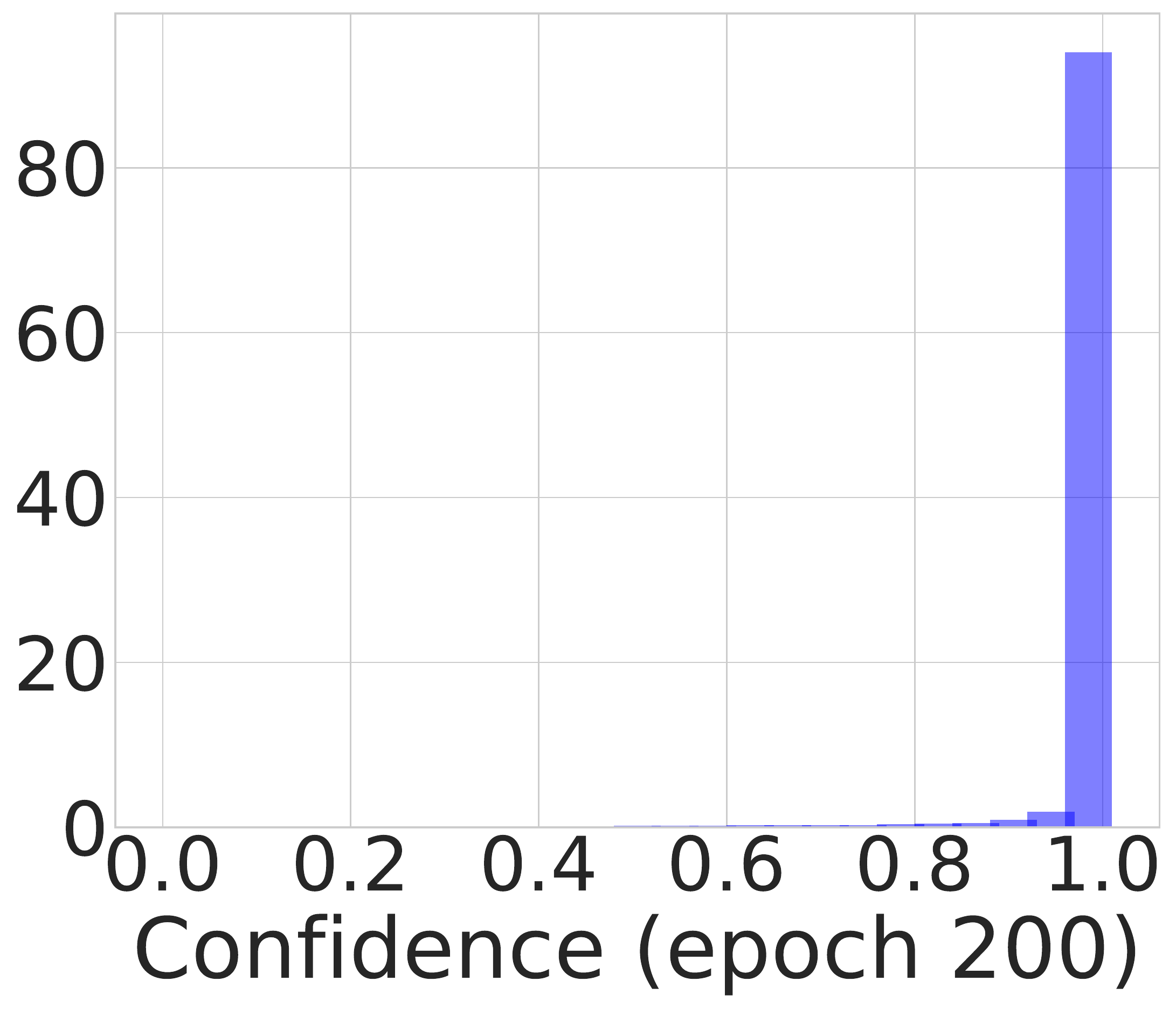}}
    \subfigure{\includegraphics[width=0.20\linewidth]{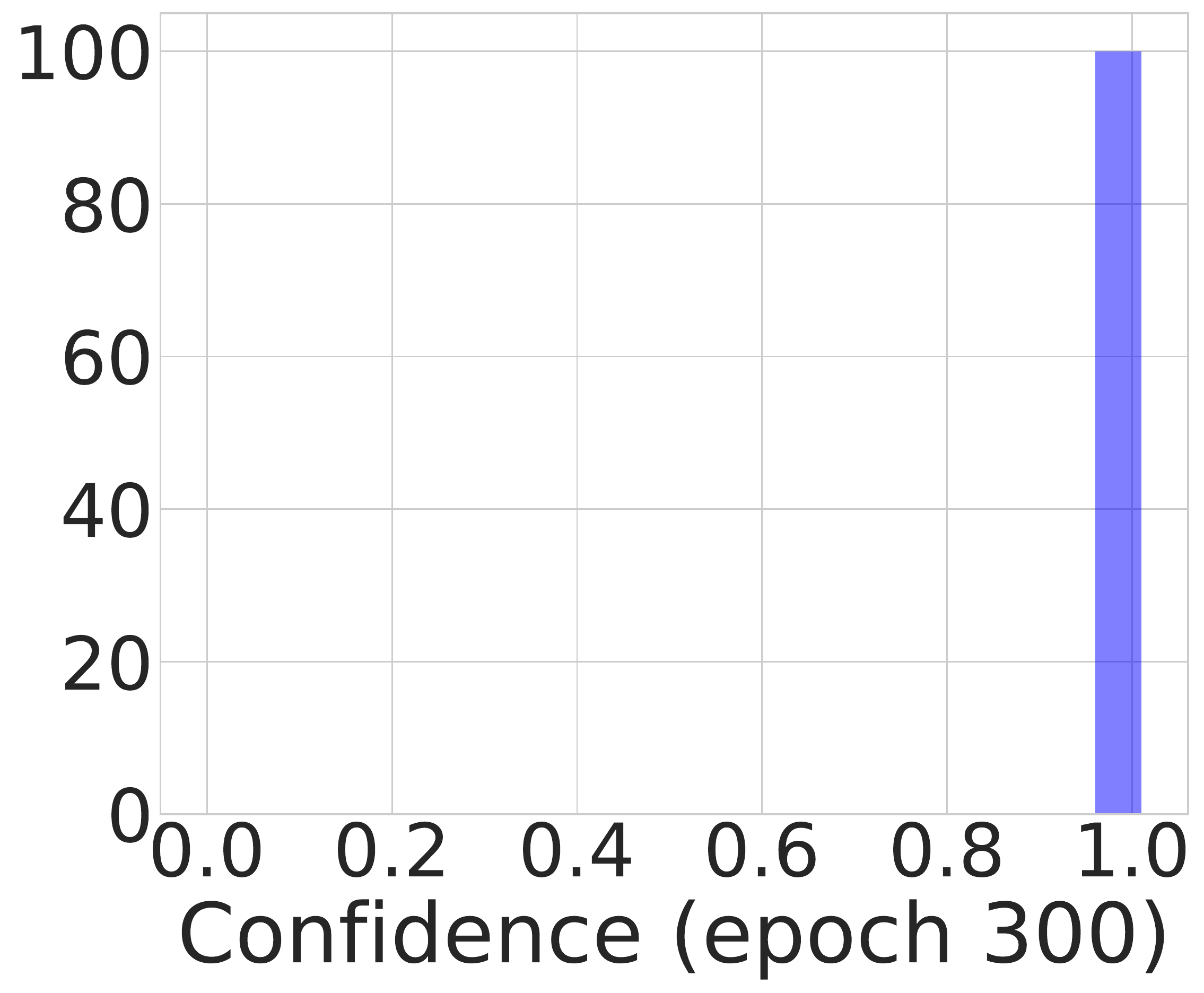}}
    \subfigure{\includegraphics[width=0.20\linewidth]{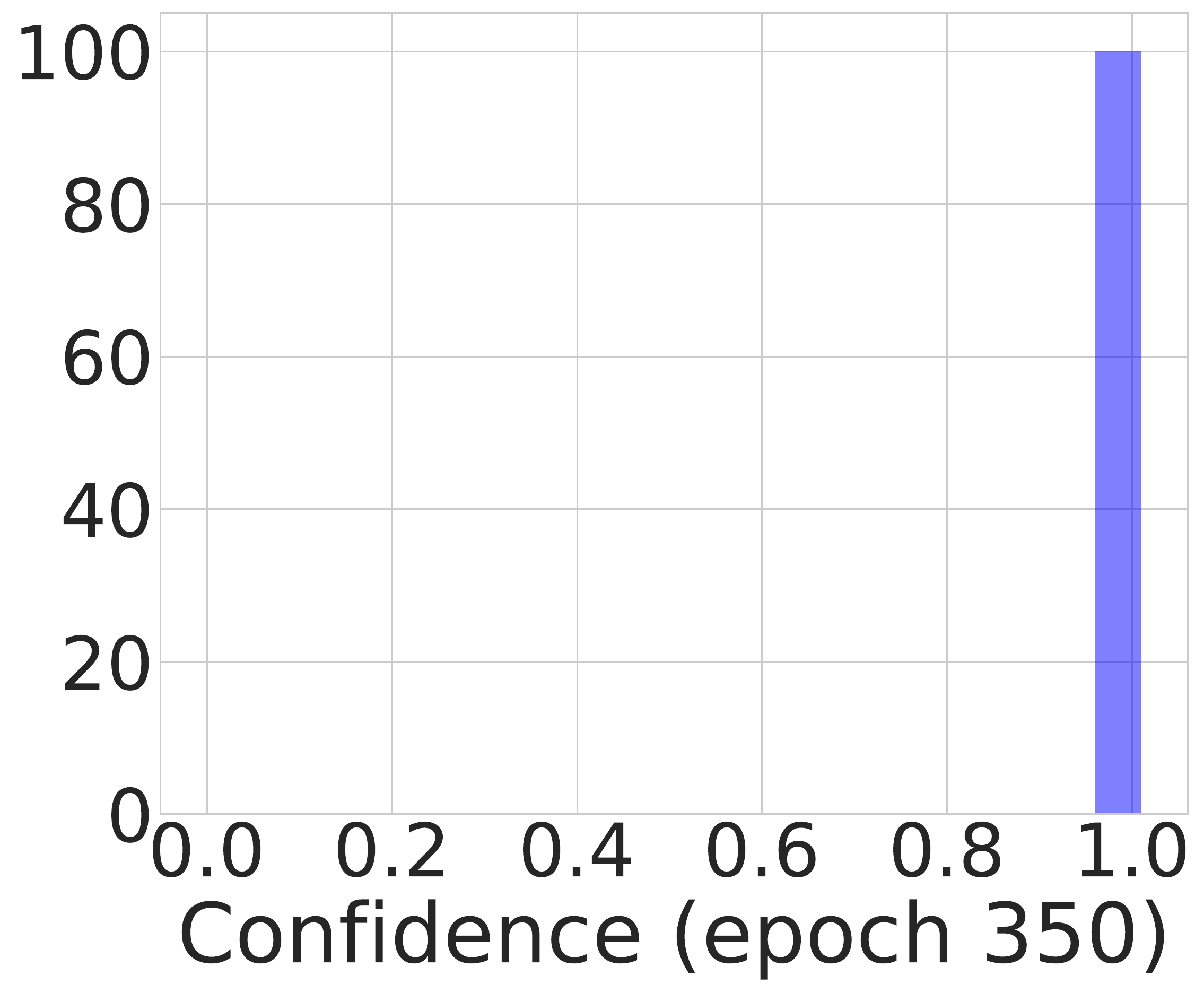}}
\vspace{-\baselineskip}
\caption{The confidence values for training samples at different epochs during the NLL training of a ResNet-50 on CIFAR-10 (see \S\ref{sec:cause_cali}). Top row: reliability plots using $25$ confidence bins; bottom row: \% of samples in each bin. As training progresses, the model gradually shifts all training samples to the highest confidence bin. Notably, it continues to do so even after achieving 100\% training accuracy by the $300$ epoch point.
}
\label{fig:rel_conf_bin_plot}
\vspace{-\baselineskip}
\end{figure}

In this section, we provide some empirical evidence to support the observation made in \S\ref{sec:cause_cali} of the main paper that a model, even after attaining perfect training accuracy, can reduce the training NLL (loss) further by increasing the prediction confidences to match the ground-truth one-hot encoding. To empirically observe this, we use the ResNet-50 network used for the analysis in \S\ref{sec:cause_cali}. We divide the confidence range $[0, 1]$ into 25 bins, and present reliability plots computed on the training set at training epochs $100$, $200$, $300$ and $350$ (see the top row of Figure~\ref{fig:rel_conf_bin_plot}). In Figure~\ref{fig:rel_conf_bin_plot}, we also show the percentage of samples in each confidence bin. It is quite clear from these plots that over time, the network gradually pushes all of the training samples towards the highest confidence bin. Furthermore, even though the network has achieved $100\%$ accuracy on the training set by epoch $300$, it still pushes some of the samples lying in lower confidence bins to the highest confidence bin by epoch $350$.

\section{Relation between Focal Loss and Entropy Regularised KL Divergence}
\label{reg_bregman}
Here we show why focal loss favours accurate but relatively less confident solutions. We show that it inherently provides a trade-off between minimizing the KL-divergence and maximizing the entropy, depending on the strength of $\gamma$. 
We use $\mathcal{L}_f$ and $\mathcal{L}_c$ to denote the focal loss with parameter $\gamma$ and cross entropy between $\hat{p}$ and $q$, respectively. $K$ denotes the number of classes and $q_y$ denotes the ground-truth probability assigned to the $y^{th}$ class (similarly for $\hat{p}_y$). We consider the following simple extension of focal loss: 
\begin{align*}
\mathcal{L}_f &= -\sum_{y=1}^K (1 -  \hat{p}_{y})^\gamma q_{y} \log{\hat{p}_{y}}\\
        &\ge -\sum_{y=1}^K(1 - \gamma \hat{p}_{y})q_{y} \log{\hat{p}_{y}} &&\text{By Bernoulli's inequality $\forall \gamma \ge 1$, since $\hat{p}_{y} \in [0,1]$}\\
        &=- \sum_{y=1}^K q_{y} \log{\hat{p}_{y}} - \gamma \left|\sum_{y=1}^K q_{y} \hat{p}_{y} \log{\hat{p}_{y}}\right|&& \text{$\forall y$, $\log{\hat{p}_{y}}\le 0$}\\
        &\ge-\sum_{y=1}^K q_{y} \log{\hat{p}_{y}}  - \gamma\max_j q_{j} \sum_{y=1}^K |\hat{p}_{y} \log{\hat{p}_{y}}|&&\text{By H\"older's inequality $||fg||_1 \leq ||f||_{\infty}||g||_1$}\\ 
        &\ge-\sum_{y=1}^K q_{y} \log{\hat{p}_{y}}  + \gamma\sum_{y=1}^K \hat{p}_{y} \log{\hat{p}_{y}}&&\forall j, q_j \in [0,1]\\
        &= \mathcal{L}_c - \gamma\mathbb{H}[\hat{p}].
\end{align*}

We know that $\mathcal{L}_c = \mathrm{KL}(q||\hat{p})+  \mathbb{H}[q]$. Combining this equality with the above inequality leads to:
\begin{align*}
\mathcal{L}_f \geq \mathrm{KL}(q||\hat{p})+  \underbrace{\mathbb{H}[q]}_{constant} - \gamma \mathbb{H}[\hat{p}].
\end{align*}

In the case of one-hot encoding (Delta distribution for $q$), focal loss will maximize $-\hat{p}_y \log \hat{p}_y$ (let $y$ be the ground-truth class index), the component of the entropy of $\hat{p}$ corresponding to the ground-truth index. Thus, it will prefer learning $\hat{p}$ such that $\hat{p}_y$ is assigned a higher value 
\begin{wrapfigure}{r}{0.25\textwidth}
  \begin{center}
    \includegraphics[width=\linewidth]{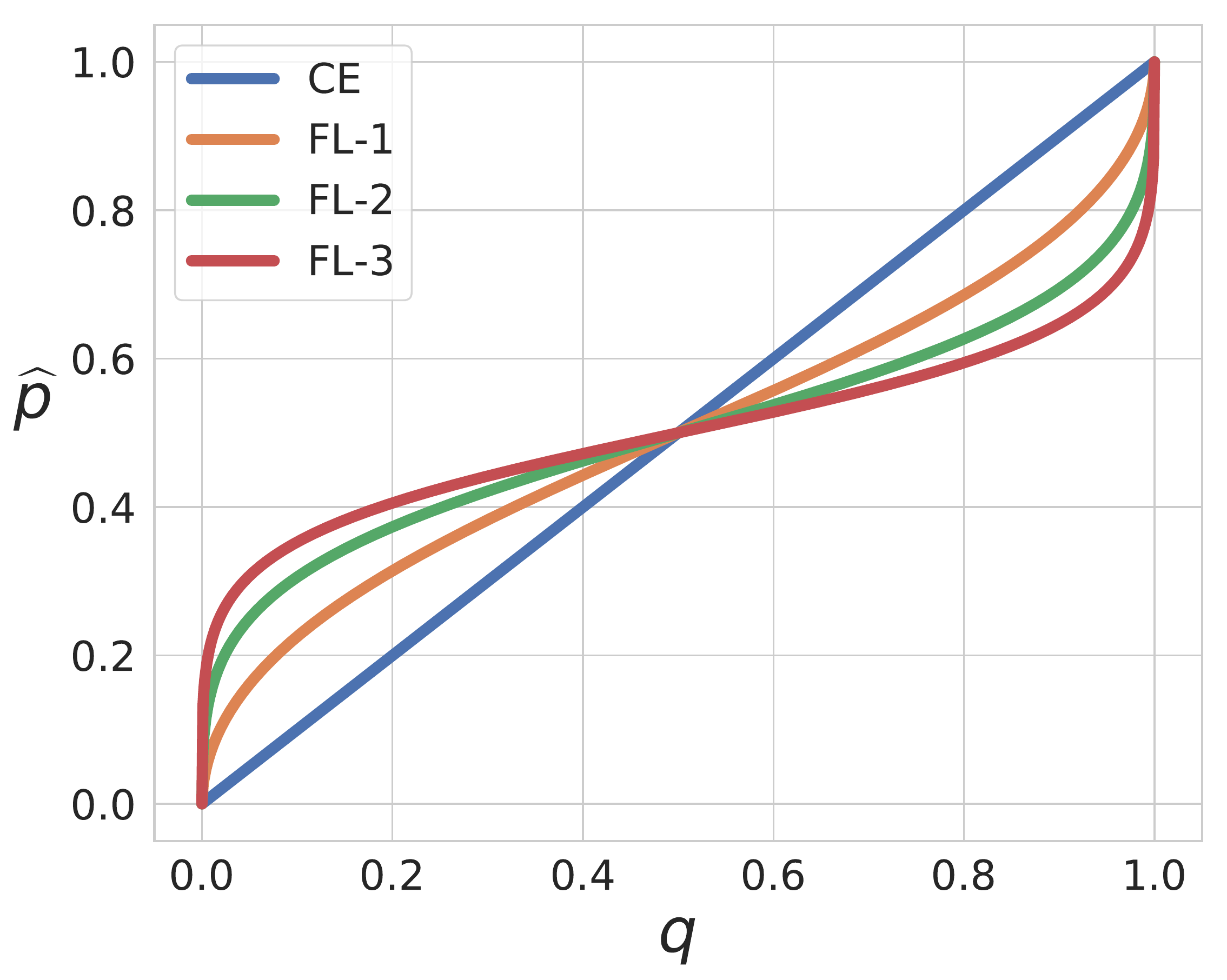}
  \end{center}
  \caption{Optimal $\hat{p}$ for various values of $q$.}
  \label{fig:soln_p}
\end{wrapfigure}
(because of the KL term), but not too high (because of the entropy term), and will ultimately avoid preferring overconfident models (by contrast to cross-entropy loss). 
Experimentally, we found the solution of the cross-entropy and focal loss equations, i.e.\ the value of the predicted probability $\hat{p}$ which minimises the loss, for various values of $q$ in a binary classification problem (i.e.\ $K=2$), and plotted it in Figure~\ref{fig:soln_p}. As expected, focal loss favours a more entropic solution $\hat{p}$ that is closer to $0.5$. In other words, as Figure~\ref{fig:soln_p} shows, solutions to focal loss (Equation~\ref{eq:fc_loss}) will always have higher entropy than those of cross-entropy, depending on the value of $\gamma$.
\begin{equation}
\hat{p} = \mathrm{argmin}_x \; -(1-x)^\gamma q \log{x} - x^\gamma (1-q) \log{(1 - x)}\quad 0\le x\le 1\label{eq:fc_loss}
\end{equation}
\section{Focal Loss and Cross-Entropy on a Linear Model}
\label{linear_model}

\begin{figure*}[!htb]
	\centering
	\subfigure[]{\includegraphics[width=0.35\linewidth]{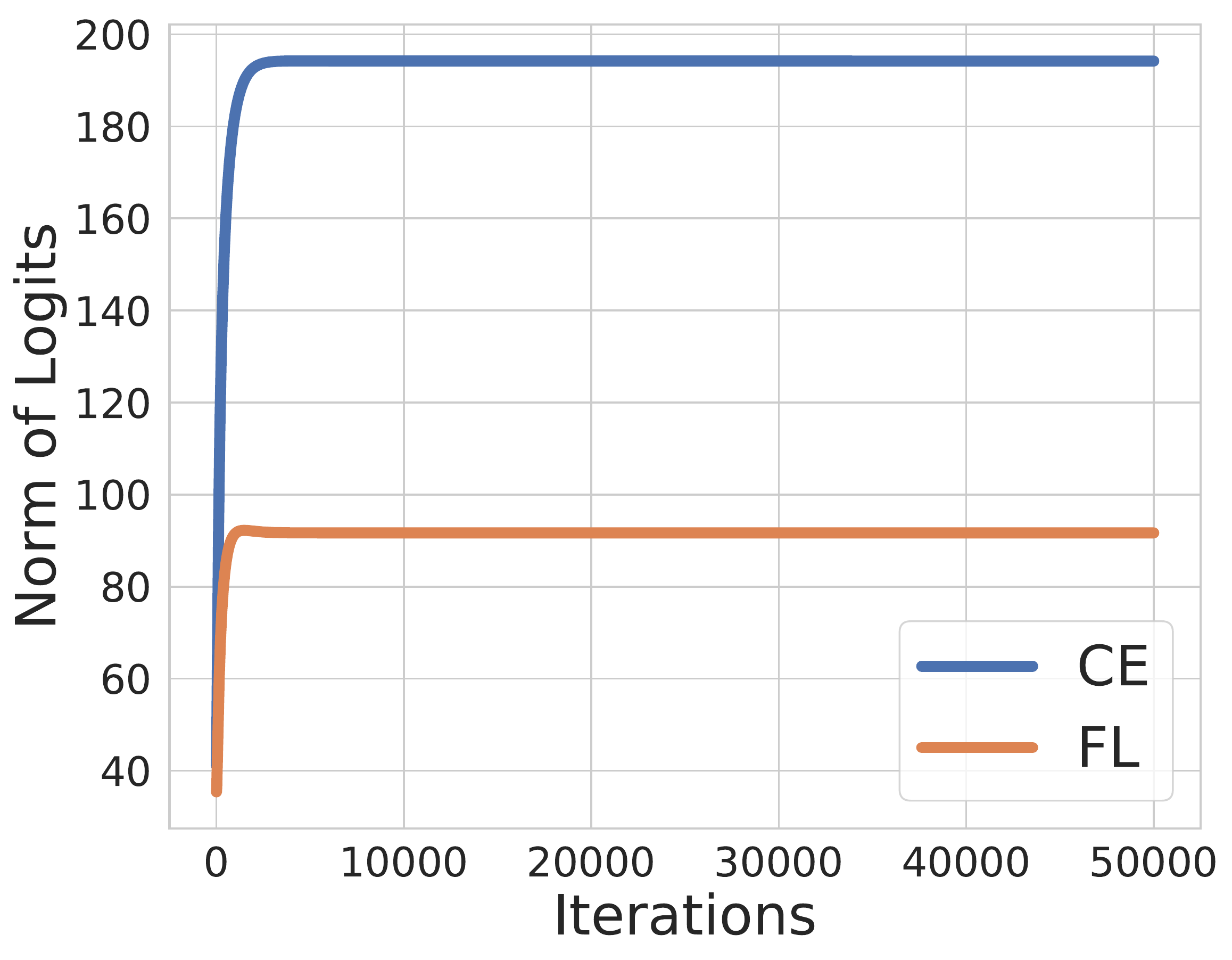}}
	\subfigure[]{\includegraphics[width=0.35\linewidth]{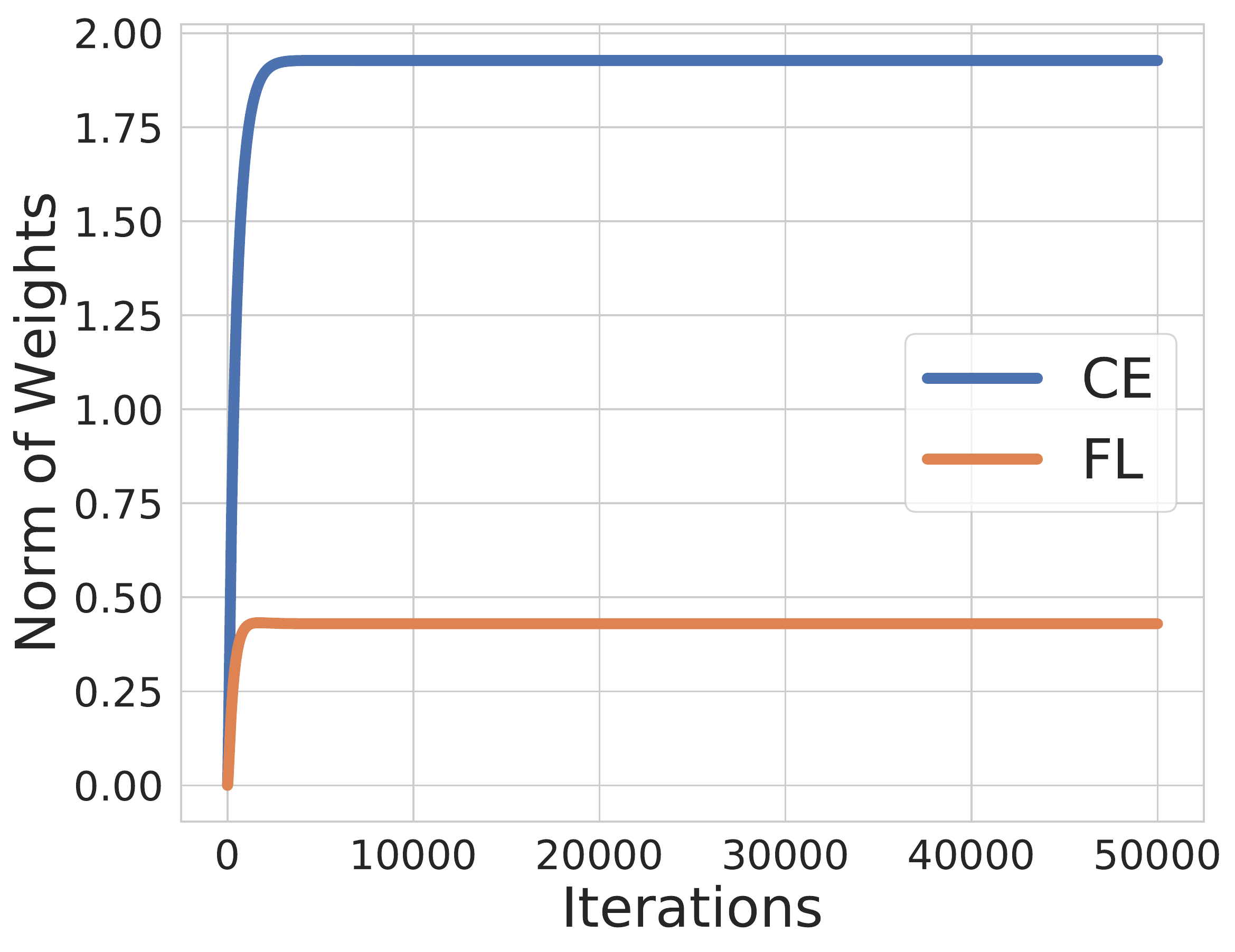}}
	\caption{(a): Norm of logits (b): Norm of weights.}
	\vspace{-\baselineskip}
	\label{fig:norms_linear}
\end{figure*}

The behaviour of deep neural networks is generally quite different from linear models and the problem of calibration is more pronounced in the case of deep neural networks, hence we focus on analysing the calibration of deep networks in the paper. However, weight norm analysis for the initial layers is complex due to batchnorm and weight decay. Hence, to see the effect of weight magnification on miscalibration, here we use a simple network without batchnorm or weight decay, which is a generalised linear model, and a simple data distribution.

\paragraph{Setup} We consider a binary classification problem. The data matrix $\mathbf{X}\in\mathbb{R}^{2\times N}$ is created by assigning each class, two normally distributed clusters such that the mean of the clusters are linearly separable. The mean of the clusters are situated on the vertices of a two-dimensional hypercube of side length 4. The standard deviation for each cluster is $1$ and the samples are randomly linearly combined within each cluster in order to add covariance. Further, for $10\%$ of the data points, the labels were flipped. $4000$ samples are used for training and $1000$ samples are used for testing. The model consists of a simple 2-parameter logistic regression model. $f_{\mathbf{w}}(\mathbf{x}) = \sigma(w_1x_1+w_2x_2)$. We train this model using both cross-entropy and focal loss with $\gamma = 1$.

\paragraph{Weight Magnification} We have argued that focal loss implicitly regularizes the weights of the model by providing smaller gradients as compared to cross-entropy. This helps in calibration as, if all the weights are large,  the logits are large and thus the confidence of the network is large on all test points, even on the misclassified points. When the model misclassifies, it misclassifies  with a high confidence. Figure~\ref{fig:norms_linear}  shows, for a generalised linear model, that the norm of the logits and the weights of a network blows for Cross Entropy as compared to Focal Loss.

\begin{figure*}[!htb]
	\centering
	\subfigure[]{\includegraphics[width=0.30\linewidth]{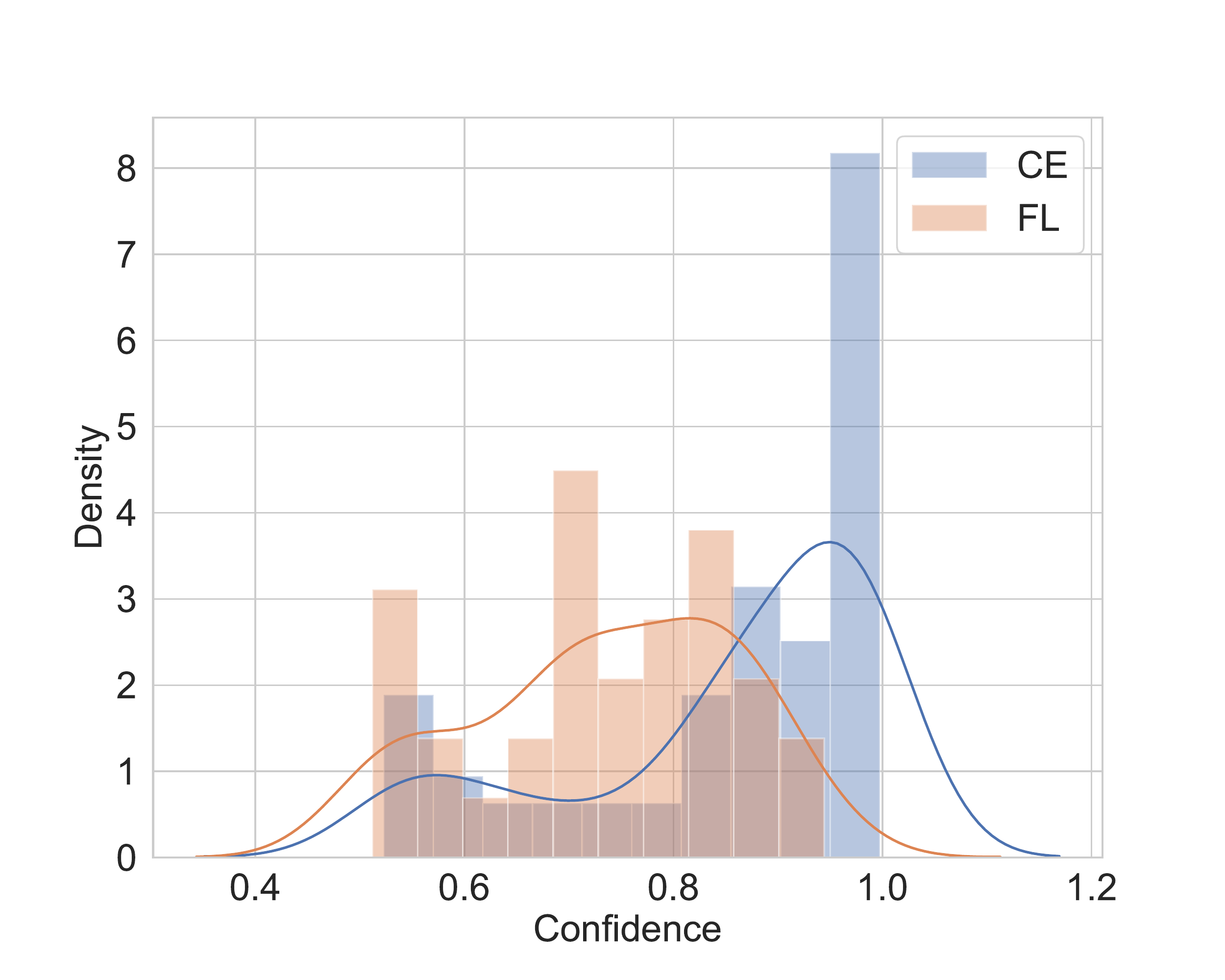}}
	\subfigure[]{\includegraphics[width=0.30\linewidth]{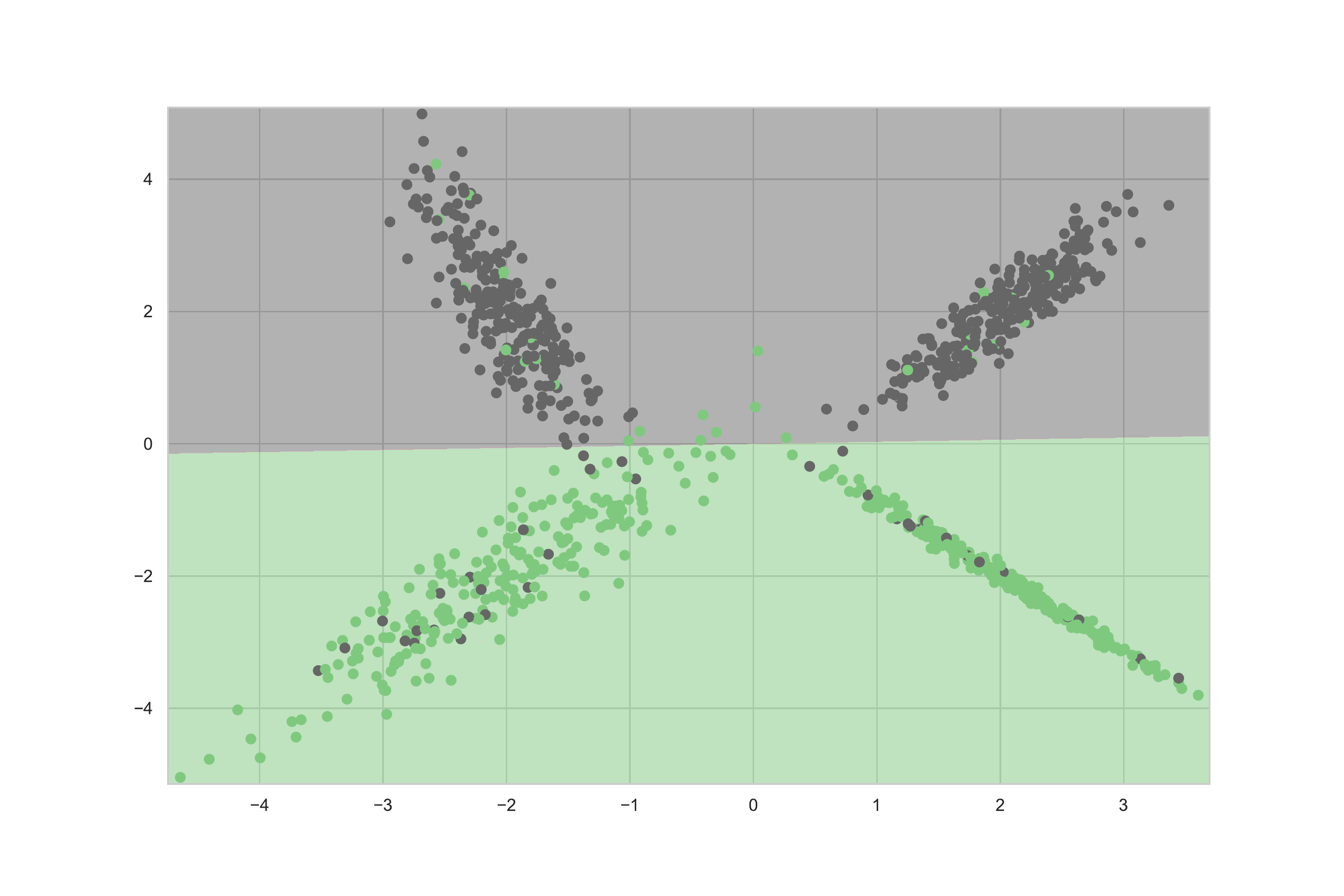}}
	\subfigure[]{\includegraphics[width=0.30\linewidth]{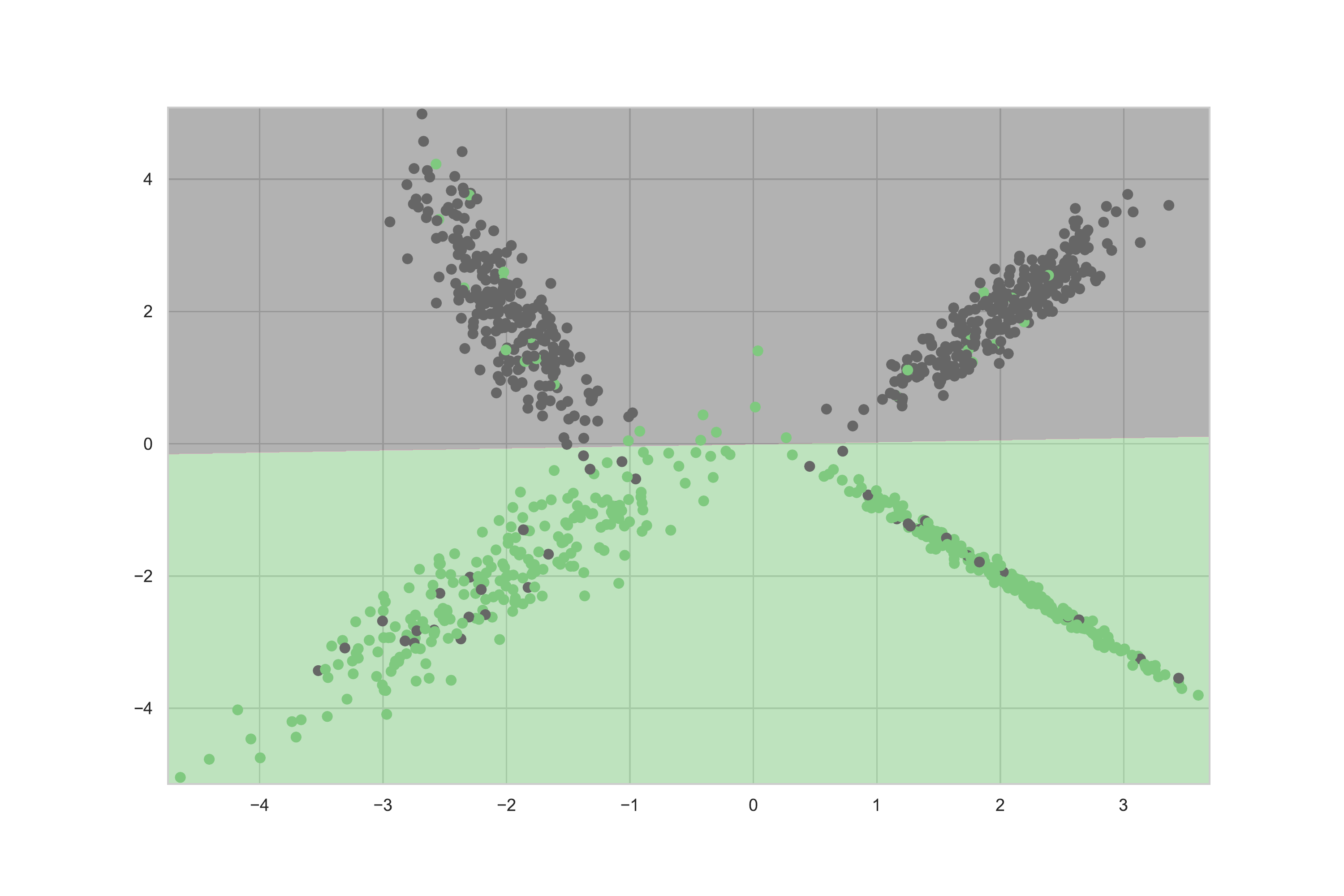}}
	\caption{(a): Confidence of mis-classifications (b): Decision boundary of linear classifier trained using cross entropy (c): Decision boundary of linear classifier trained using focal loss}
	\vspace{-\baselineskip}
	\label{fig:conf_wrong_dec_bound}
\end{figure*}

\paragraph{High Confidence for mistakes} Figures~\ref{fig:conf_wrong_dec_bound} (b) and (c) show that running gradient descent with cross-entropy (CE) and focal loss (FL) both gives the same decision regions i.e. the weight vector points in the same region for both FL and CE. However, as we have seen that the norm of the weights is much larger for CE as compared to FL, we would expect the confidence of misclassified test points to be large for CE as compared to FL. Figure~\ref{fig:conf_wrong_dec_bound} (a) plots a histogram of the confidence of the misclassified points and it shows that CE misclassifies almost always with greater than $90\%$ confidence whereas FL misclassifies with much lower confidence.

\section{Proofs}
\label{sec:proof}
Here we provide the proofs of both the propositions presented in the main text. While Proposition \textcolor{red}{1} helps us understand the regularization effect of focal loss, Proposition \textcolor{red}{2} provides us the $\gamma$ values in a principled way such that it is sample-dependent. Implementing the sample-dependent $\gamma$ is very easy as implementation of the Lambert-W function~\citep{corless1996lambertw} is available in standard libraries (e.g.\ python scipy).

\begin{prop1}
For focal loss $\loss_f$ and cross-entropy $\loss_c$, the gradients $\frac{\partial \loss_f}{\partial \bfw} = \frac{\partial \loss_c}{\partial \bfw} g(\hat{p}_{i,y_i}, \gamma)$, where $g(p, \gamma) = (1-p)^\gamma - \gamma p (1-p)^{\gamma - 1} \log(p)$, $\gamma \in \mathbb{R}^+$ is the focal loss hyperparameter, and $\bfw$ denotes the parameters of the last linear layer. Thus $\norm{\frac{\partial  \loss_f}{\partial \bfw}} \leq \norm{\frac{\partial  \loss_c}{\partial \bfw}}$ if $g(\hat{p}_{i,y_i}, \gamma) \in [0, 1]$.

\end{prop1}
\begin{proof}
Let $\bfw$ be the linear layer parameters connecting the feature map to the logit $s$. Then, using the chain rule, $\frac{\partial \loss_f}{\partial \bfw} = \Big( \frac{\partial s}{\partial \bfw} \Big) \Big( \frac{\partial \hat{p}_{i,y_i}}{\partial s} \Big) \Big( \frac{\partial \loss_f}{\partial \hat{p}_{i,y_i}} \Big)$. Similarly, $\frac{\partial \loss_c}{\partial \bfw} = \Big( \frac{\partial s}{\partial \bfw} \Big) \Big( \frac{\partial \hat{p}_{i,y_i}}{\partial s} \Big) \Big( \frac{\partial \loss_c}{\partial \hat{p}_{i,y_i}} \Big)$. The derivative of the focal loss with respect to $\hat{p}_{i,y_i}$, the softmax output of the network for the true class $y_i$, takes the form
\begin{equation}
\begin{split}
        \frac{\partial \loss_f}{\partial \hat{p}_{i,y_i}} & = - \frac{1}{\hat{p}_{i,y_i}} \Big( (1-\hat{p}_{i,y_i})^\gamma - \gamma \hat{p}_{i,y_i} (1-\hat{p}_{i,y_i})^{\gamma -1} \log(\hat{p}_{i,y_i}) \Big) \nonumber \\
        & = \frac{\partial \loss_c}{\partial \hat{p}_{i,y_i}} g(\hat{p}_{i,y_i}, \gamma),
\end{split}
\end{equation}
in which $g(\hat{p}_{i,y_i}, \gamma) = (1-\hat{p}_{i,y_i})^\gamma - \gamma \hat{p}_{i,y_i} (1-\hat{p}_{i,y_i})^{\gamma -1} \log(\hat{p}_{i,y_i})$ and $ \frac{\partial \loss_c}{\partial \hat{p}_{i,y_i}}  = -\frac{1}{\hat{p}_{i,y_i}}$. It is thus straightforward to verify that if $g(\hat{p}_{i,y_i}, \gamma) \in [0,1]$, then $\norm{\frac{\partial  \loss_f}{\partial \hat{p}_{i,y_i}}} \leq \norm{\frac{\partial  \loss_c}{\partial \hat{p}_{i,y_i}}}$, which itself implies that $\norm{\frac{\partial  \loss_f}{\partial \bfw}} \leq \norm{\frac{\partial  \loss_c}{\partial \bfw}}$.
\end{proof}

\begin{prop2}
\label{pro:2}
Given a $p_0$, for $1 \geq p \geq p_0 > 0$, $g(p, \gamma) \leq 1$ for all $\gamma \geq \gamma^* = \frac{a}{b} + \frac{1}{\log a}W_{-1} \big(-\frac{a^{(1-a/b)}}{b} \log a \big)$, where $a = 1-p_0$, $b = p_0 \log p_0$, and $W_{-1}$ is the Lambert-W function~\citep{corless1996lambertw}. Moreover, for $p \geq p_0 > 0$ and $\gamma \geq \gamma^*$, the equality $g(p, \gamma) = 1$ holds only for $p = p_0$ and $\gamma = \gamma^*$.
\end{prop2}

\begin{proof}
We derive the value of $\gamma > 0$ for which $g(p_0, \gamma)=1$ for a given $p_0 \in [0, 1]$. From Proposition 4.1, we already know that
\begin{equation}
    \frac{\partial \loss_f}{\partial \hat{p}_{i,y_i}} = \frac{\partial \loss_c}{\partial \hat{p}_{i,y_i}} g(\hat{p}_{i,y_i}, \gamma),
\end{equation}
where $\loss_f$ is focal loss, $\loss_c$ is cross entropy loss, $\hat{p}_{i,y_i}$ is the probability assigned by the model to the ground-truth correct class for the $i^{th}$ sample, and
\begin{equation}
    g(\hat{p}_{i,y_i}, \gamma) = (1-\hat{p}_{i,y_i})^\gamma - \gamma \hat{p}_{i,y_i} (1-\hat{p}_{i,y_i})^{\gamma -1} \log(\hat{p}_{i,y_i}).
\end{equation}
For $p \in [0, 1]$, if we look at the function $g(p, \gamma)$, then we can clearly see that $g(p, \gamma) \rightarrow 1$ as $p \rightarrow 0$, and that $g(p, \gamma) = 0$ when $p = 1$. To observe the behaviour of $g(p, \gamma)$ for intermediate values of $p$, we first take its derivative with respect to $p$:
\begin{equation}
    \label{eq:11}
        \frac{\partial g(p, \gamma)}{\partial p} = \gamma (1-p)^{\gamma-2} \big[-2(1-p)-(1-p)\log p + (\gamma-1) p \log p\big]
\end{equation}
In Equation \ref{eq:11}, $\gamma(1-p)^{\gamma-2} > 0$ except when $p = 1$ (in which case $\gamma(1-p)^{\gamma-2} = 0$). Thus, to observe the sign of the gradient $\frac{\partial g(p, \gamma)}{\partial p}$, we focus on the term
\begin{equation}
\label{eq:12}
    -2(1-p)-(1-p)\log p + (\gamma-1)p \log p.
\end{equation}
Dividing Equation \ref{eq:12} by $(-\log p)$, the sign remains unchanged and we get
\begin{equation}
    k(p, \gamma) = \frac{2(1-p)}{\log p} + 1 - \gamma p.
\end{equation}
We can see that $k(p,\gamma) \rightarrow 1$ as $p \rightarrow 0$ and $k(p,\gamma) \rightarrow -(1+\gamma)$ as $p \rightarrow 1$ (using l'H{\^o}pital's rule). Furthermore, $k(p, \gamma)$ is monotonically decreasing for $p \in [0, 1]$. Thus, as the gradient $\frac{\partial g(p, \gamma)}{\partial p}$ is positive initially starting from $p = 0$ and negative later till $p = 1$, we can say that $g(p, \gamma)$ first monotonically increases starting from $1$ (as $p\rightarrow 0$) and then monotonically decreases down to $0$ (at $p = 1$). Thus, if for some threshold $p_0 > 0$ and for some $\gamma > 0$, $g(p, \gamma) = 1$, then $\forall p > p_0$, $g(p, \gamma) < 1$. We now want to find a $\gamma$ such that $\forall p \geq p_0$, $g(p, \gamma) \le 1$. First, let $a=(1-p_0)$ and $b=p_0\log p_0$. Then:
\begin{equation}
\begin{split}
\label{eq:exp_gamma}
    &g(p_0, \gamma) = (1-p_0)^\gamma-\gamma p_0 (1-p_0)^{\gamma-1}\log p_0 \le 1\\
    \implies &(1-p_0)^{\gamma-1}[(1-p_0)-\gamma p_0\log p_0] \le 1\\
    \implies &a^{\gamma -1}(a-\gamma b) \le 1\\
    \implies &(\gamma -1)\log a + \log (a-\gamma b) \le 0\\
    \implies &\Big( \gamma-\frac{a}{b} \Big) \log a+\log(a-\gamma b) \le \Big( 1-\frac{a}{b} \Big)\log a\\
    \implies &(a-\gamma b)e^{(\gamma -a/b)\log a} \le a^{(1-a/b)}\\
    \implies &\Big( \gamma-\frac{a}{b} \Big) e^{(\gamma -a/b)\log a} \le -\frac{a^{(1-a/b)}}{b}\\
    \implies &\Big( \Big( \gamma-\frac{a}{b} \Big) \log a\Big) e^{(\gamma -a/b)\log a} \ge -\frac{a^{(1-a/b)}}{b}\log a
\end{split}
\end{equation}
where $a=(1-p_0)$ and $b=p_0\log p_0$. We know that the inverse of $y=x e^x$ is defined as $x=W(y)$, where $W$ is the Lambert-W function~\citep{corless1996lambertw}. Furthermore, the r.h.s. of the inequality in Equation \ref{eq:exp_gamma} is always negative, with a minimum possible value of $-1/e$ that occurs at $p_0=0.5$. Therefore, applying the Lambert-W function to the r.h.s.\ will yield two real solutions (corresponding to a principal branch denoted by $W_0$ and a negative branch denoted by $W_{-1}$). We first consider the solution corresponding to the negative branch (which is the smaller of the two solutions):
\begin{equation}
\begin{split}
\label{eq:exp_gamma_cont}
    &\Big((\gamma-\frac{a}{b})\log a\Big) \le W_{-1}\Big(-\frac{a^{(1-a/b)}}{b}\log a\Big)\\
    \implies &\gamma \ge \frac{a}{b}+\frac{1}{\log a}W_{-1}\Big(-\frac{a^{(1-a/b)}}{b}\log a\Big)\\
\end{split}
\end{equation}
If we consider the principal branch, the solution is
\begin{equation}
    \gamma \le \frac{a}{b}+\frac{1}{\log a}W_{0}\Big(-\frac{a^{(1-a/b)}}{b}\log a\Big),
\end{equation}
which yields a negative value for $\gamma$ that we discard. Thus Equation \ref{eq:exp_gamma_cont} gives the values of $\gamma$ for which if $p>p_0$, then $g(p,\gamma) < 1$. In other words, $g(p_0, \gamma) = 1$, and for any $p < p_0$, $g(p, \gamma) > 1$.
\end{proof}

\section{Dataset Description and Implementation Details}
\label{dataset}
We use the following image and document classification datasets in our experiments:
\vspace{-1.5mm}
\begin{enumerate}[leftmargin=*]
\item \textbf{CIFAR-10} \citep{Krizhevsky2009}: This dataset has 60,000 colour images of size $32 \times 32$, divided equally into 10 classes. We use a train/validation/test split of 45,000/5,000/10,000 images.
\item \textbf{CIFAR-100} \citep{Krizhevsky2009}: This dataset has 60,000 colour images of size $32 \times 32$, divided equally into 100 classes. (Note that the images in this dataset are not the same images as in CIFAR-10.) We again use a train/validation/test split of 45,000/5,000/10,000 images.
\item \textbf{Tiny-ImageNet} \citep{deng2009imagenet}: Tiny-ImageNet is a subset of ImageNet with 64 x 64 dimensional images, 200 classes and 500 images per class in the training set and 50 images per class in the validation set. The image dimensions of Tiny-ImageNet are twice that of CIFAR-10/100 images.
\item \textbf{20 Newsgroups} \citep{Lang1995}: This dataset contains 20,000 news articles, categorised evenly into 20 different newsgroups based on their content. It is a popular dataset for text classification. Whilst some of the newsgroups are very related (e.g.\ rec.motorcycles and rec.autos), others are quite unrelated (e.g.\ sci.space and misc.forsale). We use a train/validation/test split of 15,098/900/3,999 documents.
\item \textbf{Stanford Sentiment Treebank (SST)} \citep{Socher2013}: This dataset contains movie reviews in the form of sentence parse trees, where each node is annotated by sentiment. We use the dataset version with binary labels, for which 6,920/872/1,821 documents are used as the training/validation/test split. In the training set, each node of a parse tree is annotated as positive, neutral or negative. At test time, the evaluation is done based on the model classification at the root node, i.e.\ considering the whole sentence, which contains only positive or negative sentiment.
\end{enumerate}
\vspace{-1.5mm}

All our experiments required a single 12 GB TITAN Xp GPU. For training networks on CIFAR-10 and CIFAR-100, we use SGD with a momentum of 0.9 as our optimiser, and train the networks for 350 epochs, with a learning rate of 0.1 for the first 150 epochs, 0.01 for the next 100 epochs, and 0.001 for the last 100 epochs. We use a training batch size of 128. Furthermore, we augment the training images by applying random crops and random horizontal flips. For Tiny-ImageNet, we train for 100 epochs with a learning rate of 0.1 for the first 40 epochs, 0.01 for the next 20 epochs and 0.001 for the last 40 epochs. We use a training batch size of 64. It should be noted that for Tiny-ImageNet, we saved 50 samples per class (i.e., a total of 10000 samples) from the training set as our own validation set to fine-tune the temperature parameter (hence, we trained on 90000 images) and we use the Tiny-ImageNet validation set as our test set.

For 20 Newsgroups, we train the Global Pooling Convolutional Network \citep{Lin2013} using the Adam optimiser, with learning rate $0.001$, and betas $0.9$ and $0.999$. The code is a PyTorch adaptation of \cite{Ng}. We used Glove word embeddings \citep{pennington2014glove} to train the network. We trained all the models for 50 epochs and used the models with the best validation accuracy.

For the SST Binary dataset, we train the Tree-LSTM~\citep{Tai2015} using the AdaGrad optimiser with a learning rate of $0.05$ and a weight decay of $10^{-4}$, as suggested by the authors. We used the constituency model, which considers binary parse trees of the data and trains a binary Tree-LSTM on them. The Glove word embeddings \citep{pennington2014glove} were also tuned for best results. The code framework we used is inspired by \cite{TreeLSTM}. We trained these models for 25 epochs and used the models with the best validation accuracy.

For all our models, we use the PyTorch framework, setting any hyperparameters not explicitly mentioned to the default values used in the standard models. For MMCE, we used $\lambda = 2$ for all our experiments as we found it to perform better over all the values we tried. A calibrated model which does not generalise well to an unseen test set is not very useful. Hence, for all the experiments, we set the training parameters in a way such that we get best test set accuracies on all datasets for each model.

\section{Additional Results}
\label{results}
In addition to the sample-dependent $\gamma$ approach, we try the following focal loss approaches as well:

\textbf{Focal Loss (Fixed $\gamma$)}: We trained models on focal loss with $\gamma$ fixed to $1, 2$ and $3$. We found $\gamma = 3$ to produce the best ECE among models trained using a fixed $\gamma$. This corroborates the observation we made in \S4 of the main paper that $\gamma = 3$ should produce better results than $\gamma=1$ or $\gamma=2$, as the regularising effect for $\gamma=3$ is higher.

\textbf{Focal Loss (Scheduled $\gamma$)}: As a simplification to the sample-dependent $\gamma$ approach, we also tried using a schedule for $\gamma$ during training, as we expect the value of $\hat{p}_{i,y_i}$ to increase in general for all samples over time. In particular, we report results for two different schedules: (a) Focal Loss (scheduled $\gamma$ 5,3,2): $\gamma=5$ for the first 100 epochs, $\gamma=3$ for the next 150 epochs, and $\gamma=2$ for the last 100 epochs, and (b) Focal Loss (scheduled $\gamma$ 5,3,1): $\gamma=5$ for the first 100 epochs, $\gamma=3$ for the next 150 epochs, and $\gamma=1$ for the last 100 epochs. We also tried various other schedules, but found these two to produce the best results on the validation sets.

Finally, for the sample-dependent $\gamma$ approach, we also found the policy: Focal Loss (sample-dependent $\gamma$ 5,3,2) with $\gamma=5$ for $\hat{p}_{i,y_i} \in [0, 0.2)$, $\gamma=3$ for $\hat{p}_{i,y_i} \in [0.2, 0.5)$ and $\gamma=2$ for $\hat{p}_{i, y_i} \in [0.5, 1]$ to produce competitive results.

In Tables~\ref{table:ada_ece_tab1} and \ref{table:sce_tab1}, we present the AdaECE and Classwise-ECE scores for all the baselines discussed in Table~\ref{table:ece_tab1} of the main paper.

In Table~\ref{table:error_tab1} of the main paper, we present the classification errors on the test datasets for all the major loss functions we considered. Here we also report the classification errors for the different focal loss approaches in Table~\ref{table:error_tab2}. We also report the ECE, Ada-ECE and Classwise-ECE for all the focal loss approaches in Table~\ref{table:ece_tab2}, Table~\ref{table:ada_ece_tab2} and Table~\ref{table:sce_tab2} respectively.

Finally, calibrated models should have a higher logit score (or softmax probability) on the correct class even when they misclassify, as compared to models which are less calibrated. Thus, intuitively, such models should have a higher Top-5 accuracy. In Table~\ref{table:top5}, we report the Top-5 accuracies for all our models on datasets where the number of classes is relatively high (i.e., on CIFAR-100 with 100 classes and Tiny-ImageNet with 200 classes). We observe focal loss with sample-dependent $\gamma$ to produce the highest top-5 accuracies on all models trained on CIFAR-100 and the second best top-5 accuracy (only marginally below the highest accuracy) on Tiny-ImageNet.

\begin{table*}[!t]
	\centering
	\scriptsize
	\resizebox{\linewidth}{!}{%
		\begin{tabular}{cccccccccccccc}
			\toprule
			\textbf{Dataset} & \textbf{Model} & \multicolumn{2}{c}{\textbf{Cross-Entropy}} &
			\multicolumn{2}{c}{\textbf{Brier Loss}} & \multicolumn{2}{c}{\textbf{MMCE}} &
			\multicolumn{2}{c}{\textbf{LS-0.05}} & \multicolumn{2}{c}{\textbf{FL-3 (Ours)}} &
			\multicolumn{2}{c}{\textbf{FLSD-53 (Ours)}} \\
			&& Pre T & Post T & Pre T & Post T & Pre T & Post T & Pre T & Post T & Pre T & Post T & Pre T & Post T \\
			\midrule
			
			\multirow{4}{*}{CIFAR-100} & ResNet-50&17.52&3.42(2.1)&6.52&3.64(1.1)&15.32&2.38(1.8)&7.81&4.01(1.1)&\tikzmark{top left}5.08&2.02(1.1)&\textbf{4.5}&\textbf{2.0(1.1)}\\
			& ResNet-110&19.05&5.86(2.3)&\textbf{7.73}&4.53(1.2)&19.14&4.85(2.3)&11.12&8.59(1.1)&8.64&4.14(1.2)&8.55&\textbf{3.96(1.2)}\\
			& Wide-ResNet-26-10&15.33&2.89(2.2)&4.22&2.81(1.1)&13.16&4.25(1.9)&5.1&5.1(1)&\textbf{2.08}&2.08(1)&2.75&\textbf{1.63(1.1)}\\
			& DenseNet-121&20.98&5.09(2.3)&5.04&2.56(1.1)&19.13&3.07(2.1)&12.83&8.92(1.2)&4.15&\textbf{1.23(1.1)}&\textbf{3.55}&\textbf{1.24(1.1)}\\
			\midrule
			\multirow{4}{*}{CIFAR-10} & ResNet-50&4.33&2.14(2.5)&1.74&\textbf{1.23(1.1)}&4.55&2.16(2.6)&3.89&2.92(0.9)&1.95&1.83(1.1)&\textbf{1.56}&1.26(1.1)\\
			& ResNet-110&4.4&1.99(2.8)&2.6&1.7(1.2)&5.06&2.52(2.8)&4.44&4.44(1)&\textbf{1.62}&\textbf{1.44(1.1)}&2.07&1.67(1.1)\\
			& Wide-ResNet-26-10&3.23&1.69(2.2)&1.7&1.7(1)&3.29&1.6(2.2)&4.27&2.44(0.8)&1.84&1.54(0.9)&\textbf{1.52}&\textbf{1.38(0.9)}\\
			& DenseNet-121&4.51&2.13(2.4)&2.03&2.03(1)&5.1&2.29(2.5)&4.42&3.33(0.9)&\textbf{1.22}&1.48(0.9)&1.42&\textbf{1.42(1)}\\
			\midrule
			Tiny-ImageNet & ResNet-50&15.23&5.41(1.4)&4.37&4.07(0.9)&13.0&5.56(1.3)&15.28&6.29(0.7)&1.88&1.88(1)&\textbf{1.42}&\textbf{1.42(1)}\\
			
			\midrule
			20 Newsgroups & Global Pooling CNN&17.91&\textbf{2.23(3.4)}&13.57&3.11(2.3)&15.21&6.47(2.2)&\textbf{4.39}&2.63(1.1)&8.65&3.78(1.5)&6.92&2.35(1.5)\\
			\midrule
			SST Binary & Tree-LSTM&7.27&3.39(1.8)&8.12&2.84(2.5)&\textbf{5.01}&4.32(1.5)&5.14&4.23(1.2)&16.01&2.16(0.5)&9.15&\textbf{1.92(0.7)}\tikzmark{bottom right}\\
			\bottomrule
		\end{tabular}%
	}
	\vspace{-2mm}
	\caption{Adaptive ECE $(\%)$ computed for different approaches both pre and post temperature scaling (cross-validating T on ECE). Optimal temperature for each method is indicated in brackets. \vspace{-3mm}}
	\label{table:ada_ece_tab1}
	\vspace{-0\baselineskip}
\end{table*}

\begin{table*}[!t]
	\centering
	\scriptsize
	\resizebox{\linewidth}{!}{%
		\begin{tabular}{cccccccccccccc}
			\toprule
			\textbf{Dataset} & \textbf{Model} & \multicolumn{2}{c}{\textbf{Cross-Entropy}} &
			\multicolumn{2}{c}{\textbf{Brier Loss}} & \multicolumn{2}{c}{\textbf{MMCE}} &
			\multicolumn{2}{c}{\textbf{LS-0.05}} & \multicolumn{2}{c}{\textbf{FL-3 (Ours)}} &
			\multicolumn{2}{c}{\textbf{FLSD-53 (Ours)}} \\
			&& Pre T & Post T & Pre T & Post T & Pre T & Post T & Pre T & Post T & Pre T & Post T & Pre T & Post T \\
			\midrule
			
			\multirow{4}{*}{CIFAR-100} & ResNet-50 & 0.38 & 0.22(2.1)&0.22&0.20(1.1)&0.34&0.21(1.8)&0.23&0.21(1.1)&\tikzmark{top left}\textbf{0.20}&\textbf{0.20(1.1)}&\textbf{0.20}&\textbf{0.20(1.1)}\\
			& ResNet-110&0.41&0.21(2.3)&0.24&0.23(1.2)&0.42&0.22(2.3)&0.26&0.22(1.1)&\textbf{0.24}&0.22(1.2)&\textbf{0.24}&\textbf{0.21(1.2)}\\
			& Wide-ResNet-26-10&0.34&0.20(2.2)&0.19&0.19(1.1)&0.31&0.20(1.9)&0.21&0.21(1)&\textbf{0.18}&\textbf{0.18(1)}&\textbf{0.18}&0.19(1.1)\\
			& DenseNet-121&0.45&0.23(2.3)&0.20&0.21(1.1)&0.42&0.24(2.1)&0.29&0.24(1.2)&0.20&0.20(1.1)&\textbf{0.19}&\textbf{0.20(1.1)}\\
			\midrule
			\multirow{4}{*}{CIFAR-10} & ResNet-50&0.91&0.45(2.5)&0.46&0.42(1.1)&0.94&0.52(2.6)&0.71&0.51(0.9)&0.43&0.48(1.1)&\textbf{0.42}&\textbf{0.42(1.1)}\\
			& ResNet-110&0.91&0.50(2.8)&0.59&0.50(1.2)&1.04&0.55(2.8)&0.66&0.66(1)&\textbf{0.44}&\textbf{0.41(1.1)}&0.48&0.44(1.1)\\
			& Wide-ResNet-26-10&0.68&0.37(2.2)&0.44&0.44(1)&0.70&0.35(2.2)&0.80&0.45(0.8)&0.44&0.36(0.9)&\textbf{0.41}&\textbf{0.31(0.9)}\\
			& DenseNet-121&0.92&0.47(2.4)&0.46&0.46(1)&1.04&0.57(2.5)&0.60&0.50(0.9)&0.43&0.41(0.9)&\textbf{0.41}&\textbf{0.41(1)}\\
			\midrule
			Tiny-ImageNet & ResNet-50&0.22&0.16(1.4)&0.16&0.16(0.9)&0.21&0.16(1.3)&0.21&0.17(0.7)&0.16&0.16(1)&\textbf{0.16}&\textbf{0.16(1)}\\
			
			\midrule
			20 Newsgroups & Global Pooling CNN&1.95&0.83(3.4)&1.56&\textbf{0.82(2.3)}&1.77&1.10(2.2)&\textbf{0.93}&0.91(1.1)&1.31&1.05(1.5)&1.40&1.19(1.5)\\
			\midrule
			SST Binary & Tree-LSTM&5.81&3.76(1.8)&6.38&2.48(2.5)&\textbf{3.82}&\textbf{2.70(1.5)}&3.99&3.20(1.2)&6.35&2.81(0.5)&4.84&3.24(0.7)\tikzmark{bottom right}\\
			\bottomrule
		\end{tabular}%
	}
	\vspace{-2mm}
	\caption{Classwise-ECE $(\%)$ computed for different approaches both pre and post temperature scaling (cross-validating T on ECE). Optimal temperature for each method is indicated in brackets. \vspace{-3mm}}
	\label{table:sce_tab1}
	\vspace{-0\baselineskip}
\end{table*}

In addition to ECE, Ada-ECE and Classwise-ECE, we use various other metrics to compare the proposed methods with the baselines (i.e.\ cross-entropy, Brier loss, MMCE and Label Smoothing). We present the test NLL \% before and after temperature scaling in Tables \ref{table:nll_tab1} and \ref{table:nll_tab2}, respectively. We report the test set MCE \% before and after temperature scaling in Tables \ref{table:mce_tab1} and \ref{table:mce_tab2}, respectively. 

We use the following abbreviation to report results on different varieties of Focal Loss. FL-1 refers to Focal Loss (fixed $\gamma$ 1), FL-2 refers to Focal Loss (fixed $\gamma$ 2), FL-3 refers to Focal Loss (fixed $\gamma$ 3), FLSc-531 refers to Focal Loss (scheduled $\gamma$ 5,3,1),  FLSc-532 refers to Focal Loss (scheduled $\gamma$ 5,3,2),  FLSD-532 refers to Focal Loss (sample-dependent $\gamma$ 5,3,2)  and FLSD-53 refers to Focal Loss (sample-dependent $\gamma$ 5,3).

\begin{table*}[!htb]
\renewcommand{\arraystretch}{1.3}
\centering
\small
\resizebox{\linewidth}{!}{%
\begin{tabular}{cccccccccccccccc}
\toprule
\textbf{Dataset} & \textbf{Model} & \multicolumn{2}{c}{\textbf{FL-1}} & \multicolumn{2}{c}{\textbf{FL-2}} & \multicolumn{2}{c}{\textbf{FL-3}} & \multicolumn{2}{c}{\textbf{FLSc-531}} & \multicolumn{2}{c}{\textbf{FLSc-532}} & \multicolumn{2}{c}{\textbf{FLSD-532}} & \multicolumn{2}{c}{\textbf{FLSD-53}}\\
&& Pre T & Post T & Pre T & Post T & Pre T & Post T & Pre T & Post T & Pre T & Post T & Pre T & Post T & Pre T & Post T \\

\midrule
\multirow{4}{*}{CIFAR-100} & ResNet-50&12.86&2.3(1.5)&8.61&2.24(1.3)&5.13&1.97(1.1)&11.63&2.09(1.4)&8.47&2.13(1.3)&9.09&1.61(1.3)&4.5&2.(1.1)\\
& ResNet-110&15.08&4.55(1.5)&11.57&3.73(1.3)&8.64&3.95(1.2)&14.99&4.56(1.5)&11.2&3.43(1.3)&11.74&3.64(1.3)&8.56&4.12(1.2)\\
& Wide-ResNet-26-10&8.93&2.53(1.4)&4.64&2.93(1.2)&2.13&2.13(1)&9.36&2.48(1.4)&4.98&1.94(1.2)&4.98&2.55(1.2)&3.03&1.64(1.1)\\
& DenseNet-121&14.24&2.8(1.5)&7.9&2.33(1.2)&4.15&1.25(1.1)&13.05&2.08(1.5)&7.63&1.96(1.2)&8.14&2.35(1.3)&3.73&1.31(1.1)\\
\midrule
\multirow{4}{*}{CIFAR-10} & ResNet-50&3.42&1.08(1.6)&2.36&0.91(1.2)&1.48&1.42(1.1)&4.06&1.53(1.6)&2.97&1.53(1.2)&2.52&0.88(1.3)&1.55&0.95(1.1)\\
& ResNet-110&3.46&1.2(1.6)&2.7&0.89(1.3)&1.55&1.02(1.1)&4.92&1.5(1.7)&3.33&1.36(1.3)&2.82&0.97(1.3)&1.87&1.07(1.1)\\
& Wide-ResNet-26-10&2.69&1.46(1.3)&1.42&1.03(1.1)&1.69&0.97(0.9)&2.81&0.96(1.4)&1.82&1.45(1.1)&1.31&0.87(1.1)&1.56&0.84(0.9)\\
& DenseNet-121&3.44&1.63(1.4)&1.93&1.04(1.1)&1.32&1.26(0.9)&4.12&1.65(1.5)&2.22&1.34(1.1)&2.45&1.31(1.2)&1.22&1.22(1)\\
\midrule
Tiny-ImageNet & ResNet-50&7.61&3.29(1.2)&3.02&3.02(1)&1.87&1.87(1)&7.77&3.07(1.2)&3.62&2.54(1.1)&2.81&2.57(1.1)&1.76&1.76(1)\\
\midrule
20 Newsgroups & Global Pooling CNN&15.06&2.14(2.6)&12.1&3.22(1.6)&8.67&3.51(1.5)&13.55&4.32(1.7)&12.13&2.47(1.8)&12.2&2.39(2)&6.92&2.19(1.5)\\
\midrule
SST Binary & Tree-LSTM&6.78&3.29(1.6)&3.05&3.05(1)&16.05&1.78(0.5)&4.66&3.36(1.4)&3.91&2.64(0.9)&4.47&2.77(0.9)&9.19&1.83(0.7)\\
\bottomrule

\end{tabular}}
\caption{ECE $(\%)$ computed for different focal loss approaches both pre and post temperature scaling (cross-validating T on ECE). Optimal temperature for each method is indicated in brackets. \vspace{-3mm}}
\label{table:ece_tab2}
\end{table*}

\begin{table*}[!htb]
\renewcommand{\arraystretch}{1.3}
\centering
\footnotesize
\resizebox{\linewidth}{!}{%
\begin{tabular}{cccccccccccccccc}
\toprule
\textbf{Dataset} & \textbf{Model} & \multicolumn{2}{c}{\textbf{FL-1}} & \multicolumn{2}{c}{\textbf{FL-2}} & \multicolumn{2}{c}{\textbf{FL-3}} & \multicolumn{2}{c}{\textbf{FLSc-531}} & \multicolumn{2}{c}{\textbf{FLSc-532}} & \multicolumn{2}{c}{\textbf{FLSD-532}} & \multicolumn{2}{c}{\textbf{FLSD-53}}\\
&& Pre T & Post T & Pre T & Post T & Pre T & Post T & Pre T & Post T & Pre T & Post T & Pre T & Post T & Pre T & Post T \\

\midrule
\multirow{4}{*}{CIFAR-100} & ResNet-50&12.86&2.54(1.5)&8.55&2.44(1.3)&5.08&2.02(1.1)&11.58&2.01(1.4)&8.41&2.25(1.3)&9.08&1.94(1.3)&4.39&2.33(1.1)\\
& ResNet-110&15.08&5.16(1.5)&11.57&4.46(1.3)&8.64&4.14(1.2)&14.98&4.97(1.5)&11.18&3.68(1.3)&11.74&4.21(1.3)&8.55&3.96(1.2)\\
& Wide-ResNet-26-10&8.93&2.74(1.4)&4.65&2.96(1.2)&2.08&2.08(1)&9.2&2.52(1.4)&5&2.11(1.2)&5&2.58(1.2)&2.75&1.63(1.1)\\
& DenseNet-121&14.24&2.71(1.5)&7.9&2.36(1.2)&4.15&1.23(1.1)&13.01&2.18(1.5)&7.61&2.04(1.2)&8.04&2.1(1.3)&3.55&1.24(1.1)\\
\midrule
\multirow{4}{*}{CIFAR-10} & ResNet-50&3.42&1.51(1.6)&2.37&1.69(1.2)&1.95&1.83(1.1)&4.06&2.43(1.6)&2.95&2.18(1.2)&2.5&1.23(1.3)&1.56&1.26(1.1)\\
& ResNet-110&3.42&1.57(1.6)&2.69&1.29(1.3)&1.62&1.44(1.1)&4.91&2.61(1.7)&3.32&1.92(1.3)&2.78&1.58(1.3)&2.07&1.67(1.1)\\
& Wide-ResNet-26-10&2.7&1.71(1.3)&1.64&1.47(1.1)&1.84&1.54(0.9)&2.75&1.85(1.4)&2.04&1.9(1.1)&1.68&1.49(1.1)&1.52&1.38(0.9)\\
& DenseNet-121&3.44&1.85(1.4)&1.8&1.39(1.1)&1.22&1.48(0.9)&4.11&2.2(1.5)&2.19&1.64(1.1)&2.44&1.6(1.2)&1.42&1.42(1)\\
\midrule
Tiny-ImageNet & ResNet-50&7.56&2.95(1.2)&3.15&3.15(1)&1.88&1.88(1)&7.7&2.9(1.2)&3.76&2.4(1.1)&2.81&2.6(1.1)&1.42&1.42(1)\\
\midrule
20 Newsgroups & Global Pooling CNN&15.06&2.22(2.6)&12.1&3.33(1.6)&8.65&3.78(1.5)&13.55&4.58(1.7)&12.13&2.49(1.8)&12.19&2.37(2)&6.92&2.35(1.5)\\
\midrule
SST Binary & Tree-LSTM&6.27&4.59(1.6)&3.69&3.69(1)&16.01&2.16(0.5)&4.43&3.57(1.4)&3.37&2.46(0.9)&4.42&2.96(0.9)&9.15&1.92(0.7)\\
\bottomrule
\end{tabular}}
\caption{AdaECE $(\%)$ computed for different focal loss approaches both pre and post temperature scaling (cross-validating T on ECE). Optimal temperature for each method is indicated in brackets.  \vspace{-3mm}}
\label{table:ada_ece_tab2}
\end{table*}

\begin{table*}[!htb]
\renewcommand{\arraystretch}{1.3}
\centering
\footnotesize
\resizebox{\linewidth}{!}{%
\begin{tabular}{cccccccccccccccc}
\toprule
\textbf{Dataset} & \textbf{Model} & \multicolumn{2}{c}{\textbf{FL-1}} & \multicolumn{2}{c}{\textbf{FL-2}} & \multicolumn{2}{c}{\textbf{FL-3}} & \multicolumn{2}{c}{\textbf{FLSc-531}} & \multicolumn{2}{c}{\textbf{FLSc-532}} & \multicolumn{2}{c}{\textbf{FLSD-532}} & \multicolumn{2}{c}{\textbf{FLSD-53}}\\
&& Pre T & Post T & Pre T & Post T & Pre T & Post T & Pre T & Post T & Pre T & Post T & Pre T & Post T & Pre T & Post T \\

\midrule
\multirow{4}{*}{CIFAR-100} & ResNet-50&0.30&0.21(1.5)&0.24&0.22(1.3)&0.20&0.20(1.1)&0.28&0.21(1.4)&0.24&0.21(1.3)&0.25&0.21(1.3)&0.20&0.20(1.1)\\
& ResNet-110&0.34&0.21(1.5)&0.28&0.20(1.3)&0.24&0.22(1.2)&0.34&0.22(1.5)&0.28&0.21(1.3)&0.28&0.21(1.3)&0.24&0.21(1.2)\\
& Wide-ResNet-26-10&0.23&0.19(1.4)&0.18&0.21(1.2)&0.18&0.18(1)&0.24&0.19(1.4)&0.19&0.20(1.2)&0.19&0.19(1.2)&0.18&0.19(1.1)\\
& DenseNet-121&0.33&0.22(1.5)&0.23&0.21(1.2)&0.20&0.20(1.1)&0.31&0.22(1.5)&0.23&0.20(1.2)&0.23&0.22(1.3)&0.19&0.20(1.1)\\
\midrule
\multirow{4}{*}{CIFAR-10} & ResNet-50&0.73&0.40(1.6)&0.54&0.45(1.2)&0.43&0.48(1.1)&0.85&0.51(1.6)&0.64&0.48(1.2)&0.56&0.42(1.3)&0.42&0.42(1.1)\\
& ResNet-110&0.74&0.44(1.6)&0.61&0.42(1.3)&0.44&0.41(1.1)&1.02&0.62(1.7)&0.73&0.51(1.3)&0.63&0.44(1.3)&0.48&0.44(1.1)\\
& Wide-ResNet-26-10&0.59&0.39(1.3)&0.37&0.36(1.1)&0.44&0.36(0.9)&0.61&0.37(1.4)&0.42&0.42(1.1)&0.37&0.33(1.1)&0.41&0.31(0.9)\\
& DenseNet-121&0.74&0.44(1.4)&0.45&0.39(1.1)&0.43&0.41(0.9)&0.88&0.47(1.5)&0.52&0.43(1.1)&0.56&0.47(1.2)&0.41&0.41(1)\\
\midrule
Tiny-ImageNet & ResNet-50&0.18&0.17(1.2)&0.16&0.16(1)&0.16&0.16(1)&0.18&0.17(1.2)&0.16&0.16(1.1)&0.16&0.17(1.1)&0.16&0.16(1)\\
\midrule
20 Newsgroups & Global Pooling CNN&1.73&0.98(2.6)&1.51&1.00(1.6)&1.31&1.05(1.5)&1.59&1.15(1.7)&1.54&1.10(1.8)&1.50&0.99(2)&1.40&1.19(1.5)\\
\midrule
SST Binary & Tree-LSTM&5.46&3.63(1.6)&3.34&3.34(1)&6.35&2.81(0.5)&4.32&3.60(1.4)&3.95&3.90(0.9)&3.52&3.39(0.9)&4.84&3.24(0.7)\\
\bottomrule
\end{tabular}}
\caption{Classwise-ECE $(\%)$ computed for different focal loss approaches both pre and post temperature scaling (cross-validating T on ECE). Optimal temperature for each method is indicated in brackets.  \vspace{-3mm}}
\label{table:sce_tab2}
\end{table*}

\begin{table*}[!htb]
\renewcommand{\arraystretch}{1.3}
\centering
\footnotesize
\resizebox{\linewidth}{!}{%
\begin{tabular}{ccccccccc}
\toprule

\textbf{Dataset} & \textbf{Model} & \textbf{FL-1} & \textbf{FL-2} & \textbf{FL-3} & \textbf{FLSc-531} & \textbf{FLSc-532} & \textbf{FLSD-532} & \textbf{FLSD-53}\\
\midrule
\multirow{4}{*}{CIFAR-100} & ResNet-50&22.8&23.15&22.75&23.49&23.24&23.55&23.22\\
& ResNet-110&22.36&22.53&22.92&22.81&22.96&22.93&22.51\\
& Wide-ResNet-26-10&19.61&20.01&19.69&20.13&20.13&19.71&20.11\\
& DenseNet-121&23.82&23.19&23.25&23.69&23.72&22.41&22.67\\
\midrule
\multirow{4}{*}{CIFAR-10} & ResNet-50&4.93&4.98&5.25&5.66&5.63&5.24&4.98\\
& ResNet-110&4.78&5.06&5.08&6.13&5.71&5.19&5.42\\
& Wide-ResNet-26-10&4.27&4.27&4.13&4.11&4.46&4.14&4.01\\
& DenseNet-121&5.09&4.84&5.33&5.46&5.65&5.46&5.46\\
\midrule
Tiny-ImageNet & ResNet-50&50.06&47.7&49.69&50.49&49.83&48.95&49.06\\
\midrule
20 Newsgroups & Global Pooling CNN&26.13&28.23&29.26&29.16&28.16&27.26&27.98\\
\midrule
SST Binary & Tree-LSTM&12.63&12.3&12.19&12.36&13.07&12.3&12.8\\
\bottomrule
\end{tabular}}
\caption{Error $(\%)$ computed for different focal loss approaches. \vspace{-3mm}}
\label{table:error_tab2}
\end{table*}

\begin{table*}[!htb]
\renewcommand{\arraystretch}{1.3}
\centering
\footnotesize
\resizebox{\linewidth}{!}{%
\begin{tabular}{cccccccccccc}
\toprule
\textbf{Dataset} & \textbf{Model} & \multicolumn{2}{c}{\textbf{Cross-Entropy}} & \multicolumn{2}{c}{\textbf{Brier Loss}} & \multicolumn{2}{c}{\textbf{MMCE}} & \multicolumn{2}{c}{\textbf{LS-0.05}} & \multicolumn{2}{c}{\textbf{FLSD-53 (Ours)}}\\
&& Top-1 & Top-5 & Top-1 & Top-5 & Top-1 & Top-5 & Top-1 & Top-5 & Top-1 & Top-5 \\

\midrule
\multirow{4}{*}{CIFAR-100} & ResNet-50&76.7&93.77&76.61&93.24&76.8&93.69&76.57&92.86&76.78&\textbf{94.44}\\
& ResNet-110&77.27&93.79&74.9&92.44&76.93&93.78&76.57&92.27&77.49&\textbf{94.78}\\
& Wide-ResNet-26-10&79.3&93.96&79.41&94.56&79.27&94.11&78.81&93.18&79.89&\textbf{95.2}\\
& DenseNet-121&75.48&91.33&76.25&92.76&76&91.96&75.95&89.51&77.33&\textbf{94.49}\\
\midrule
Tiny-ImageNet & ResNet-50&50.19&74.24&46.8&70.34&48.69&73.52&52.88&\textbf{76.15}&50.94&76.07\\
\bottomrule
\end{tabular}}
\caption{Top-1 and Top-5 accuracies computed for different approaches.\vspace{-3mm}}
\label{table:top5}
\end{table*}

\begin{table*}[!htb]
\renewcommand{\arraystretch}{1.3}
\centering
\footnotesize
\resizebox{\linewidth}{!}{%
\begin{tabular}{ccccccccccccc}
\toprule
\textbf{Dataset} & \textbf{Model} & \textbf{Cross-Entropy} & \textbf{Brier Loss} & \textbf{MMCE} & \textbf{LS-0.05} & \textbf{FL-1} & \textbf{FL-2} & \textbf{FL-3} & \textbf{FLSc-531} & \textbf{FLSc-532} & \textbf{FLSD-532} & \textbf{FLSD-53}\\ 
\midrule
\multirow{4}{*}{CIFAR-100} & ResNet-50&153.67&99.63&125.28&121.02&105.61&92.82&87.52&100.09&92.66&94.1&88.03\\
& ResNet-110&179.21&110.72&180.54&133.11&114.18&96.74&90.9&112.46&95.85&97.97&89.92\\
& Wide-ResNet-26-10&140.1&84.62&119.58&108.06&87.56&77.8&74.66&88.61&78.52&78.86&76.92\\
& DenseNet-121&205.61&98.31&166.65&142.04&115.5&93.11&87.13&107.91&93.12&91.14&85.47\\
\midrule
\multirow{4}{*}{CIFAR-10} & ResNet-50&41.21&18.67&44.83&27.68&22.67&18.6&18.43&25.32&20.5&18.69&17.55\\
& ResNet-110&47.51&20.44&55.71&29.88&22.54&19.19&17.8&32.77&22.48&19.39&18.54\\
& Wide-ResNet-26-10&26.75&15.85&28.47&21.71&17.66&14.96&15.2&18.5&15.57&14.78&14.55\\
& DenseNet-121&42.93&19.11&52.14&28.7&22.5&17.56&18.02&27.41&19.5&20.14&18.39\\
\midrule
Tiny-ImageNet & ResNet-50&232.85&240.32&234.29&235.04&219.07&202.92&207.2&217.52&211.42&204.71&204.97\\
\midrule
20 Newsgroups & Global Pooling CNN&176.57&130.41&158.7&90.95&140.4&115.97&109.62&128.75&123.72&124.03&109.17\\
\midrule
SST Binary & Tree-LSTM&50.2&54.96&37.28&44.34&53.9&47.72&50.29&50.25&53.13&45.08&49.23\\
\bottomrule
\end{tabular}}
\caption{NLL $(\%)$ computed for different approaches pre temperature scaling. \vspace{-3mm}}
\label{table:nll_tab1}
\end{table*}

\begin{table*}[!t]
\renewcommand{\arraystretch}{1.3}
\centering
\footnotesize
\resizebox{\linewidth}{!}{%
\begin{tabular}{ccccccccccccc}
\toprule
\textbf{Dataset} & \textbf{Model} & \textbf{Cross-Entropy} & \textbf{Brier Loss} & \textbf{MMCE} & \textbf{LS-0.05} & \textbf{FL-1} & \textbf{FL-2} & \textbf{FL-3} & \textbf{FLSc-531} & \textbf{FLSc-532} & \textbf{FLSD-532} & \textbf{FLSD-53}\\ 
\midrule
\multirow{4}{*}{CIFAR-100} & ResNet-50&106.83&99.57&101.92&120.19&94.58&91.80&87.37&92.77&91.58&92.83&88.27\\
& ResNet-110&104.63&111.81&106.73&129.76&94.65&91.24&89.92&93.73&91.30&92.29&88.93\\
& Wide-ResNet-26-10&97.10&85.77&95.92&108.06&83.68&80.44&74.66&84.11&80.01&80.40&78.14\\
& DenseNet-121&119.23&98.74&113.24&136.28&100.81&91.35&87.55&98.16&91.55&90.57&86.06\\
\midrule
\multirow{4}{*}{CIFAR-10} & ResNet-50&20.38&18.36&21.58&27.69&17.56&17.67&18.34&19.93&19.25&17.28&17.37\\
& ResNet-110&21.52&19.60&24.61&29.88&17.32&17.53&17.62&23.79&20.21&17.78&18.24\\
& Wide-ResNet-26-10&15.33&15.85&16.16&21.19&15.48&14.85&15.06&15.81&15.38&14.69&14.23\\
& DenseNet-121&21.77&19.11&24.88&28.95&18.71&17.21&18.10&21.65&19.04&19.27&18.39\\
\midrule
Tiny-ImageNet & ResNet-50&220.98&238.98&226.29&214.95&217.51&202.92&207.20&215.37&211.57&205.42&204.97\\
\midrule
20 Newsgroups & Global Pooling CNN&87.95&93.11&99.74&90.42&87.24&93.60&94.69&97.89&93.66&91.73&93.98\\
\midrule
SST Binary & Tree-LSTM&41.05&38.27&36.37&43.45&45.67&47.72&45.96&45.82&54.52&45.36&49.69\\
\bottomrule
\end{tabular}}
\caption{NLL $(\%)$ computed for different approaches post temperature scaling (cross-validating T on ECE). \vspace{-3mm}}
\label{table:nll_tab2}
\end{table*}

\begin{table*}[!htb]
\renewcommand{\arraystretch}{1.3}
\centering
\footnotesize
\resizebox{\linewidth}{!}{%
\begin{tabular}{ccccccccccccc}
\toprule
\textbf{Dataset} & \textbf{Model} & \textbf{Cross-Entropy} & \textbf{Brier Loss} & \textbf{MMCE} & \textbf{LS-0.05} & \textbf{FL-1} & \textbf{FL-2} & \textbf{FL-3} & \textbf{FLSc-531} & \textbf{FLSc-532} & \textbf{FLSD-532} & \textbf{FLSD-53}\\ 
\midrule
\multirow{4}{*}{CIFAR-100} & ResNet-50&44.34&36.75&39.53&26.11&33.22&21.03&13.02&26.76&23.56&22.4&16.12\\
& ResNet-110&55.92&24.85&50.69&36.23&40.49&32.57&26&37.24&29.56&34.73&22.57\\
& Wide-ResNet-26-10&49.36&14.68&40.13&23.79&27&15.14&9.96&27.81&17.59&13.64&10.17\\
& DenseNet-121&56.28&15.47&49.97&43.59&35.45&21.7&11.61&38.68&18.91&21.34&9.68\\
\midrule
\multirow{4}{*}{CIFAR-10} & ResNet-50&38.65&31.54&60.06&35.61&31.75&25&21.83&30.54&23.57&25.45&14.89\\
& ResNet-110&44.25&25.18&67.52&45.72&73.35&25.92&25.15&34.18&30.38&30.8&18.95\\
& Wide-ResNet-26-10&48.17&77.15&36.82&24.89&29.17&30.17&23.86&37.57&30.65&18.51&74.07\\
& DenseNet-121&45.19&19.39&43.92&45.5&38.03&29.59&77.08&33.5&16.47&17.85&13.36\\
\midrule
Tiny-ImageNet & ResNet-50&30.83&8.41&26.48&25.48&20.7&8.47&6.11&16.03&9.28&8.97&3.76\\
\midrule
20 Newsgroups & Global Pooling CNN&36.91&31.35&34.72&8.93&34.28&24.1&18.85&26.02&25.02&24.29&17.44\\
\midrule
SST Binary & Tree-LSTM&71.08&92.62&68.43&39.39&95.48&86.21&22.32&76.28&86.93&80.85&73.7\\
\bottomrule
\end{tabular}}
\caption{MCE $(\%)$ computed for different approaches pre temperature scaling. \vspace{-3mm}}
\label{table:mce_tab1}
\end{table*}

\begin{table*}[!htb]
\renewcommand{\arraystretch}{1.3}
\centering
\footnotesize
\resizebox{\linewidth}{!}{%
\begin{tabular}{ccccccccccccc}
\toprule
\textbf{Dataset} & \textbf{Model} & \textbf{Cross-Entropy} & \textbf{Brier Loss} & \textbf{MMCE} & \textbf{LS-0.05} & \textbf{FL-1} & \textbf{FL-2} & \textbf{FL-3} & \textbf{FLSc-531} & \textbf{FLSc-532} & \textbf{FLSD-532} & \textbf{FLSD-53}\\ 
\midrule
\multirow{4}{*}{CIFAR-100} & ResNet-50&12.75&21.61&11.99&18.58&8.92&8.86&6.76&7.46&6.76&5.24&27.18\\
& ResNet-110&22.65&13.56&19.23&30.46&20.13&12&13.06&18.28&13.72&15.89&10.94\\
& Wide-ResNet-26-10&14.18&13.42&16.5&23.79&10.28&18.32&9.96&13.18&11.01&12.5&9.73\\
& DenseNet-121&21.63&8.55&13.02&29.95&10.49&11.63&6.17&6.21&6.48&9.41&5.68\\
\midrule
\multirow{4}{*}{CIFAR-10} & ResNet-50&20.6&22.46&23.6&40.51&25.86&28.17&15.76&22.05&23.85&24.76&26.37\\
& ResNet-110&29.98&22.73&31.87&45.72&29.74&23.82&37.61&26.25&25.94&11.59&17.35\\
& Wide-ResNet-26-10&26.63&77.15&32.33&37.53&74.58&29.58&25.64&28.63&20.23&19.68&36.56\\
& DenseNet-121&32.52&19.39&27.03&53.57&19.68&22.71&76.27&21.05&32.76&35.06&13.36\\
\midrule
Tiny-ImageNet & ResNet-50&13.33&12.82&12.52&17.2&6.5&8.47&6.11&5.97&7.01&5.73&3.76\\
\midrule
20 Newsgroups & Global Pooling CNN&36.91&31.35&34.72&8.93&34.28&24.1&18.85&26.02&25.02&24.29&17.44\\
\midrule
SST Binary & Tree-LSTM&88.48&91.86&32.92&35.72&87.77&86.21&74.52&54.27&88.85&82.42&76.71\\
\bottomrule
\end{tabular}}
\caption{MCE $(\%)$ computed for different approaches post temperature scaling (cross-validating T on ECE). \vspace{-3mm}}
\label{table:mce_tab2}
\end{table*}

\newpage
\section{Bar plots}
\label{sec:barplots}

\begin{figure*}[!t]
    \centering
    \includegraphics[width=\linewidth]{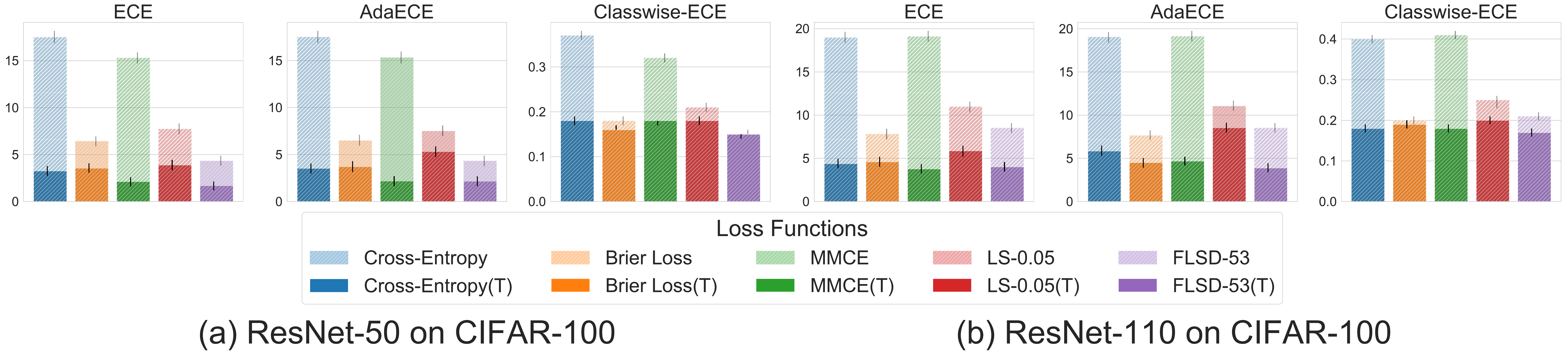}
    \vspace{-\baselineskip}
	\caption{Bar plots with confidence intervals for ECE, AdaECE and Classwise-ECE computed for ResNet-50 (first $3$ figures) and ResNet-110 (last $3$ figures) on CIFAR-100.}
  \label{fig:error_ba2}
  \vspace{-1\baselineskip}
\end{figure*}

\begin{figure*}[!t]
    \centering
    \includegraphics[width=\linewidth]{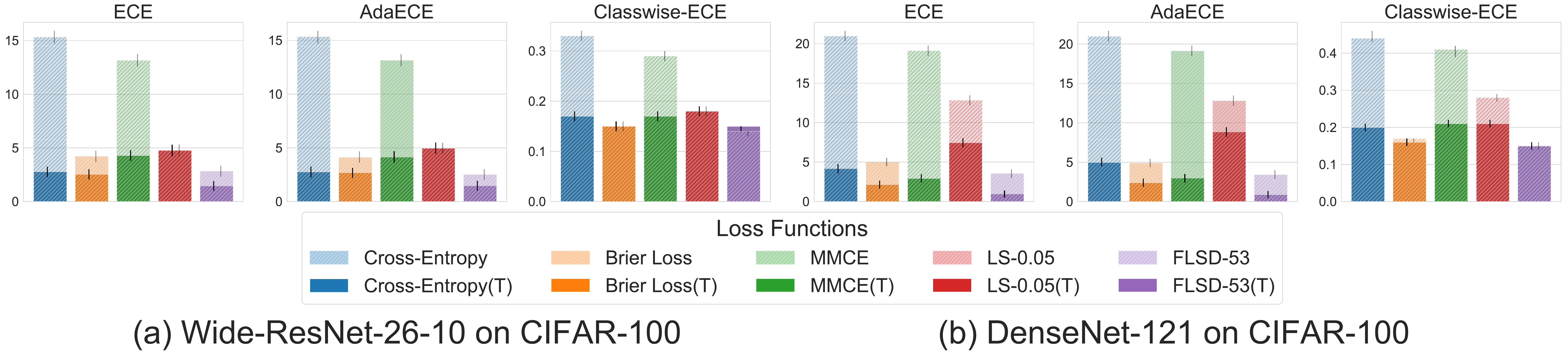}
    \vspace{-\baselineskip}
	\caption{Bar plots with confidence intervals for ECE, AdaECE and Classwise-ECE computed for Wide-ResNet-26-10 (first $3$ figures) and DenseNet-121 (last $3$ figures) on CIFAR-100.}
  \label{fig:error_ba3}
  \vspace{-1\baselineskip}
\end{figure*}

\begin{figure*}[!t]
    \centering
    \includegraphics[width=\linewidth]{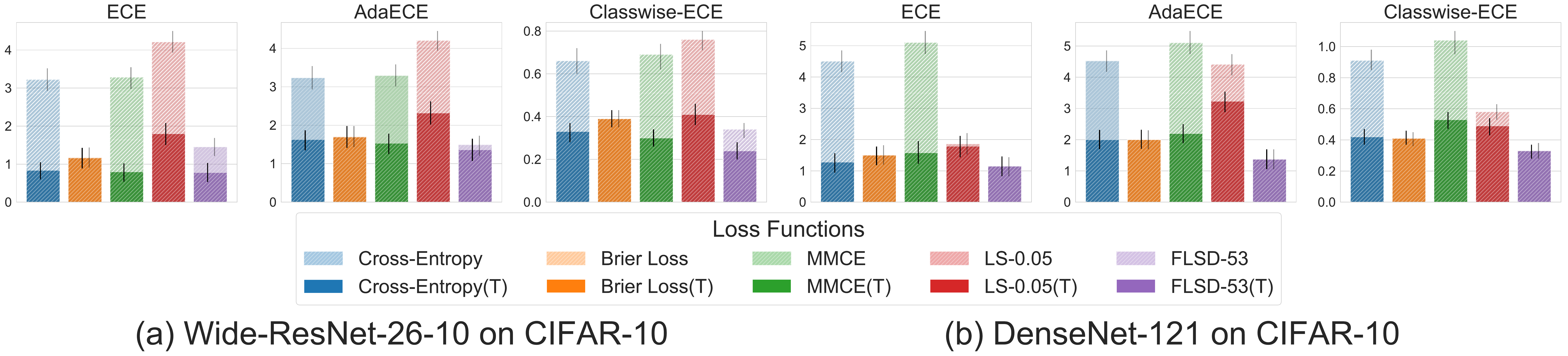}
    \vspace{-\baselineskip}
	\caption{Bar plots with confidence intervals for ECE, AdaECE and Classwise-ECE computed for Wide-ResNet-26-10 (first $3$ figures) and DenseNet-121 (last $3$ figures) on CIFAR-10.}
  \label{fig:error_ba4}
  \vspace{-1\baselineskip}
\end{figure*}

\begin{figure*}[!t]
    \centering
    \includegraphics[width=\linewidth]{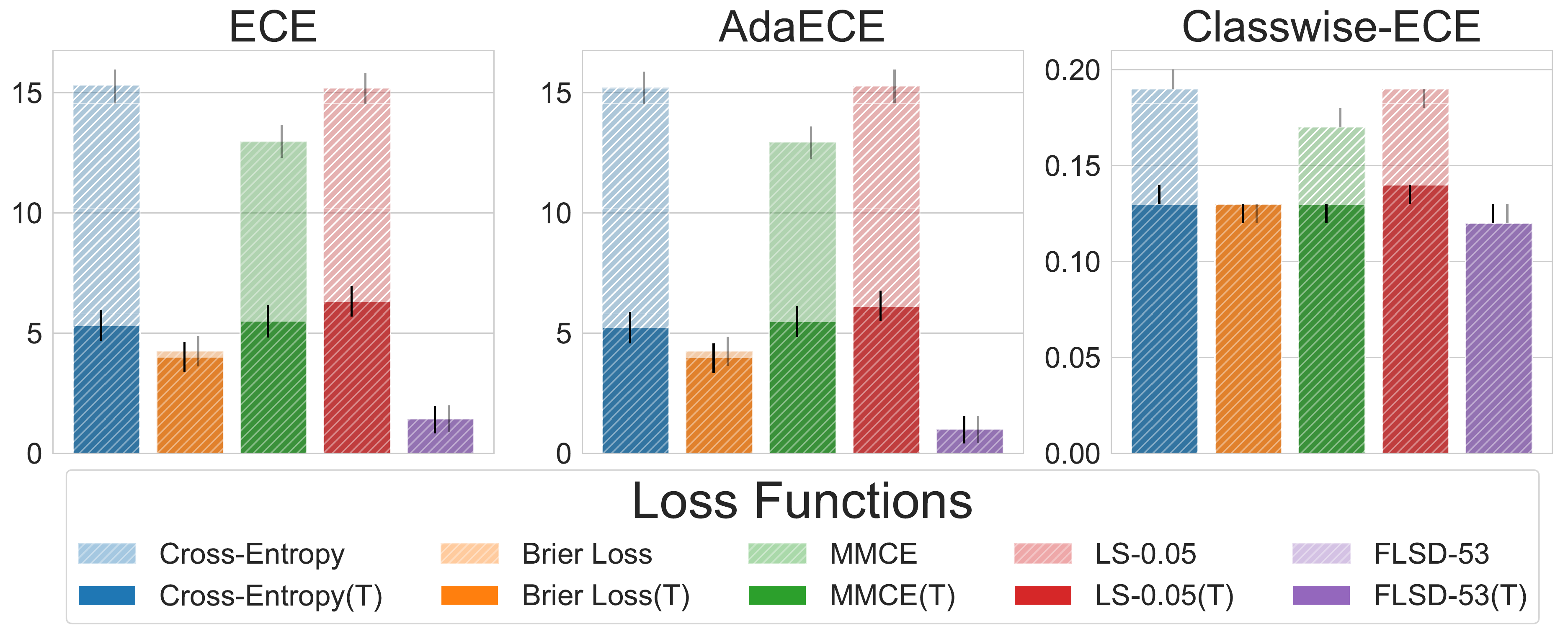}
    \vspace{-\baselineskip}
	\caption{Bar plots with confidence intervals for ECE, AdaECE and Classwise-ECE computed for ResNet-50 on Tiny-ImageNet.}
  \label{fig:error_ba5}
  \vspace{-1\baselineskip}
\end{figure*}

In this section, we present additional results in reference to Figure~\ref{fig:error_ba} in the main paper. In particular, we compute 90\% confidence intervals for ECE, AdaECE and Classwise-ECE using 1000 bootstrap samples following \cite{Kumar2019verified} and present the resulting confidence intervals as bar plots in Figures~\ref{fig:error_ba2}, ~\ref{fig:error_ba3}, ~\ref{fig:error_ba4} and ~\ref{fig:error_ba5}. These plots further corroboate the observations made in Section~\ref{sec:experiments} of the main paper. We find that FLSD-53 broadly produces the lowest calibration errors, and in quite a few cases (especially before temperature scaling) the differences in calibration errors between cross-entropy and focal loss are statistically significant.
\label{sec:bar_plots}

\section{Focal Loss is Confident and Calibrated}
\label{conf_and_cal}
\begin{table*}[!t]
\centering
\scriptsize
\resizebox{\linewidth}{!}{%
\begin{tabular}{cccccccccccccc}
\toprule
\textbf{Dataset} & \textbf{Model} & \multicolumn{2}{c}{\textbf{Cross-Entropy (Pre T)}} & \multicolumn{2}{c}{\textbf{Cross-Entropy (Post T)}} & \multicolumn{2}{c}{\textbf{MMCE (Pre T)}} & \multicolumn{2}{c}{\textbf{MMCE (Post T)}} & \multicolumn{2}{c}{\textbf{Focal Loss (Pre T)}} & \multicolumn{2}{c}{\textbf{Focal Loss (Post T)}} \\
&& |S99|$\%$ & Accuracy & |S99|$\%$ & Accuracy & |S99|$\%$ & Accuracy & |S99|$\%$ & Accuracy & |S99|$\%$ & Accuracy & |S99|$\%$ & Accuracy \\
\midrule
CIFAR-10 & ResNet-110 & $97.11$ & $96.33$ & $11.5$ & $97.39$ & $97.65$ & $96.72$ & $10.62$ & $99.83$ & $61.41$ & $99.51$ & $31.10$ & $99.68$ \\ 
CIFAR-10 & ResNet-50 & $95.93$ & $96.72$ & $7.33$ & $99.73$ & $92.33$ & $98.24$ & $4.21$ & $100$ & $46.31$ & $99.57$ & $14.27$ & $99.93$ \\ 
\bottomrule
\end{tabular}}
\caption{Percentage of test samples predicted with confidence higher than $99\%$ and the corresponding accuracy for Cross Entropy, MMCE and Focal loss computed both pre and post temperature scaling (represented in the table as pre T and post T respectively).}
\label{table:side_table}
\vspace{-1\baselineskip}
\end{table*}
\begin{figure*}[!b]
\centering
\includegraphics[width=\linewidth]{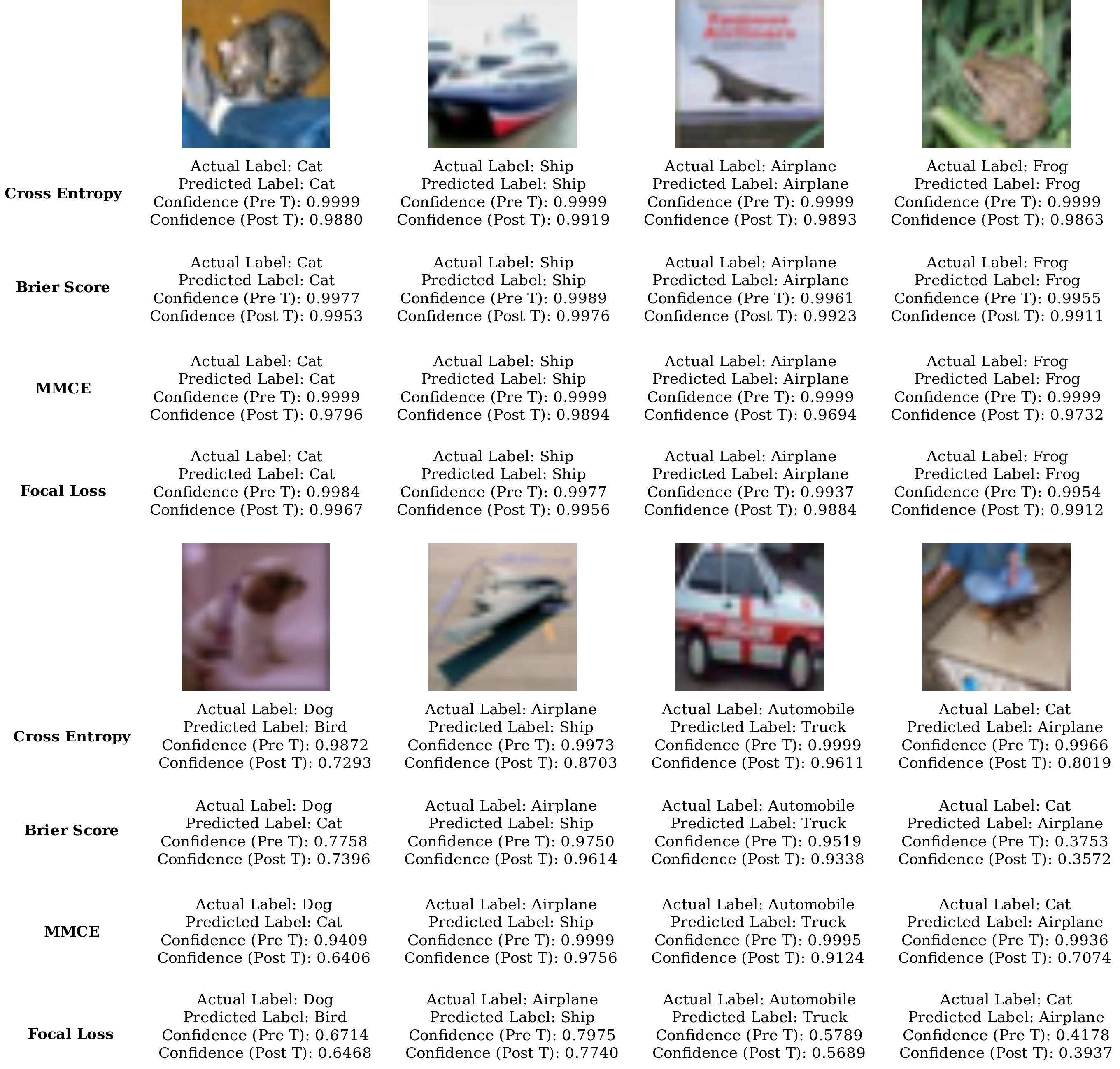}
\caption{Qualitative results showing the performance of Cross Entropy, Brier loss, MMCE and Focal Loss (sample-dependent $\gamma$ 5,3) for a ResNet-50 trained on CIFAR-10. The first row of images have been correctly classified by networks trained on all four loss functions and the second row of images have all been incorrectly classified. For each image, we present the actual label, the predicted label and the confidence of the prediction both before and after temperature scaling.
}
\label{fig:qualitative}
\vspace{-\baselineskip}
\end{figure*}

In extension to what we present in Section 5 of the main paper, we also follow the approach adopted in \cite{Kumar2018}, and measure the percentage of test samples that are predicted with a confidence of 0.99 or more (we call this set of test samples $S99$). In Table \ref{table:side_table}, we report $|S99|$ as a percentage of the total number of test samples, along with the accuracy of the samples in $S99$ for ResNet-50 and ResNet-110 trained on CIFAR-10, using cross-entropy loss, MMCE loss, and focal loss. We observe that $|S99|$ for the focal loss model is much lower than for the cross-entropy or MMCE models before temperature scaling. However, after temperature scaling, $|S99|$ for focal loss is significantly higher than for both MMCE and cross-entropy. The reason is that with an optimal temperature of 1.1, the confidence of the temperature-scaled model for focal loss does not reduce as much as those of the models for cross-entropy and MMCE, for which the optimal temperatures lie between 2.5 to 2.8. We thus conclude that models trained on focal loss are not only more calibrated, but also better preserve their confidence on predictions, even after being post-processed with temperature scaling.

In Figure \ref{fig:qualitative}, we present some qualitative results to support this claim and show the improvement in the confidence estimates of focal loss in comparison to other baselines (i.e., cross entropy, MMCE and Brier loss). For this, we take ResNet-50 networks trained on CIFAR-10 using the four loss functions (cross entropy, MMCE, Brier loss and Focal loss with sample-dependent $\gamma$ 5,3) and measure the confidence of their predictions for four correctly and four incorrectly classified test samples. We report these confidences both before and after temperature scaling. It is clear from Figure \ref{fig:qualitative} that for all the correctly classified samples, the model trained using focal loss has very confident predictions both pre and post temperature scaling. However, on misclassified samples, we observe a very low confidence for the focal loss model. The ResNet-50 network trained using cross entropy is very confident even on the misclassified samples, particularly before temperature scaling. Apart from focal loss, the only model which has relatively low confidences on misclassified test samples is the one trained using Brier loss. These observations support our claim that focal loss produces not only a calibrated model but also one which is confident on its correct predictions. 
\section{Ordering of Feature Norms}
\label{feature_norm}
\begin{figure}[b]
    \centering
    \includegraphics[width=0.5\linewidth]{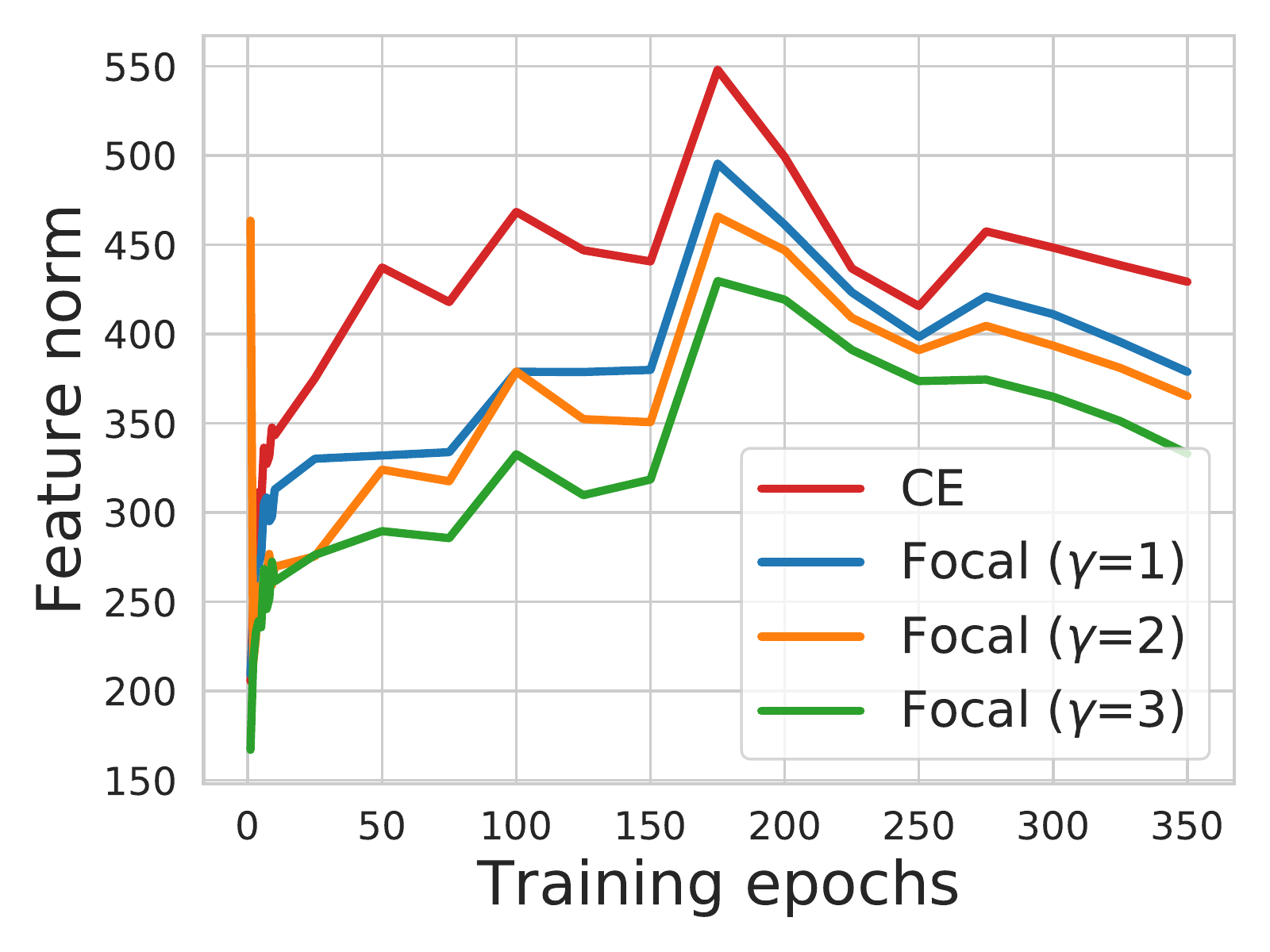}
    \caption{$L_{2}$ norm of features obtained from the last ResNet block (before the linear layer) of ResNet-50 averaged over entire training dataset of CIFAR-10 using a batch size of 128.}
    \label{fig:featureNorm}
\end{figure}
As an extension to the discussion related to Figure \ref{fig:nll_corr_incorr_entropy}(e) in the main paper, we plot the $L_2$ norm of the features/activations obtained from the last ResNet block (right before the linear layer is applied on these features to get the logits). We plot these norms throughout the training period for networks trained on cross-entropy and focal loss with $\gamma$ set to $1, 2$ and $3$ in Figure~\ref{fig:featureNorm}. We observe that there is a distinct ordering of feature norms for the four models: cross-entropy has the highest feature norm, followed by focal loss with $\gamma=1$, followed by focal loss with $\gamma = 2$ and finally focal loss with $\gamma = 3$. Furthermore, this ordering is preserved throughout training. As we saw from Figure \ref{fig:nll_corr_incorr_entropy}(e) in the main paper, from epoch 150 onwards (i.e., the epoch from which the networks start getting miscalibrated), there is a flip in the ordering of weight norms of the last linear layer. From epoch 150 onwards, the weight norms also follow the exact same ordering that we observe from Figure~\ref{fig:featureNorm} here. 
This shows that throughout training the initial layer weights (before last linear layer) of the network trained using focal loss are also regularized to favor lower norm of the output features, thus possibly leading to less peakiness in final prediction as compared to that of cross-entropy loss (see the `Peak at the wrong place' paragraph of Section 3 of the main paper).

\section{Early stopping}
\label{sec:early_stopping}
From Figure \ref{fig:nll_corr_incorr_entropy}(a), one may think that intermediate models (using early stopping) might provide better accuracy and calibration. However, there is no ideal approach for early stopping. For fair comparison, we train ResNet50, CIFAR-10 using cross-entrpy and focal loss with the best (in hindsight) possible early stopping. We train each model for 350 epochs and choose the 3 intermediate models with the best val set ECE, NLL and classification error, respectively. We present the test set performance in Table~\ref{table:early_stopping_table}.

\begin{table}[!t]
	\centering
	\scriptsize
	\begin{tabular}{ccccc}
		\toprule
		\textbf{Criterion} & \textbf{Loss} & \textbf{Epoch} & \textbf{Error} & \textbf{ECE \%}\\
		\midrule
		ECE & CE & 151 & 7.34 & 1.69 \\
		ECE & FLSD-53 & 257 & 5.52 & 0.85 \\
		NLL & CE & 153 & 6.69 & 2.28 \\
		NLL & FLSD-53 & 266 & 5.34 & 1.33 \\
		Error & CE & 344 & 5.0 & 4.46 \\
		Error & FLSD-53 & 343 & 4.99 & 1.43 \\
		\midrule
		Full & CE & 350 & 4.95 & 4.35 \\
		Full & FLSD-53 & 350 & 4.98 & 1.55 \\
		\bottomrule
	\end{tabular}
	\caption{Classification errors and ECE scores obtained from ResNet-50 models trained using cross-entropy and focal loss with different early stopping criteria (best in hindsight ECE, NLL and classification error on the validation set) applied during training. In the table CE and FL stand for cross-entropy and focal loss respectively and the Full Criterion indicates models where early stopping has not been applied.}
	\label{table:early_stopping_table}
	\vspace{-2mm}
\end{table}

From the table, we can observe that: 1) On every early stopping criterion, the model trained on focal loss outperforms the one trained on cross-entropy in both error and ECE, 2) ECE as a stopping criterion provides better test set ECE, but increases the test error significantly, 3) even without early stopping, focal loss achieves consistently better error and ECE compared to cross-entropy using any stopping criterion.

\section{Machine Translation: A Downstream Task}
\label{sec:downstream_task}

In this section, we explore machine translation with beam search as a relevant downstream task for calibration. Following the setup in \cite{muller2019does}, we train the Transformer architecture \citep{vaswani2017attention} on the WMT 2014 English-to-German translation dataset. The training settings (like optimiser, LR schedule, etc.) are the same as \cite{vaswani2017attention}. We chose machine translation as a relevant task because the softmax vectors produced by the transformer model are directly fed into the beam search algorithm, and hence softmax outputs from a calibrated model should intuitively produce better translations and a better BLEU score.

\begin{figure}[t!]
    \centering
    \includegraphics[width=0.7\linewidth]{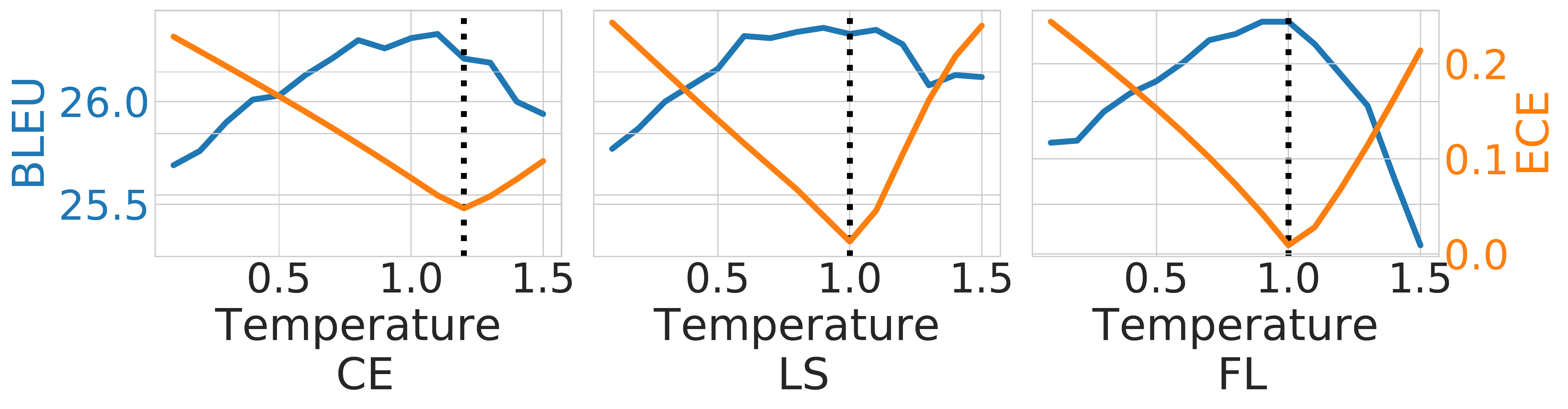}
    \vspace{-2mm}
    \caption{Changes in \emph{test set} BLEU score and \emph{validation set} ECE with temperature, for models trained using (a) cross-entropy with hard targets (CE) (b) cross-entropy with label smoothing (LS) ($\alpha = 0.1$), and (c) focal loss (FL) ($\gamma = 1$).}
    \label{fig:mt_plots}
\end{figure}

We train three transformer models, one on cross-entropy with hard target labels, the second on cross-entropy with label smoothing (with smoothing factor $\alpha = 0.1$) and the third on focal loss with $\gamma = 1$. In order to compare these models in terms of calibration, we report the test set ECE (\%) both before and after temperature scaling in the first row of Table \ref{table:mt_table}. Furthermore, to evaluate their performance on the English-to-German translation task, we also report the test set BLEU score of these models in the second row of Table \ref{table:mt_table}. Finally, to study the variation of test set BLEU score and validation set ECE with temperature, we plot them against temperature for all three models in Figure \ref{fig:mt_plots}.
\begin{table}[!b]
	\centering
	\scriptsize
	\begin{tabular}{cccc}
		\toprule
		\textbf{Metrics} & \textbf{CE ($\alpha = 0.0$)} & \textbf{LS ($\alpha = 0.1$)} & \textbf{FL ($\gamma = 1.0$)} \\
		\midrule
		ECE$\%$ Pre T / Post T / T & $10.16/2.59/1.2$ & $3.25/3.25/1.0$ & $\bm{1.69}/\bm{1.69}/1.0$ \\ 
		BLEU Pre T / Post T & $26.31/26.21$ & $26.33/26.33$ & $\bm{26.39}/\bm{26.39}$ \\ 
		\bottomrule
	\end{tabular}
	\caption{Test set ECE and BLEU score both pre and post temperature scaling for cross-entropy (CE) with hard targets, cross-entropy with label smoothing (LS) ($\alpha = 0.1$) and focal loss (FL) ($\gamma = 1$).}
	\label{table:mt_table}
\end{table}

We observe from Table \ref{table:mt_table} that the model trained on focal loss outperforms its competitors on ECE and also has a competitive edge over other methods on BLEU score as well. The focal loss model also has an optimal temperature of 1, just like the model trained on cross-entropy with label smoothing. From Figure~\ref{fig:mt_plots}, we can see that the models obtain the highest BLEU scores at around the same temperatures at which they obtain low ECEs, thereby confirming our initial notion that a more calibrated model provides better translations. However, since the optimal temperatures are tuned on the validation set, they don't often correspond to the best BLEU scores on the test set. On the test set, the highest BLEU scores we observe are 26.33 for cross-entropy, 26.36 for cross-entropy with label smoothing, and 26.39 for focal loss. Thus, the performance of focal loss on machine translation (a downstream task related to calibration) is also very encouraging.

\end{document}